%% file: main.tex
\documentclass{article} %
\usepackage{iclr2027_conference,times}
\iclrfinalcopy

\input{math_commands.tex}

\usepackage{hyperref}
\usepackage{url}
\usepackage{graphicx}
\usepackage{booktabs}
\usepackage{longtable}
\usepackage{adjustbox}
\usepackage[most]{tcolorbox}
\usepackage{fvextra}
\usepackage{multirow}
\usepackage{amsmath,amssymb}
\usepackage[table]{xcolor}
\usepackage{subcaption}
\usepackage{tikz}
\usepackage{pdflscape}   %
\usepackage{soul}
\usepackage{etoc}
\usepackage{float}
\usepackage{array}
\usepackage{tabularx}
\usepackage[normalem]{ulem}
\usepackage{enumitem}
\usepackage{xspace}
\usetikzlibrary{positioning,arrows.meta,fit,backgrounds,calc}
\input{macros}

\input{tables/captions}
\usepackage[section]{placeins} 
\usepackage{xurl}
\usepackage{longtable} 
\usepackage{fontawesome5}  %
\definecolor{lenscolor}{HTML}{1F7A8C}   %
\definecolor{ideacolor}{HTML}{E3A008}
\definecolor{prosecolor}{HTML}{9B4F96}
\newcommand{\aflens}{\textcolor{lenscolor}{\faLightbulb}}     %
\newcommand{\afbulb}{\textcolor{ideacolor}{\faSearch}}  %
\newcommand{\afpen}{\textcolor{prosecolor}{\faPenNib}}     %
\newcommand{\hficon}{\raisebox{-0.15em}
{\includegraphics[height=1.0em]{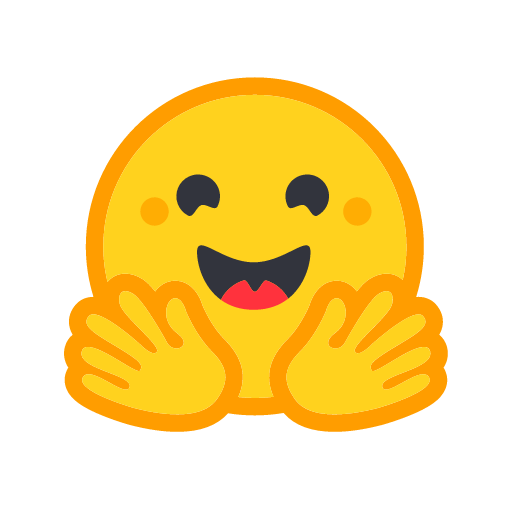}}}
\newcommand{\authorrow}[1]{\hspace*{-\tabcolsep}\makebox[\linewidth][c]{#1}}
\hypersetup{pdftitle={IdeaLens: Detecting AI Ideas in Long-form Writing}}
\hypersetup{pdfborder={0 0 0}}

\title{{\centering\fontsize{15}{18}\selectfont IdeaLens: Detecting AI Ideas in Long-form Writing}}

\author{%
\authorrow{%
  Rishanth Rajendhran\textsuperscript{\aflens}\hspace{1.5em}%
  Minjoon Choi\textsuperscript{\aflens}\hspace{1.5em}%
  Jenna Russell\textsuperscript{\aflens}\hspace{1.5em}%
  Ramya Namuduri\textsuperscript{\aflens}}\\[3pt]
\authorrow{\bfseries
  Deniz B\"{o}l\"{o}ni-Turgut\textsuperscript{\aflens}\hspace{1.5em}%
  Marzena Karpinska\textsuperscript{\afpen}\hspace{1.5em}%
  John Wieting\textsuperscript{\afbulb}\hspace{1.5em}%
  Mohit Iyyer\textsuperscript{\aflens}}\\[10pt]
\authorrow{\normalfont\small
  \aflens\ University of Maryland\hspace{2em}%
  \afbulb\ Google DeepMind\hspace{2em}%
  \afpen\ Simon Fraser University}\\[4pt]
\authorrow{\normalfont\small\texttt{\{rishanth, miyyer\}@umd.edu}}\\[8pt]
\authorrow{\normalfont\small
  \href{https://github.com/RishanthRajendhran/IdeaLens}{\textcolor{lenscolor}{\faGithub}\;\textsc{Code}}%
  \hspace{1.2em}\textcolor{black!25}{\textbar}\hspace{1.2em}%
  \href{https://huggingface.co/collections/rishanthrajendhran/idealens-6abee785ce6196fc0be9200f}{\hficon\;\textsc{Models \& Data}}%
  \hspace{1.2em}\textcolor{black!25}{\textbar}\hspace{1.2em}%
  \href{http://ideadetector.ai/}{\textcolor{lenscolor}{\faGlobe}\;\textsc{Demo}}}%
}

\newcommand{\model}{\textsc{IdeaLens}\xspace}
\newcommand{\twinmodel}{\textsc{ProseLens}\xspace}
\newcommand{\modelModernBERT}[0]{\textsc{IdeaLens-ModernBERT-L}\xspace}
\newcommand{\twinmodelModernBERT}[0]{\textsc{ProseLens-ModernBERT-L}\xspace}
\newcommand{\modelQwen}[0]{\textsc{IdeaLens-Qwen3.5-9B}\xspace}

\newcommand{\modelLR}[0]{\textsc{IdeaLens-LogisticClassifier}\xspace}

\newcommand{\aiipset}[0]{\twicetold}      %
\newcommand{\collabLadder}[0]{\textsc{IdeaShift}\xspace}
\newcommand{\xLadder}[0]{\textsc{IdeaShift-X}\xspace}
\newcommand{\twicetold}[0]{\textsc{TwiceTold}\xspace}

\begin{document}

\maketitle

\lhead{Preprint}

\etocdepthtag.toc{mainpaper}
\vspace{-1em}
\input{sections/00_abstract}

\input{sections/01_introduction}

\input{sections/03_method}

\input{sections/04_experiments}

\input{sections/05_results}

\input{sections/06_analysis}

\input{sections/07_related_work}
\input{sections/09_conclusion}

\input{sections/10_statements}

\bibliography{references}
\bibliographystyle{iclr2027_conference}

\appendix
\input{sections/appendix}

\end{document}

%% file: math_commands.tex
\usepackage{amsmath,amsfonts,bm}

\def\eqref#1{equation~\ref{#1}}

\def\1{\bm{1}}

\DeclareMathAlphabet{\mathsfit}{\encodingdefault}{\sfdefault}{m}{sl}
\SetMathAlphabet{\mathsfit}{bold}{\encodingdefault}{\sfdefault}{bx}{n}

%% file: macros.tex
\newcommand{\corpus}{\textsc{WildOutlines}}   %
\newcommand{\nemo}{Nemotron-30B}
\newcommand{\mb}{ModernBERT-L}

\newcommand{\featscore}{\textsc{Human-association Score}}
\newcommand{\featscores}{\textsc{Human-association Scores}}

\definecolor{ours}{HTML}{1F4E8C}
\definecolor{twin}{HTML}{C4622D}
\definecolor{pangram}{HTML}{8A8A8A}
\definecolor{muted}{HTML}{6B6B6B}
\definecolor{rule}{HTML}{CCCCCC}
\definecolor{cellgrey}{HTML}{EDEDED}

\definecolor{target}{HTML}{B03A2E}
\definecolor{gap}{HTML}{7B3F8C}
\definecolor{base}{HTML}{2E7D4F}
\definecolor{ok}{HTML}{2E7D4F}
\definecolor{ink}{HTML}{222222}

%% file: tables/captions.tex
\newcommand{\capAblations}{\textbf{In-domain results and ablations} for the two outline-trained models. Top: the held-out test split. Middle: the same models read a transformed outline, with the 1\% cut refitted on the transformed calibration split; the roles-only row also gives a separately trained roles-only \mb{}. Bottom: the share of verdicts that survive paraphrasing, sentence-level paraphrase of the source and a change of extractor, and the cost of reading a 500-word excerpt.}

\newcommand{\capMultiling}{\textbf{\xLadder{}, selected languages (\%).} FineWeb2 pages in each language (human: false-positive rate), a
model writing natively in that language from one line naming the document type and broad subject (level 0: TPR), and a model rendering the page's outline in that language with nothing
added (level 5: false-positive rate under the provenance rule). \model{} reads English outlines extracted from every language; it has never
seen a non-English document. Global 1\% cut for our models, Pangram~4 at its published threshold. All 24 languages in \S\ref{app:multiling}.}

\newcommand{\capProbe}{\textbf{Paraphrasing removes most of the surface signal.} Pangram~4 scores 1,000 held-out test documents three ways: as the document, the raw outline, and the paraphrased outline. We report the mean and median (shaded) of Pangram's human, AI-assisted, and AI fractions and of P(AI) $=$ AI $+\,\frac{1}{2}$ AI-assisted, in percent. On paraphrased outlines, Pangram rates almost every human document as AI, while \model{} still separates them. AUC is against the Pangram~3.3.2 labels, so the document row is partly circular.}

\newcommand{\capQone}{\textbf{Human ideas, human prose: accuracy (\%).} Ordinary human writing, so accuracy is the share labeled human. The top two rows give each detector's mean accuracy in this quadrant and in the other three, so no detector can look strong by always predicting one label. Outline column: global 1\% cut\,/\,per-format cut. $^{\dagger}$All documents under 500 words. All arms are in \S\ref{app:evals}.}

\newcommand{\capQfour}{\textbf{Human ideas, model prose: accuracy (\%).} A model wrote, rewrote, polished, or translated a person's content, so accuracy is the share labeled human. The top two rows give each detector's mean accuracy in this quadrant and in the other three, so no detector can look strong by always predicting one label. Outline column: global 1\% cut\,/\,per-format cut. $^{\dagger}$All documents under 500 words. All arms are in \S\ref{app:evals}.}

\newcommand{\capQthree}{\textbf{Model ideas, model prose: accuracy (\%).} Conventional machine-text detection; accuracy is TPR, and attacked or humanized arms are marked. PeerPrism \citep{peerprism} and the personal-style set \citep{baumler-etal-2026-make} are left out (\S\ref{app:benchmarks}). $^{\dagger}$every document under 500 words. Every arm of every evaluation is in \S\ref{app:evals}.}

\newcommand{\capExtractBench}{\textbf{Outline extraction benchmark.} Twelve extractors on 198 documents, scored against Claude Fable 5 reference outlines. We report (a) order-preserving match of outline items (\emph{alignment}) (b) reference/candidate items matched (\emph{coverage}) (c) one-to-one matches (\emph{granularity agreement}) (d) candidate minus reference items per match (\emph{split$-$merge}) (e) exact label match (\emph{role agreement}) (f) cost per document with batching (g) mean rank (1 is best) and mean score from GPT-5.6 Sol, which judges eight configurations per document against the reference without seeing their source; a cross-check on the alignment score (``--'': not judged). Corpus extractors are in bold.
}

%% file: sections/00_abstract.tex
\begin{abstract}

While modern AI detectors identify \textbf{\emph{who wrote the words}}, emerging policies on AI use increasingly hinge on a different question:  \textbf{\emph{who came up with the ideas?}} We introduce \model, a detector that identifies whether a document's ideas came from a human or AI (\emph{idea provenance}), regardless of who wrote its words. To focus \model on ideas rather than prose, we represent documents as outlines: lists of items that each pair a discourse role with a brief, paraphrased description of the content, minimizing word-level overlap with the raw text. We train \model on 1M FineWeb documents with silver labels from Pangram, a prose provenance detector. Since the outlines are largely stripped of surface-level information, the labels must be fit mainly through the ideas.
In a controlled study, \model's AI flag rate drops from 95\% to 7\% as models write from increasingly detailed human plans, while Pangram 4 still flags 92\%; from AI-derived plans, \model stays above 96\%. Conversely, on a new dataset of 50 stories that human authors wrote from AI-generated plans, \model flags 68\% of the stories as AI, compared to 8\% for Pangram 4. On a comprehensive suite of 19 existing detection benchmarks, we show that \model maintains strong detection rates at low false positive rates, suggesting that ideas themselves provide a powerful discriminative signal, and its performance holds across domains, formats, and languages.
Finally, we examine 90K predictions from \model to characterize systematic differences between human and AI ideation. We release our models and labeled datasets to facilitate future research on idea provenance detection.

\end{abstract}

%% file: sections/01_introduction.tex
\section{Introduction}
\label{sec:intro}

\begin{figure}[t]
\centering
\includegraphics[width=\textwidth]{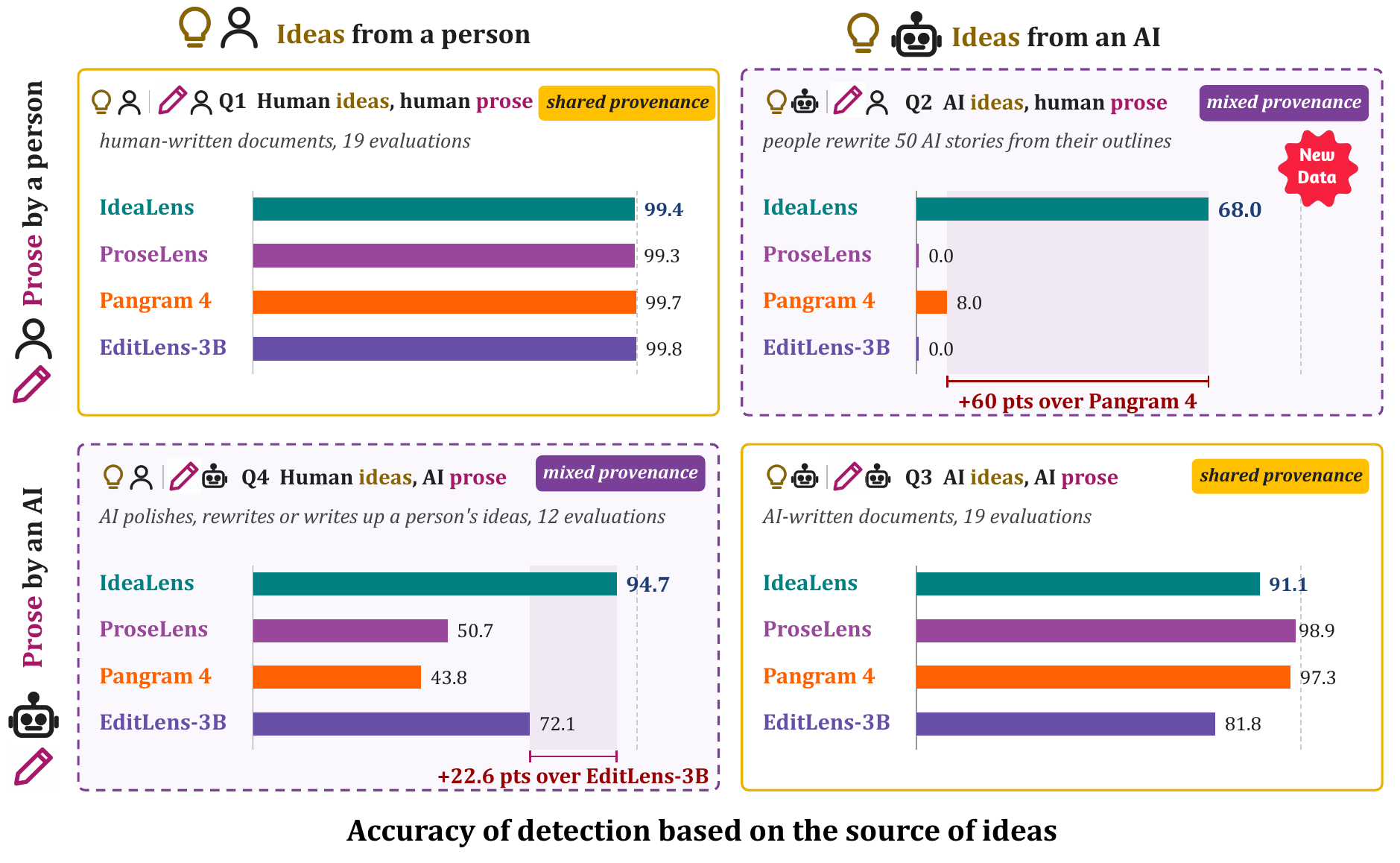}
\caption{
    \textbf{\model\ tracks ideas; traditional detectors track wording.} We present the four quadrants covering both \textit{shared} provenance (fully human or fully AI) and \textit{mixed} provenance (human ideas and AI prose, or AI ideas and human prose) scenarios. Unlike traditional detectors (e.g., Pangram 4, \citealp{glickenhaus2026pangram4technicalreport}; EditLens-3B, \citealp{thai2026editlens}), \model\ can accurately detect the source of ideas, not the source of the prose. \model\ classifies 94.7\% of AI-generated documents based on human ideas as human, while Pangram labels only 43.8\% as human. On our new dataset of human-written stories following AI ideas (\S\ref{sec:experiments}), \model\ flags 68.0\% as AI, compared to Pangram and EditLens flagging only 8.0\% and 0.0\%, respectively.
}
\label{fig:teaser}
\end{figure}

In August 2026, the AI detector Pangram flagged a \emph{Wall Street Journal}  op-ed \citep{druckenmiller2026bond} by billionaire investor Stanley Druckenmiller as 100\% AI-generated, to which
Druckenmiller unapologetically responded: ``Of course I used AI''~\citep{stein2026druckenmiller}. Paul Gigot, the editor of WSJ's editorial pages, defended Druckenmiller by arguing
that what matters is whether an op-ed ``reflects an author's original argument'' and that ``nobody can doubt
that his op-ed is his genuine opinion''~\citep{wapo2026wsjai}, though he provided no supporting evidence from the writing process (e.g., chat logs). Detectors like Pangram answer only \emph{who wrote the words} (\textbf{prose provenance}), while Gigot's defense, like many emerging policies on AI use in writing \citep{springernature2026ai, iclr2027aipolicy}, hinges on \emph{who came up with the ideas} (\textbf{idea provenance}).\footnote{\model\ flags the op-ed's ideas as AI-generated, with P(AI)  of $98.3\%$
against a decision threshold of 86.3\%. This is a model prediction and of course does
not constitute direct evidence of the writing process.}

Idea and prose provenance can differ within the same document: a human can brainstorm the ideas for a document and ask an AI to write it, or vice versa. Prior work separates the two by building detectors around simulated human--AI collaborations, such as human writing polished by AI or AI writing run through humanizers \citep{Bao2025DecouplingCA, li-etal-2026-beyond-final}. These simulations are expensive to create at scale, and detectors trained on them may not transfer to  real collaborations. In contrast, prose detectors like Pangram are trained on documents whose ideas and words come from the same source, which are cheap to collect: human text can be sampled from the pre-ChatGPT era, and AI text can be generated on demand.  

We propose \model, an idea provenance detector that learns from the same cheap labels as prose detectors and can therefore be trained at their scale. Our key methodological contribution is to change the classifier's \emph{input}. Following the cognitive process model of writing~\citep{FlowerHayes1981,
hayes2012}, we define a document's \emph{ideas} as what its author decided to say
(e.g., its claims, events, and organization), as opposed to
the words used to say it. We represent these ideas as an \emph{outline} whose items each consist of a discourse role paired with a short description of its content, paraphrased to remove the original wording (\autoref{fig:pipeline}).  We train \model\ on a dataset of 1M outlines extracted from FineWeb \citep{penedo2024the} documents, with silver labels from Pangram. As a control, we also train a separate model, \twinmodel, with the same backbone,
documents, and labels, but on the raw text instead of outlines. Despite identical supervision, the
two behave in opposite ways whenever a document's ideas and words come from different
sources, showing that the
outline forces \model\ to fit the labels through ideas rather than words.

\begin{itemize}[leftmargin=*,itemsep=1pt,topsep=2pt]
\item \textbf{\model\ detects human ideas in AI-generated documents.}  On a dataset of AI documents generated from increasingly detailed human plans, \model's flag rate falls from 94.9\%  with minimal human input to 6.8\% with a full human outline, while remaining above 96\% when plans come instead from AI.  \twinmodel\ and Pangram 4 flag 86.5\% and 92.4\% of documents generated from full human outlines. 
This behavior is further supported by results on 10 external benchmarks of AI polishing,
paraphrasing, and rewriting, where \model\ flags 5.6\% of documents on average,
versus 46.0\% for \twinmodel\ and 52.0\% for Pangram.

\item \textbf{\model\ detects AI ideas in human writing.} 
The opposite scenario is harder to study: prior work touches on it only through AI text that people or models edit. To this end, we collect the first dataset of long-form documents that people wrote from scratch following AI ideas. Our dataset contains 50 stories, each written from an outline extracted from an AI-generated story, without access to the raw text of that story.  \model\ flags 68.0\% of the human-written stories and 96.0\% of the AI originals, while \twinmodel\ and Pangram flag all of the AI original stories but only 0\% and 8.0\% of the human-written stories, respectively (\autoref{fig:teaser}).

\item \textbf{Despite not seeing the raw text, \model\ rivals the strongest open-source detector on conventional benchmarks and transfers across 24 languages.} On 19 evaluations that require separating fully AI (ideas and writing) from fully human documents, \model\ flags 91.1\% of AI documents on average with a false positive rate of 0.6\%, compared to 81.8\% and 0.2\% for EditLens-3B, the strongest existing open-weight detector. \twinmodel\ is even stronger here (98.9\% and 0.7\%), rivaling the commercial Pangram 4 at its published threshold (97.3\% and 0.3\%). Although trained only on English, \model\ transfers to 24
languages through English outlines, detecting 95.3\% of AI documents on average at a false positive rate of 0.1\% compared with 82.5\% and 0.0\% for
\twinmodel.

\item \textbf{\model\ flags ideas whose details could apply to anything or can't be
verified.} Based on an analysis of 90K item-level predictions, we find that \model\ scores ideas as
more AI when
they cite numbers or methods without the context needed to check them, or claim more
than their evidence supports. \model\ scores ideas as more human when they include specific sources that a reader could look up or discuss downsides and open problems.
\end{itemize}

We publicly release both \model\ and \twinmodel\ along with their training data, the evaluation datasets that we create, and our idea-level analysis.

%% file: sections/03_method.tex
\definecolor{roletxt}{RGB}{35, 79, 138}
\definecolor{contenttxt}{RGB}{176, 65, 48}
\definecolor{rolehl}{RGB}{221, 231, 245}
\definecolor{contenthl}{RGB}{246, 213, 212}

\begin{figure}[t]
\centering
\includegraphics[width=\textwidth]{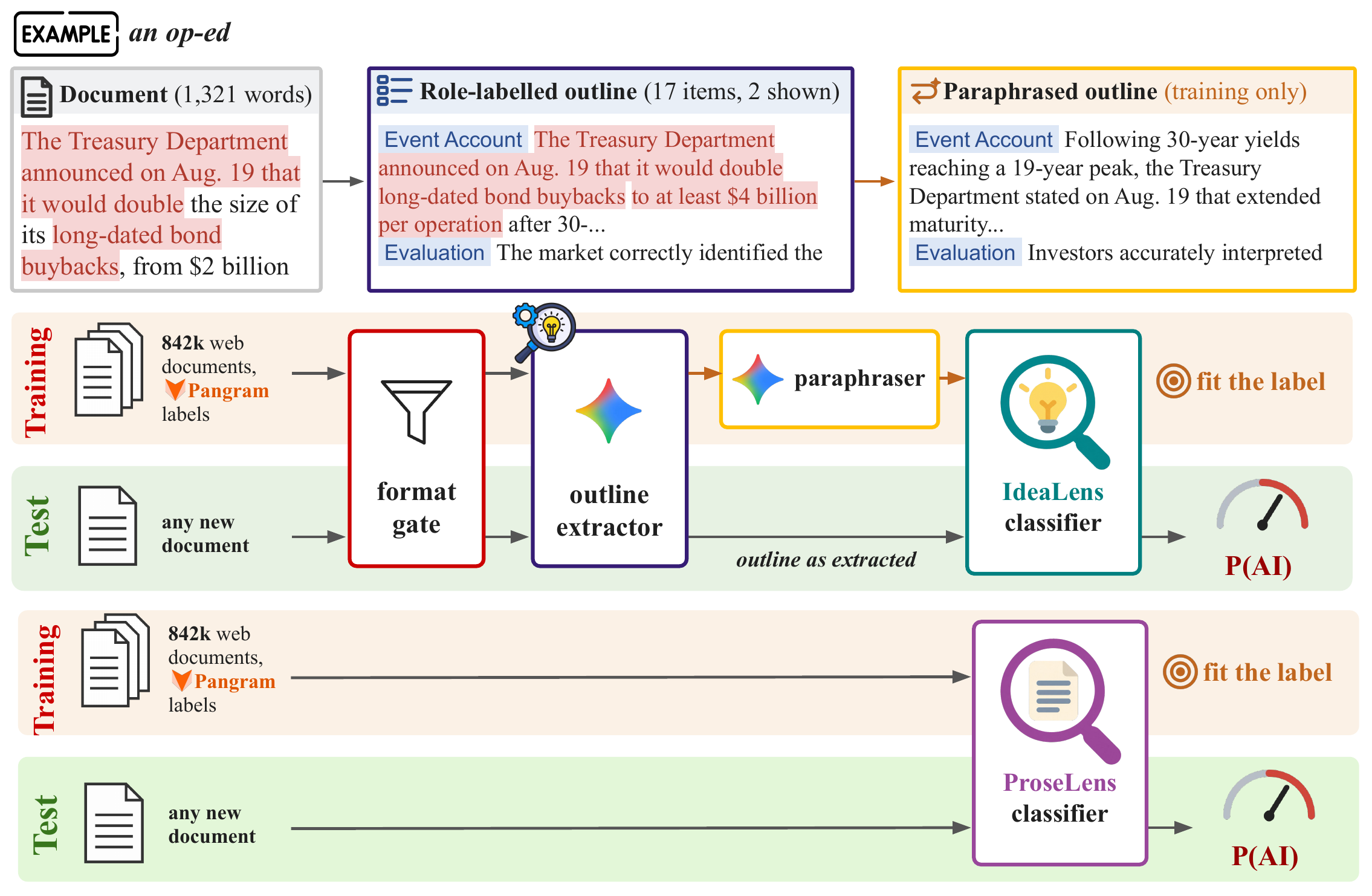}
\caption{\textbf{\model\ sees an outline of each document, not the whole text.}
Documents are assigned a format and reduced to an outline, each labeled with 
{\setlength{\fboxsep}{1pt}%
 \colorbox{rolehl}{\textcolor{roletxt}{one of the format's roles}}}. This example shows two of the 17 outline items extracted from an op-ed.
For training, the outlines are paraphrased to remove 
{\setlength{\fboxsep}{1pt}%
 \colorbox{contenthl}{\textcolor{contenttxt}{wording copied from original documents}}}. \model\ learns the Pangram label from the paraphrased outline during training, and extracted outlines are provided at inference.
}
\label{fig:pipeline}
\end{figure}

\section{Learning Idea Provenance from Outlines}
\label{sec:method}

The fundamental challenge in training an idea detector is that we have no large-scale labels for who came up with a document's ideas, only for who wrote its prose. Thus, we train on prose labels while replacing each document with an outline that preserves its ideas but not its wording. To create these outlines (\autoref{fig:pipeline}), we  identify discourse roles typical of each writing format, use them to extract an outline of each document, and paraphrase the outline to remove wording copied from the raw document. We train \model\ to predict each document's Pangram label from its paraphrased outline, while  \twinmodel\ predicts the same labels from the raw document.

\paragraph{Web documents with silver labels from Pangram.} We sample a 1M subset from WildAI \citep{russell2026aitoken}, which labels documents from FineWeb \citep{penedo2024the} with WebOrganizer \citep{wettig2025weborganizer} formats and topics along with authorship labels from Pangram 3.3.2 \citep{thai2026editlens}.\footnote{We note that Pangram 3.3.2 only provides training labels, while our evaluations benchmark \model\ and \twinmodel\ against the more recent Pangram 4.} We focus on long-form writing, selecting English documents of at least 500 words that belong to one of eight relevant WebOrganizer formats (\autoref{tab:corpus}), and that Pangram labels as either human-written or AI-generated. We assume that the ideas and prose in these documents share the same source, which enables us to use them as noisy \emph{silver} labels for idea provenance. We discard documents that Pangram labels as \emph{mixed}, since AI only wrote part of these texts and thus idea provenance is unclear. We split WildOutlines, the full corpus of 1M documents into training ($\sim$842k), validation ($\sim$30k), calibration ($\sim$80k, human documents only)\footnote{The calibration split is used to set a threshold for each detector, which turns its predicted probabilities into binary decisions (AI vs. human). Because the calibration split only contains human-written documents, we can set the threshold to achieve a target false positive rate (FPR). Unless explicitly stated otherwise, we report detector accuracy at thresholds calibrated for an FPR of 1\%.} and test ($\sim$50k) sets (\autoref{tab:corpus}).\footnote{Building the corpus cost about \$66.4k (\S\ref{app:cost}).} 

\paragraph{Outlines of discourse roles.}
Cognitive models of writing~\citep{FlowerHayes1981, hayes2012} separate the act of \emph{planning} from the act of \emph{translating} these plans into prose. To build a document representation of the former without the latter, we turn to discourse theory, which describes the roles a passage plays within the broader context of the document, for example through rhetorical structure theory~\citep{MANNTHOMPSON+1988+243+281,Mann2006}. We represent a document as an \emph{outline}: an ordered list of discourse roles, each paired with brief descriptions of how those roles are implemented in the document. For example, the op-ed displayed in \autoref{fig:pipeline} contains 17 items, starting with an \emph{Event Account} of a Treasury announcement followed by an \emph{Evaluation} of the market's reaction. 
Many of these roles are format-specific~\citep{swales1990genre, SunWangXiong+2021+599+624, van2019macrostructures}; for instance, creative writing uses different roles than nonfiction. Thus, we induce separate role sets for each of the eight formats in our dataset (\autoref{tab:roles_example}). 

\paragraph{Automatically extracting outlines.}
We build a training dataset, \corpus, for \model\ by converting documents to outlines in three automated steps:

\begin{enumerate}[label=(\roman*), wide=0pt, itemsep=1pt, topsep=2pt, parsep=0pt, partopsep=0pt]
    \item \textbf{Role induction:} We adapt the TopicGPT framework~\citep{pham2024topicgptpromptbasedtopicmodeling} by prompting GPT-5.6 Sol \citep{openai2026gpt56}
    to propose base roles for each format given a detailed description of that format. Then, we label outlines of 200 documents in that format using those roles, while asking the model to invent new roles as needed. Afterwards, we deduplicate and prune the resulting role set (e.g., to 42 roles for nonfiction); more details in \S\ref{app:roles}.
    \item \textbf{Outline extraction:} Given a document, we prompt Gemini 3.7 Flash and Gemini 3.1 Pro \citep{gemini2023,gemini37flash,gemini31pro} with its format's role set (and few-shot demonstrations from the much more powerful Claude Fable 5 \citep{anthropic2026fable5}) to produce an outline of the document's ideas in order, where each item consists of a role and a short content description. 
    \item \textbf{Paraphrasing:} Extracted outlines can copy phrasing from the original document, which a model could use as a stylistic shortcut to the label (i.e., human wording can leak into the outlines). Therefore, we prompt Gemini 3.1 Pro \citep{gemini31pro} to paraphrase each outline. This pushes all outlines toward a homogeneous, AI-like style: for human-written documents, the fraction of outline text that Pangram 4 labels as human drops from 8.5\% to 0.3\% (\autoref{tab:probe}). Meaning is still preserved (\autoref{tab:probe_sim:0}), as paraphrased outlines have a cosine similarity of 0.93 with the originally extracted outlines (vs. 0.39 for unrelated outlines) and 0.76 with the source document (vs. 0.26). We paraphrase only for training, to keep the model from learning style; at inference, \model\ sees the unparaphrased outline.
\end{enumerate}

\paragraph{Training \model\ and \twinmodel.}
Both of our models start from a backbone of Nemotron-3.5-Lightning-30B-A3B \citep{nvidia2026nemotron35lightning}, fine-tuned for one epoch with LoRA~\citep{hu2022lora} using Tinker \citep{tml2026tinker}.\footnote{Details and results for other backbones are in \S\ref{app:configs}.} We threshold each model's probability that a document is AI-generated at a 1\% false positive rate on our calibration split. The only difference between the two models is that \model\ is trained on  paraphrased outlines, whereas \twinmodel\ is trained on the original documents. Since they share their backbone, training documents, and labels, any difference in their behavior is a direct result of transforming the input into our outline representation.

%% file: sections/04_experiments.tex
\section{Experimental setup}
\label{sec:experiments}

We evaluate \model\ and \twinmodel\ against 17 existing detectors, including the state-of-the-art Pangram 4, on 19 external benchmarks and three datasets that we create ourselves: \collabLadder, \xLadder, and \twicetold. 
These datasets cover all four quadrants in \autoref{fig:teaser}: documents with \emph{shared} provenance (fully human or fully AI) as well as documents with \emph{mixed} provenance (human ideas and AI prose, or AI ideas and human prose). Idea detectors should achieve high accuracy in all four quadrants, while prose detectors should fail in the two mixed quadrants. 

\paragraph{Evaluation metrics and baselines.} We report \textbf{accuracy} at detecting AI ideas, which we define as the percentage of documents whose predicted label (AI or human) agrees with the source of their ideas. 
Our baselines include Pangram 4~\citep{glickenhaus2026pangram4technicalreport}, EditLens-Llama-3.2-3B \citep{thai2026editlens}, Fast-DetectGPT \citep{bao2024fastdetectgpt}, Binoculars \citep{hans2024spotting}, MELD \citep{li2026meldmultitaskequilibratedlearning} and Desklib \citep{desklib2025aitextdetector}.  For \model, \twinmodel, and EditLens, we set a global threshold across all formats at 1\% FPR on our calibration set. We binarize Pangram's ternary labels (AI, AI-assisted, human) following its tech report, and other baselines use their default thresholds.\footnote{Details on thresholding, including the Pangram conversion, are in \S\ref{app:thresholds}.}%

\paragraph{Existing benchmarks.} Of the 19 external detection benchmarks (see \S\ref{app:benchmarks} for full list), 14 contain documents with shared provenance (fully human or fully AI). We evaluate accuracy on these documents and on our in-domain test set of 50K documents from our Pangram-labeled web corpus. We also measure FPR on a set of 10K documents from C4~\citep{c4dataset}, which was created prior to ChatGPT's release and is thus much more likely to be human-written than our silver-labeled test set. Finally, 10 external benchmarks also contain documents written by humans but edited by an AI (e.g., via paraphrasing, polishing, or rewriting), which we use as instances of human ideas rendered in AI prose; this includes OpAI-Bench~\citep{Bsharat2026OperationGuidedPH} and GEDE~\citep{Gehring2025LimitsOL}.

\paragraph{\collabLadder: AI writing from increasingly detailed human plans.} 
A common collaborative writing pattern is for a human to give an AI their ideas via a prompt, which it uses to write a document from scratch \citep{10.1145/3698061.3726910}. 
Only two external datasets we evaluate on contain instances of this pattern~\citep{saha2026policies,Gehring2025LimitsOL}, and neither  controls the \emph{amount} of human input, so it is unclear how much of the final document's ideas came from the human. Thus, we create \collabLadder\ from 500 seed documents (both human-written and AI-generated) sampled from our in-domain test set. For each seed document,  GPT-5.6 Sol~\citep{openai2026gpt56} writes five progressively more detailed prompts: the level-0 prompt contains only the topic (e.g., ``Write a dining and lifestyle blog post reviewing a restaurant visit.''), while  the level-5 prompt contains the full outline extracted as per \S\ref{sec:method} with instructions to add nothing else to it.  GPT-5.6 models then generate a new document for each prompt. Since all of the prose in \collabLadder\ is AI-generated, prose detectors should flag every document as AI, and an idea detector should do the same for AI seeds. However, on documents generated from human seeds, idea detectors should flag fewer and fewer documents as the prompts carry more of the human plan.

\paragraph{\twicetold: humans writing from AI plans.} 
We are also interested in the reverse pattern: a human obtains ideas from an AI and then writes the document themselves. Existing datasets approximate this pattern with AI-generated text that humans edit or AI tools humanize,\footnote{E.g., LAMP \citep{LAMP}, HART \citep{Bao2025DecouplingCA}, and \citet{baumler-etal-2026-make}.} but these edited documents are still largely AI-generated and thus do not cleanly separate idea and prose provenance. To fill this gap, we create \twicetold,\footnote{Named after \emph{Twice-Told Tales}, a  collection of previously published short stories by Nathaniel Hawthorne.} a dataset with 50 stories written \emph{from scratch} by humans given only the outline of a 500-word story generated by GPT-5.6 Sol (see \S\ref{app:aiip} for more details).\footnote{The stories were written in person by 17 computer science graduate students or faculty. We avoided crowdsourcing to reduce the chance of AI use during the writing process \citep{10.1145/3685527}.} Prose detectors should label these stories as human, while idea detectors should label them as AI. We also include the 50 original AI stories, which all detectors should label as AI.

\paragraph{\xLadder: idea detection in multilingual documents.} 
While prose detectors rely on language-specific features (e.g., vocabulary choice, sentence structure), \model\ uses an LLM extractor that can generate \emph{English} outlines for documents in any language. Thus, even though \model\ is trained only on English documents, we can apply it to documents in other languages without any additional training. To test this, we create \xLadder\ by sampling 10K human-authored documents in 24 languages (\autoref{tab:multiling}) from FineWeb2 \citep{penedo2025fineweb}, to measure FPR across different languages. For each, we follow the \collabLadder\ protocol to generate AI versions in the original language at level 0 (AI ideas) and level 5 (human ideas).

%% file: sections/05_results.tex
\section{\model detects AI ideas, not AI prose}
\label{sec:results}

\begin{figure}[!t]
\centering
\includegraphics[width=\textwidth]{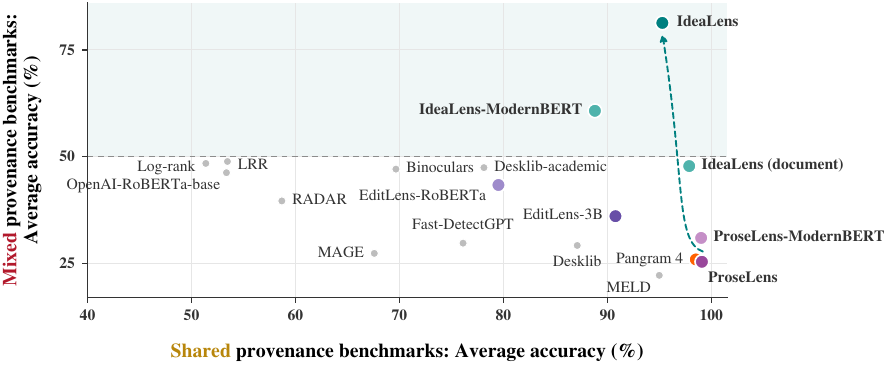}
\caption{\textbf{Outline-trained \model detects idea authorship while traditional detectors track prose authorship.}
The average accuracy across all benchmarks where ideas and prose are shared by one author (x-axis) and where they are by different authors (y-axis). \model\ is the only detector accurate with both shared (95.3\%) and mixed (81.3\%) authorship.\protect\footnotemark
}
\label{fig:quadrants}
\end{figure}

\autoref{fig:quadrants} summarizes our main results across 51 evaluation splits from 22 benchmarks, where each point is a detector plotted by its aggregate idea detection accuracy on documents with shared vs. mixed provenance. 
\model\ is the only detector that achieves high accuracy in both shared (95.3\%) and mixed (81.3\%) settings. \twinmodel\ and Pangram 4 are extremely accurate on documents with shared provenance (99.1\% and 98.5\%, respectively) but fail to identify the source of a document's ideas in mixed settings (25.4\% and 25.9\%). 
Since \twinmodel\ has the same backbone, training documents, and training labels as \model, this highlights the impact of our outline representation, which a ModernBERT-based pair \citep{warner-etal-2025-smarter} replicates with a different backbone. \model\ also transfers unexpectedly well to \emph{raw text} inputs at test time, exhibiting high shared provenance accuracy and substantially higher mixed provenance accuracy than \twinmodel.

\footnotetext{With weighted average accuracy, \model\ is more accurate at 92.6\% compared to \twinmodel's 46.8\% and Pangram's 41.1\% on mixed provenance benchmarks.}

\begin{figure}[t]
    \centering
    \begin{subfigure}[b]{0.43\textwidth}
        \centering
        \includegraphics[width=\textwidth]{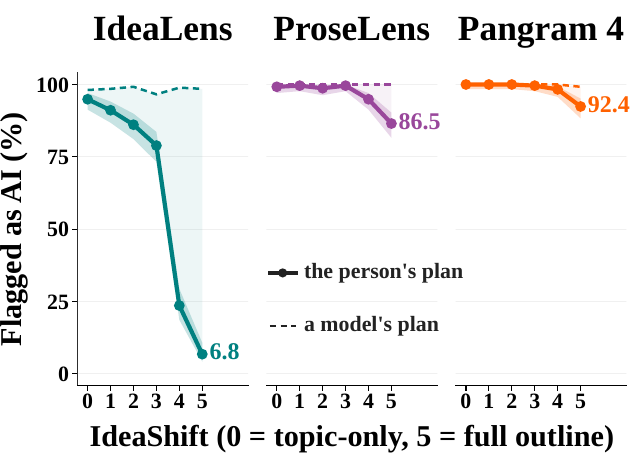}
        \caption{\collabLadder}
        \label{fig:ladder_attacks_a}
    \end{subfigure}
    \hfill
    \begin{subfigure}[b]{0.53\textwidth}
        \centering
        \includegraphics[width=\textwidth]{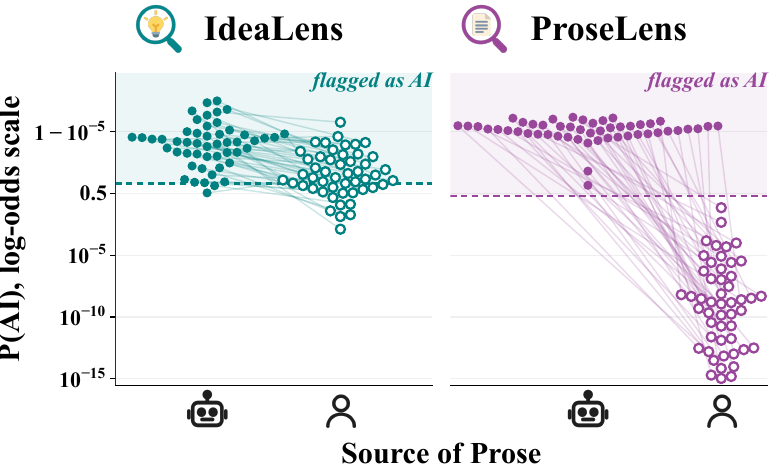}
        \caption{\twicetold}
        \label{fig:ladder_attacks_b}
    \end{subfigure}
    \caption{\textbf{\model\ detects varying levels of human input in AI-generated text (left) and assigns similar P(AI) for \twicetold and its AI-versions (right).} (a) On \collabLadder, \model\ is the only model whose flag rate tracks with varying levels of human input in the AI-generated text; on the AI control, all models flag nearly entirely at all levels. (b) The difference in P(AI) between the AI and human-written stories in \twicetold is much smaller for \model than for \twinmodel.}
\label{fig:ladder_attacks}
\end{figure}

\paragraph{\model\ detects idea provenance on \collabLadder.} As the human plan in the prompts from \collabLadder\ becomes more detailed, \model's flag rate drops from 94.9\% at level 0 to 6.8\% at level 5 (\autoref{fig:ladder_attacks_a}). In contrast, \twinmodel\ and Pangram 4 flag 86--100\% of documents regardless of the level of human ideation. When the plans come from AI instead of humans, \model\ consistently flags more than  96\% of documents at every level (\autoref{tab:ladder_a:0}), showing that it is sensitive to idea provenance rather than level of prompt detail. Finally, we observe that the biggest drop in flag rate occurs between levels 3 and 4, which is when the prompts begin including outlines with \emph{content} in addition to structural role information (\autoref{tab:ladder_h:0}).

\paragraph{\model\ is insensitive to AI rewording of human ideas.}
Aggregated over 12 benchmarks
where AI makes surface-level changes to human text (e.g., polishing, paraphrasing), \model\ labels 94.7\% of documents as human, compared to 50.7\% for \twinmodel\ and 43.8\% for Pangram 4 (\autoref{tab:q4}). In GEDE \citep{Gehring2025LimitsOL}, \model\ flags only 0.5\% of human essays as AI after minor corrections made by an LLM (vs Pangram 4's 54.6\% and \twinmodel's 45.8\%) and 2.6\% after a rewrite preserving the original content (vs Pangram 4's 94.3\% and \twinmodel's 77.1\%). In HART \citep{Bao2025DecouplingCA}, \model\ flags 4.9\% of LLM paraphrases of human documents, compared to 42.8\% by Pangram 4 and 44.5\% by \twinmodel. In the Academic Integrity dataset \citep{Karr2026WhyAD}, it flags 8.2\% of abstracts that an LLM generated from a human-written reference abstract, compared to Pangram 4's 96.4\% and \twinmodel's 83.9\%.

\paragraph{\model\ detects AI ideas in \twicetold's human-written prose.}
\model\ flags 68\% (95\% CI 54--79\%) of \twicetold\ stories as AI, compared to 0\% for \twinmodel\ and 8\% for Pangram 4.\footnote{Pangram 4's flags are likely due to usage of character names typical of AI writing (e.g., \emph{Mara}, \emph{Bellweather}).} All detectors are reliable at flagging the AI-generated original stories (96\% for \model\ and 100\% for \twinmodel\ and Pangram), but only \model\ sticks with its prediction for the majority of the human stories, even though the prose is completely different (\autoref{fig:ladder_attacks_b}). The stories that \model marks as human generally follow their outlines while adding new events or world-building details: 12 out of 16 errors are on stories that contain speculative or fantasy elements. %

\paragraph{From ideas alone, \model\ outperforms the strongest existing open detectors.} On 19 evaluation splits with shared provenance, \model\ detects 91.1\% of AI documents at 0.6\% FPR compared with 81.8\% at 0.2\% FPR for EditLens-3B,\footnote{While \model's 30B backbone is larger than EditLens-3B, it is a mixture-of-experts model with only 3B active parameters. Our \model-ModernBERT variant, with only 0.4B parameters, detects 79.1\% of AI documents at an FPR of 1.5\%.} the strongest existing open baseline (\autoref{tab:cons}), with both calibrated to a 1\% target. Training on full text does better: \twinmodel\ reaches 98.9\% at 0.7\% FPR, rivaling Pangram 4 (97.3\% at 0.3\%) and making it the strongest open detector in our evaluations. 
To cleanly measure FPR, we collect 10K pre-ChatGPT documents from C4: \model\ flags only 0.01\% compared to 0\% for \twinmodel\ and Pangram 4 (\S\ref{app:c4}). 
Against humanization and adversarial attacks, \model's detection rate drops 6.4 points on average, between Pangram's 4.9 and \twinmodel's 10.6 (\autoref{tab:attacks_app}). One weakness of \model\ is DIPPER paraphrasing \citep{krishna2023paraphrasing} in GEDE \citep{Gehring2025LimitsOL}, where it detects 77.6\% versus 97.1\% for \twinmodel\ and 95.1\% for Pangram 4; we suspect this attack changes content as well as style (\S\ref{app:attack_similarity}).

\paragraph{\model\ detects idea provenance across 24 languages on \xLadder.}
\xLadder\ contains documents generated in 24 different languages, and  \model\ flags 95.3\% of the level-0 documents but only 0.7\% of the level-5 documents despite being trained only on English. These results show that signatures of AI ideas are largely invariant to the language in which they are written. In contrast, prose detectors like 
\twinmodel\ and Pangram 4 flag 82--85\% of level-0 and 45--62\% of level-5 documents, and this gap between \model\ and these detectors widens in low-resource languages like Tamil and Amharic. 
Additionally, \model\ achieves a low FPR on non-English human-written documents:
on 10K pre-ChatGPT documents in these 24 languages sampled from FineWeb2 \citep{penedo2025fineweb}, it flags only 0.1\% as AI (\autoref{tab:multiling}).

\input{tables/feature/diff_examples} %
\paragraph{\model\ transfers unexpectedly well to raw text at test time.} While trained on outlines, \model\ also works on raw documents during inference. On shared provenance, it detects 96.1\% of AI documents at a 0.3\% realized FPR, nearly matching \twinmodel\ and Pangram 4. Thus, features learned from outlines are also present in documents and even augmented with additional details not present in the outline. On the 12 benchmarks with human ideas and AI prose, \model\ given raw text still labels 89.6\% of   documents as human, compared to 94.7\% for \model\ with outlines and 50.7\% for \twinmodel (\autoref{tab:q4}). However, it struggles with \twicetold's AI ideas and human writing, detecting only 6\% of the stories as AI. We hypothesize that the raw stories expose more of the human author's added details, which \model\ associates with human ideas.

%% file: tables/feature/diff_examples.tex
\providecommand{\humanicon}[1][teal]{\tikz[baseline=-0.35ex]{%
  \fill[#1] (0,0.55ex) circle (0.42ex);
  \fill[#1] (-0.75ex,-0.75ex) arc (180:0:0.75ex) -- cycle;}}
\providecommand{\aiiconcore}[2]{\tikz[baseline=-0.35ex]{%
  \fill[#1, rounded corners=0.28ex] (-0.78ex,-0.72ex) rectangle (0.78ex,0.62ex);
  \fill[#2] (-0.34ex,-0.02ex) circle (0.17ex) (0.34ex,-0.02ex) circle (0.17ex);
  \draw[#1, line width=0.13ex] (0,0.62ex) -- (0,0.98ex);
  \fill[#1] (0,1.08ex) circle (0.16ex);}}
\providecommand{\aiicon}[1][purple]{\aiiconcore{#1}{white}}
\providecommand{\aiiconwhite}{\aiiconcore{white}{purple}}
\providecommand{\diffnum}[1]{\tikz[baseline=(n.base)]{%
  \node[circle, fill=black!70, text=white, inner sep=0.5pt, minimum size=1.9ex,
        font=\bfseries\tiny] (n) {#1};}}
\providecommand{\humanul}{\bgroup\markoverwith{\textcolor{teal}{\rule[-0.55ex]{2pt}{0.7pt}}}\ULon}
\providecommand{\aiul}{\bgroup\markoverwith{\textcolor{purple}{\rule[-0.55ex]{2pt}{0.7pt}}}\ULon}
\providecommand{\humanpoint}[1]{\textcolor{teal}{\humanul{#1}}}
\providecommand{\aipoint}[1]{\textcolor{purple}{\aiul{#1}}}
\providecommand{\difflabel}[4]{%
  \diffnum{#1}~\textbf{#2}\par\vspace{3pt}
  {\color{black!75}\humanicon~#3\par\vspace{2pt}\aiicon~#4}}

\begin{table*}[t]
  \centering
  \caption{\textbf{Examples of items for four core differences between  human and AI ideas.}
    The first column describes each contrast and summarizes how it is realized in
    \textcolor{teal}{\textbf{human}} and \textcolor{purple}{\textbf{AI}} ideas.}
  \label{tab:feat-diff-examples}
  \scriptsize
  \setlength{\tabcolsep}{5pt}
  \renewcommand{\arraystretch}{1}
  \newlength{\celltop}\newlength{\cellbot}
  \setlength{\celltop}{2.5ex}\setlength{\cellbot}{0.9ex}%
  \newcolumntype{L}[1]{>{\raggedright\arraybackslash\rule{0pt}{\celltop}}p{#1}<{\rule[-\cellbot]{0pt}{0pt}}}
  \newcolumntype{Y}{>{\raggedright\arraybackslash\rule{0pt}{\celltop}}X<{\rule[-\cellbot]{0pt}{0pt}}}
  \setlength{\aboverulesep}{0pt}\setlength{\belowrulesep}{0pt}
  \begin{tabularx}{\textwidth}{L{0.28\textwidth}YY}
    \rowcolor{black!80}
    \footnotesize\textcolor{white}{\textsc{Difference}} &
    \cellcolor{teal}\footnotesize\textcolor{white}{\humanicon[white]~\textsc{Human idea}} &
    \cellcolor{purple}\footnotesize\textcolor{white}{\aiiconwhite~\textsc{AI idea}} \\
    \noalign{\global\setlength{\celltop}{2.8ex}\global\setlength{\cellbot}{1.4ex}}%

    \difflabel{1}{Decisive vs.\ Universal Details}
      {Details could alter the author's point or conclusion.}
      {Details could apply to any similar subject.} &
    The bottom plate of the Sony Cyber-shot RX100 contains a mini-HDMI output alongside a shared compartment for the power cell and SD/Memory Stick media---\humanpoint{a hatch that becomes inaccessible during tripod use}. (\ldots) &
    Physical attributes of the Jony Ive speaker include \aipoint{a resilient, pared-down aesthetic, premium audio fidelity, low power consumption}, and control mechanisms driven by straightforward tactile and vocal inputs. \\

    \rowcolor{black!5}
    \difflabel{2}{Functional vs.\ Apparent Specificity}
      {Numbers have comparisons, and instructions can be followed.}
      {Methods or numbers look specific without being usable.} &
    Over a period of merely ten days, the shop generates 700 yuan in profit from selling athletic gear. This amount exceeds \humanpoint{twice the highest anticipated expense for the prizes}, demonstrating that (\ldots) &
    Development teams are advised to handle technical obstacles preemptively by engaging in \aipoint{thorough initial preparation, continuous education, risk reduction strategies, Agile methodologies}, strict quality testing, and discussions with domain specialists. \\

    \difflabel{3}{Bounded vs.\ Overextended Claims}
      {Claims state clear limits on when they apply.}
      {Claims extend beyond evidence to uncertain futures.} &
    The initial deployment of the new fleet in passenger service is \humanpoint{anticipated for mid-2022}, originating from the Montreal Maintenance Centre. (\ldots) &
    The 2025 Kia K4 is projected to \aipoint{establish a permanent legacy} as an updated standard for the category, demonstrating that (\ldots) \\

    \rowcolor{black!5}
    \difflabel{4}{Preserved vs.\ Erased Costs}
      {Existing costs, limitations, and unresolved problems are retained.}
      {Costs vanish or become ideal trade-offs.} &
    The writer intends to work toward the Schutzhund 1 certification for Quasi during the upcoming summer, limiting advancement to that initial tier \humanpoint{due to the logistical challenge of a 250-mile journey} to the closest training facility, (\ldots) &
    The overall assessment concludes that the M3 MacBook Air represents the definitive lightweight notebook, providing \aipoint{an ideal equilibrium} among computational strength, electrical endurance, structural excellence, and operating system cohesion. \\
    \bottomrule
  \end{tabularx}
\end{table*}

%% file: sections/06_analysis.tex
\section{What separates AI ideas from human ideas?}
\label{sec:feat_analysis}

Human writing can differ from AI writing both in the ideas that they express and in how those ideas are organized. To disentangle the two, we first conduct an ablation (\autoref{tab:ablations})  that compares \model's AUROC on our in-domain test set given outlines with role labels removed (0.993), shuffled outlines (0.993), and role labels alone (0.601), showing that \model\ relies primarily on content. To characterize this content, we follow the methodology of~\citet{russell2026storyscope} to extract 79 interpretable features\footnote{Examples of features are provided in \autoref{tab:feat-list-human} and \autoref{tab:feat-list-ai}.} from 90K item-level predictions, most of which are predictive of \model's scores on held-out items.\footnote{On 13,616 held-out items, 64 of the 79 features align with \model's scores in their proposed direction (AI vs. human), while only one shows a conflict. Detailed analysis is presented in \S\ref{app:feat-verification}.} \autoref{tab:feat-diff-examples} synthesizes these features into a short list of systematic differences that \model\ uses to separate human and AI ideas.

\paragraph{AI ideas use details that could apply to anything.}
Human ideas contain details that are crucial to the author's point: the human camera review in the first row of \autoref{tab:feat-diff-examples} points out a problem during tripod use, while an AI speaker review mentions ``premium audio fidelity'' that could apply to many other speakers. 
Going beyond such vagueness, which has also been noted in prior work~\citep{kim2026argumentcollapsellmsflatten,russell2026storyscope}, we find that AI ideas can also \emph{look} specific without being usable: the AI idea in the second row advises development teams to handle obstacles via ``risk reduction strategies'' and ``Agile methodologies'' but never says how to carry these out, whereas the human counterpart reveals specifically which baseline amount is more than doubled.

\paragraph{AI ideas overstate claims and hide downsides.}
Human ideas generally state the limitations of their claims, while AI ideas extend their claims beyond the evidence: the AI idea in the third row of \autoref{tab:feat-diff-examples} predicts that a newly released Kia model will ``establish a permanent legacy'', whereas the human idea provides a hedged projection of when a fleet deployment will occur. AI ideas also omit limitations: the AI review in the fourth row describes the MacBook as at an ``ideal equilibrium'' without any downsides, while human writers explicitly state the limitations that constrain them.\footnote{The alignment and scores of features supporting each core difference are presented in \autoref{tab:feat-diff-summary}.}

%% file: sections/07_related_work.tex
\section{Related work}
\label{sec:related}

\paragraph{Detecting AI and mixed authorship.}
Standard AI text detectors \citep{mitchell2023detectgpt,hu2023radar,hans2024spotting,emi2024pangram}, including several of our baselines, perform binary classification of prose authorship. Existing work on mixed authorship settings either focuses on AI-edited text, including polishing and paraphrasing~\citep{zhang-etal-2024-llm,ta-etal-2026-faid,saha-feizi-2025-almost,Bsharat2026OperationGuidedPH} or attempts to explicitly separate content from expression~\citep{Bao2025DecouplingCA,chen2026d2crouting,chen-etal-2026-coconuts,peerprism}. Other work aims to log the writing process~\citep{krebs2026fai2rdidwhatchecked, siddiqui2026draftmarks} and theoretically examines how provenance should be defined~\citep{schnell2026wrotethisturingtotal}.
In contrast, \model\ learns idea provenance from prose-level labels, and we ground our formalization of provenance  in cognitive theories of writing \citep{FlowerHayes1981,hayes2012} rather than in the text itself or formal semantics \citep{10.3115/990403.990490}.

\paragraph{Learning from transformed inputs.}
Transforming training inputs can change which features a model learns without changing its supervision. \citet{geirhos2019texture} suppress texture cues through image stylization, while \citet{ilyas2019adversarial} show that training on a dataset constructed to emphasize robust features yields adversarial robustness.~\citet{krishna-etal-2020-reformulating} use paraphrasing to normalize stylistic variation, enabling style transfer through a content-preserving intermediate representation, and \citet{stamatatos-2017-authorship} masks content words so that authorship classifiers must rely on style. We apply the same principle in the opposite direction by keeping the labels of a prose detector but replacing its inputs with paraphrased outlines, so that the labels can only be fit through the ideas.

%% file: sections/09_conclusion.tex
\section{Conclusion}
\label{sec:conclusion}

We show that AI writing and human writing can be meaningfully distinguished from one another through their ideas alone. Using a corpus of 1M outlines, we train \model\ on silver labels from a prose detector and benchmark it across 22 datasets. \model\ tracks idea provenance when it differs from prose provenance, unlike a control model (\twinmodel) that is trained on the same labels but raw text instead of outlines. \model\ transfers to 24 languages despite being trained only on English, and it outperforms the strongest existing open detector on documents where provenance is shared. \model\ has limitations (\S\ref{app:limitations}): its training labels noisily approximate idea provenance, its binary formulation does not cleanly handle documents whose ideas came from both humans and AI, it is weaker on short texts and some adversarial perturbations, its outline extraction process relies on proprietary models, and our \twicetold\ dataset covers only 50 stories. We release our models and datasets to support future research on idea provenance detection.

%% file: sections/10_statements.tex
\section*{Acknowledgements}
We extend our special gratitude to Chau Pham, Yapei Chang, Yekyung Kim, Shufan Wang, Atrey Desai, Vinay Samuel, Chuanhai Xu, Sergiusz Rzepkowski, Piotr Nonas, Adam Podoxin, and Adrian Chang for participating in dataset creation. We also thank the University of Maryland Computational Linguistics and Information Processing
(CLIP) Lab for their feedback and support. This project was partially supported by awards IIS-2046248, IIS-2553892, and IIS-2312949 from the National Science Foundation (NSF). We thank Tinker and Pangram for a research credit award and Google for a Cloud Credit award that enabled this research.

\subsection*{AI use statement}

We used AI tools for the following tasks: 
\begin{itemize}
    \item \textbf{Prompt engineering:} We used coding agents to aid in constructing prompts for various tasks grounded in descriptions and principles we laid down. 
    \item \textbf{Dataset construction:} We prompt a frontier LLM to automatically extract outlines from documents that we then use to train our models. We also prompt an LLM to construct some synthetic evaluation data. 
    \item \textbf{Design or provide feedback on research methodology or experiments:} Coding agents were asked to provide feedback many times throughout the project, as well as help design and run the training of all models. 
    \item \textbf{Implement methods:} Coding agents implemented most of the experimental code.
    \item \textbf{Support qualitative and thematic data analysis:} We often asked coding agents to create tables and figures of results so we could interpret the results. Many of these figures and tables ended up in the final draft. 
\end{itemize}

Additional disclosures include the use of AI for the following purposes: create or modify scientific figures or images, create or edit software code,  creation of artifacts, brainstorming, sourcing/searching for information, edit a research paper to improve readability, and identify relevant literature. 

We have reviewed and validated all AI-assisted work and take responsibility for the final content of the work, including text, claims, and artifacts produced with the aid of AI.

\subsection*{Ethics statement}
Our work is a methodological contribution and is not intended to be used for invalidating human or AI-assisted writing. All predictions made by \model\ and \twinmodel\ are statistical estimates and must not be used by themselves as conclusive evidence to establish AI use in writing. All datasets and non-commercial models are publicly available and the original license permissions were followed. This work includes a human-curated dataset, but participants provided consent and no sensitive information was asked.

\subsection*{Reproducibility statement}

We release all data and trained models on Hugging Face (\url{https://huggingface.co/collections/rishanthrajendhran/idealens-6abee785ce6196fc0be9200f}). The code to train and evaluate \model\ and \twinmodel\ is released on GitHub (\url{https://github.com/RishanthRajendhran/IdeaLens}), along with a PyPI package (\url{https://pypi.org/project/idealens/}) that simplifies inference with our trained models. We also publish \model\ and \twinmodel\ on Tinker (\url{https://thinkingmachines.ai/tinker/}) which makes it possible to test our models without GPUs.

%% file: sections/appendix.tex
\renewcommand{\topfraction}{0.95}
\renewcommand{\bottomfraction}{0.9}
\renewcommand{\textfraction}{0.05}
\renewcommand{\floatpagefraction}{0.85}
\setcounter{topnumber}{4}
\setcounter{bottomnumber}{3}
\setcounter{totalnumber}{6}
\etocdepthtag.toc{appendixpart}
\etocsettagdepth{mainpaper}{none}
\etocsettagdepth{appendixpart}{subsection}
\etocsettocstyle{\section*{Appendix contents}}{}
\tableofcontents

\noindent Every table of every evaluation (all arms, every operating point, per-format cuts, the other model families and the
length splits), all prompts and role vocabularies, and the code that produced the numbers are in our GitHub repository: \url{https://github.com/RishanthRajendhran/IdeaLens/tree/main/paper}.

\input{sections/08_limitations}

\section{Labels and quadrants}
\label{app:labels}
We label each document by who conceived its ideas, not who wrote its prose: a document with a person's ideas is human however much of its prose a model wrote, and vice versa. Crossing the two axes gives four quadrants:
\begin{itemize}[leftmargin=*,itemsep=1pt,topsep=2pt]
\item \textbf{Shared authorship}, as in conventional detection benchmarks: \textbf{HH} (Q1), human ideas and human prose, and \textbf{AA} (Q3), model ideas and model prose;
\item \textbf{Mixed authorship}: \textbf{HA} (Q4), human ideas in model prose (e.g., a model polishing a person's text), and \textbf{AH} (Q2), model ideas in human prose (e.g., a person writing up a model's ideas).
\end{itemize}
Accuracy is the share of documents whose predicted label matches the source of their ideas: TPR when the ideas are a model's and $100-\text{FPR}$ when they are a person's. Arms with genuinely shared ideas, such as \collabLadder{} levels 2--4 or a model imitating an author from a few passages, have no correct binary label, so we report their mean P(AI) instead (\S\ref{app:mixed}). \autoref{tab:labels} maps every arm to its label.

\section{Benchmarks}
\label{app:benchmarks}
We evaluate on 19 external benchmarks and on four sets we build or sample ourselves (\collabLadder{}, \xLadder{}, \twicetold{} and a C4 sample). We use every external benchmark as released, with three exceptions: we keep the English rows of DetectRL-X and HART, which also carry other languages; we score MELD on a 14,713-document sample of its 220,136 documents; and we leave out two further benchmarks we ran, PeerPrism \citep{peerprism} and a personal-style post-editing set \citep{baumler-etal-2026-make}, because of data-quality problems (their tables are in the repository). Before scoring, every document goes through the same format gate and outline extractor as at test time (\S\ref{sec:method}). \autoref{tab:benchmarks} lists what each benchmark contributes.
\begin{table}[!ht]\caption{\textbf{Every arm we evaluate and its label under the idea-authorship rule.} ``Attacked'' marks model-written text that was humanized or adversarially rewritten; its ideas stay the model's. ``Mixed'' arms are reported as P(AI) only; ``not placed'' arms keep most of a model's prose after a person's edit and are left out of the quadrants.}\label{tab:labels}
\begin{center}\scriptsize\setlength{\tabcolsep}{3pt}\begin{tabular}{p{0.17\linewidth}p{0.37\linewidth}p{0.14\linewidth}p{0.13\linewidth}p{0.11\linewidth}}
\toprule Benchmark & Arm & Ideas by & Prose by & Label \\ \midrule
AI Peer Review Detection & human reviews / model reviews & person / model & person / model & HH / AA \\
AI in Peer Review (Saha et al.) & H: human review & person & person & HH \\
 & H-AI: human review polished by a model (and humanized) & person & model & HA \\
 & AI-BP, AI-EP: model review from the paper (and humanized) & model & model & AA (attacked) \\
 & AI-HI: model review from the reviewer's notes & mixed & model & mixed \\
DetectRL-X & human / model / attacked model text (English rows) & person / model / model & person / model / model & HH / AA / AA attacked \\
StoryScope & human stories / stories by five models & person / model & person / model & HH / AA \\
DetectionAI & human essays / model essays / humanized (\citealp{stealthgpt}) & person / model / model & person / model / model & HH / AA / AA attacked \\
OpAI-Bench & v0: the human draft / v1--v8: successive model edits & person & person / model & HH / HA \\
Sem-Detect & human reviews / model rewrite of a human review / model reviews & person / person / model & person / model / model & HH / HA / AA \\
MELD & human / model / six attacks & person / model / model & person / model / model & HH / AA / AA attacked \\
LAMP & model draft / the draft after a professional's edit & model & model / model, edited & AA / not placed \\
ARB & human / model rewrite of a human article (h2l) / model (free\_llm) / llm2l & person / person / model / model & person / model / model / model & HH / HA / AA / AA attacked \\
GEDE & human essay / model improves or rewrites it & person & person / model & HH / HA \\
 & model essay from the assignment / DIPPER, rewrite attack & model & model & AA / AA attacked \\
 & model essay from the student's summary & model (GEDE's own label) & model & AA, shown but not averaged \\
Perkins et al. & model text under six evasion techniques & model & model & AA attacked \\
HART & human / model rephrase of human text / model / humanized & person / person / model / model & person / model / model / model & HH / HA / AA / AA attacked \\
 & model text rewritten by a person & model & model, edited & not placed \\
Epoch author imitation & the author's text / model, plain prompt & person / model & person / model & HH / AA \\
 & model imitating the author from five passages & mixed & model & mixed \\
CoCoNUTS & human review / model polishes or translates it / model review & person / person / model & person / model / model & HH / HA / AA \\
 & model review from the person's review & mixed & model & mixed \\
ELLIPSE & essays by English learners & person & person & HH \\
AI-Researcher & expert reviews / proposals from expert ideas / model proposals & person / person / model & person / model / model & HH / HA / AA \\
 & papers people wrote from a model's idea & mixed & person & mixed \\
Academic Integrity & original abstract / model abstract from it (and humanized) & person & person / model & HH / HA \\
GEN & human / edited by an AI editor / written by an AI writer & person / person / model & person / model / model & HH / HA / AA \\
 & an AI editor adds details to a person's text & mixed & model & mixed (repository only) \\
\midrule
\collabLadder{}, human sources & level 0--1 / levels 2--3 / level 4 / level 5 & model / mixed / mixed / person & model & AA / mixed / mixed / HA \\
\collabLadder{}, AI sources & levels 0--5 & model & model & AA \\
\xLadder{} & human pages / level 0 / level 5 & person / model / person & person / model / model & HH / AA / HA \\
\twicetold{} & human rewrites / the model originals & model & person / model & AH / AA \\
C4 (2019) & web pages & person & person & HH \\
\bottomrule\end{tabular}\end{center}\end{table}
\input{tables/appx/benchmarks}

\section{Data pipeline}
\label{app:pipeline}
This section documents how a web document becomes the outline \model{} reads: the discourse roles for each format, the extractor that writes the outline and how we chose it, what the outlines look like, and the paraphrase used for training.

\FloatBarrier
\subsection{Role discovery}
\label{app:roles}
For each format, GPT-5.6 Sol \citep{openai2026gpt56} first proposes a base set of roles from a description of the format alone. It then labels every item of the outlines of 200 documents in that format (100 human, 100 model-written), inventing a role wherever none of the base roles fits, and finally consolidates the base and invented roles by merging duplicates and near-synonyms. Roles used by fewer than 1\% of outline items in the format are dropped and replaced by a catch-all \emph{Other} role (\autoref{tab:roles}). Each role comes with a definition, a test for when to assign it, the roles it is distinguished from, and example excerpts; \autoref{tab:roles_example} shows eight of the 42 roles for nonfiction writing. 
\input{tables/appx/roles}
\input{tables/appx/roles_example}

\input{sections/extraction_benchmark}

\FloatBarrier
\subsection{Outline statistics}
\label{app:outline_stats}
\autoref{tab:outline_stats} describes the outlines of 1,000 training documents per format: how many items and distinct roles an outline has, how long its items are before and after paraphrasing, and how little of the extracted wording survives the paraphrase (4.2\% of word 4-grams over all formats).
\input{tables/appx/outline_stats}

\FloatBarrier
\subsection{Paraphrasing removes the surface signal}
\label{app:probe}
The paraphrase should remove wording that could reveal the source's author while keeping the ideas. \emph{It removes the surface signal:} Pangram~4 scores 1,000 test documents as the document, the raw outline, and the paraphrased outline. On paraphrased outlines it labels almost every human document as AI, while \model{} still separates them (\autoref{tab:probe}).
\input{tables/main_pangram_probe}
\emph{It keeps the content:} a paraphrased outline has cosine 0.93 with its raw outline (0.39 for another document's outline of the same label and format), and 99.2\% of raw items are closest to their own paraphrase. Both outlines stay close to the document (0.79 raw, 0.76 paraphrased; 0.26 by chance), yet the paraphrase shares under 1\% of its word 5-grams with it (\autoref{tab:probe_sim:0}).
\input{tables/appx/pangram_probe_similarity}

\FloatBarrier
\subsection{A worked example}
\label{app:oped}
\autoref{tab:oped_example} gives the complete outline of the op-ed in \autoref{fig:pipeline}: the 17 items the extractor writes and their paraphrases. The paraphrase keeps each item's content and changes only its wording.
\input{tables/appx/oped_example}

\FloatBarrier
\subsection{Corpus}
\label{app:corpus}
The corpus is a 1,001,145-document sample of FineWeb, drawn from the WildAI release \citep{russell2026aitoken}, which labels each document with a WebOrganizer format and topic and a Pangram~3.3.2 authorship verdict. We keep English documents of at least 500 words in eight long-form formats that Pangram labels entirely human or entirely AI, and split them into training, validation, calibration (human only) and test sets (\autoref{tab:corpus}).
\input{tables/appx/corpus}

\section{Training, calibration and thresholds}
\label{app:training}
\FloatBarrier
\subsection{Model configurations}
\label{app:configs}
Every model is trained on the 842,301-document training split with the Pangram labels. \model{} and \twinmodel{} share the recipe exactly and differ only in their input (paraphrased outline or raw document) and the one word of the system prompt that names it; the loss is on the answer token only. Besides the Nemotron backbone of the main paper we train ModernBERT-large \citep{warner-etal-2025-smarter} and Qwen3.5-9B \citep{qwen3.5}, and fit a logistic regression on text-embedding-3-large embeddings \citep{openai2024newembeddingmodels}, on the same data (\autoref{tab:train_config}); their results on every quadrant are in \autoref{tab:cons}. The neural models train for one epoch and are scored on validation documents at evenly spaced probes; we keep the checkpoint with the highest mean TPR at 0.5, 1 and 2\% FPR. For \model{} and \twinmodel{} the probes use a stratified sample of 4,000 validation documents and the selected checkpoint is the final one. The ModernBERT and Qwen checkpoints are selected on the full 29,975-document validation split: the end-of-training checkpoint for both ModernBERT models, and step 12,056 of 13,161 (0.92 of the epoch) for Qwen. The logistic regression has no checkpoints.
\begin{table}[!ht]\caption{\textbf{Training configurations.} The neural models' learning rates follow a cosine schedule after the warm-up.}\label{tab:train_config}
\begin{center}\scriptsize\resizebox{\textwidth}{!}{\begin{tabular}{lllllll}
\toprule Model & Base & Method & Learning rate & Batch & Max length & Probes \\ \midrule
\model{} / \twinmodel{} & Nemotron-3.5-Lightning-30B-A3B & LoRA rank 64, $\alpha = 32$ (attention, MLP, Mamba projections, unembedding), Tinker & $4\times10^{-4}$, 3\% warm-up & 128 documents & 65,536 tokens & 25, and the final step \\
ModernBERT & ModernBERT-large & full fine-tuning, AdamW, weight decay 0.01 & $2\times10^{-5}$, 6\% warm-up & 16,384 tokens $\times$ 4 accumulation & 8,192 tokens & 10 (outline), 4 (document) \\
Qwen & Qwen3.5-9B & full fine-tuning, classification head & $1\times10^{-5}$ & 64 documents & 8,192 tokens & 12 \\
Logistic regression & text-embedding-3-large (3,072 dimensions) & L2-regularized, $C = 1$, L-BFGS & -- & full training split & -- & -- \\
\bottomrule\end{tabular}}\end{center}\end{table}

\FloatBarrier
\subsection{Calibration}
\label{app:calibration}
A detector outputs a probability; turning it into a verdict needs a threshold, and a threshold is only meaningful together with the false-positive rate it produces. We set ours on a held-out calibration split of 80,000 human documents, 10,000 per format, disjoint from the validation split used for checkpoint selection. The global cut at a target FPR $\alpha$ is the $(1-\alpha)$ quantile of the calibration scores pooled over formats; all results in the paper use $\alpha = 1\%$. A per-format cut, the same quantile within the document's format, is reported after a slash in some appendix tables. We fit a cut only where it rests on at least 25 expected false positives, so a global cut exists down to a 0.03\% target and a per-format cut down to 0.25\%; of our standard operating points (0.1, 0.5, 1, 2 and 5\%), the per-format cut is therefore reported from 0.5\%. A cut calibrated on in-domain humans need not hold its FPR out of domain; we therefore report the realized FPR on every evaluation's human arm instead of assuming it (\S\ref{app:quadrants}).

\FloatBarrier
\subsection{Thresholds and baselines}
\label{app:thresholds}
\autoref{tab:thresholds} lists how every detector's score becomes a verdict. Our detectors, EditLens and the baselines that do not ship a threshold use a cut calibrated on our 80,000 human documents, so their FPR on in-domain human text is 1\% by construction; detectors that ship a threshold keep it. Pangram~4 returns the fractions of a text it judges AI-written, AI-assisted and human; following its technical report \citep{glickenhaus2026pangram4technicalreport}, we count a text as AI when its AI fraction plus half its AI-assisted fraction is at least 0.5.
\input{tables/appx/thresholds}

\section{Cost}
\label{app:cost}
All figures are in US dollars at the rates in force in September 2026. \autoref{tab:cost} gives what we paid once, to build the corpus and train the two detectors; \autoref{tab:cost_doc} gives what it costs to score a new document.

\paragraph{Building the corpus and training.} Building the 1,001,145-document corpus cost \$66,428 in batch API calls, about \$0.07 per document: \$40,255 for outline extraction (Gemini 3.7 Flash for 613k documents and Gemini 3.1 Pro for the 388k documents of an earlier build) and \$26,173 for the paraphrase (Gemini 3.1 Pro for all of them). These figures are computed from our token logs at the published batch rates; an invoice check on part of the run implies up to \$78,437. Role discovery cost a further \$116. Training on Tinker cost \$354 for \model{} and \$710 for \twinmodel{}, which reads whole documents and so trains on about twice as many tokens.
\input{tables/appx/cost_build}

\paragraph{Scoring a new document.} At inference a document passes through the format gate, one outline extraction, and one forward pass of \model{} per answer token; the paraphrase is used only to build training data. For a document of average corpus length this costs about \$0.06 through the standard API and \$0.03 through the batch API. Prompt caching lowers it further in bulk: the six in-context examples are shared by every document of a format, and cached tokens bill at a tenth of the input rate, giving \$0.018 per document (\$18 per 1,000) at our corpus run's cache hit rate (\autoref{tab:cost_doc}). Extraction is over 90\% of the cost in every setting, mostly the extractor's thinking (5,678 of 6,765 output tokens); scoring on Tinker costs a fraction of a cent. \twinmodel{} skips the gate and the extractor and costs about \$0.0005 per document, so the extra cost of reading ideas rather than words is the extraction step.
\input{tables/appx/cost_doc}

\section{Results by quadrant}
\label{app:quadrants}

The tables in this section show the accuracy of every evaluation at the global 1\% cut per quadrant and where reported, at the per-format cut. Every table compares the mean accuracy of other quadrants versus the quadrant in question.
Where an evaluation has documents on both sides of our corpus's 500-word minimum, sub-rows split it by length. 
\model{} (\nemo{}) is trained on outlines and reads either the outline or the raw document; \twinmodel{} is \nemo{} trained on and reading raw documents; Pangram~4 is used at its published threshold and EditLens-Llama-3B at our calibrated 1\% cut. The per-format cut follows the slash in the \model{} outline column (\S\ref{app:calibration}); every other column and every mean row uses the global cut.

\FloatBarrier
\subsection{Human ideas, human prose}
\label{app:q1}
Nineteen evaluations contribute human-written documents. Every detector keeps its false-positive rate low here: averaged over evaluations, \model{} flags 0.6\% of them, \twinmodel{} 0.7\%, Pangram~4 0.3\% and EditLens-3B 0.2\% (\autoref{tab:q1}).
\input{tables/main_q1}

\FloatBarrier
\subsection{Model ideas, model prose}
\label{app:q3}
Nineteen evaluations contribute model-written documents. The detectors that read the text do best here: \twinmodel{} detects 98.9\% on average and Pangram~4 97.3\%, against 91.1\% for \model{} and 81.8\% for EditLens-3B (\autoref{tab:q3}). Humanized and adversarially attacked arms are marked; \S\ref{app:attacks} compares them with their clean counterparts.
\input{tables/main_q3}

\FloatBarrier
\subsection{Human ideas, model prose}
\label{app:q4}
Twelve evaluations contain a person's ideas written up, rewritten or polished by a model. This is where a prose detector and an idea detector part ways: \model{} flags 5.3\% of these documents on average, against 49.3\% for \twinmodel{}, 56.2\% for Pangram~4 and 27.9\% for EditLens-3B (\autoref{tab:q4}).
\input{tables/main_q4}

\FloatBarrier
\subsection{Model ideas, human prose}
\label{app:q2}
Only \twicetold{} places a model's ideas in prose a person wrote from scratch (\S\ref{app:aiip}). External sets where a person edited a model's draft (LAMP, HART's human-edited arm) keep most of the model's prose and are not counted here.
\input{tables/appendix/q2_full}

\FloatBarrier
\subsection{All detectors and baselines}
\label{app:cons}
\autoref{tab:cons} gives the per-quadrant means for every detector we ran, including our other backbones and the open-source baselines of \autoref{tab:thresholds}. Among the open baselines, those with a high true-positive rate on model-written text pay for it with a high false-positive rate on human text at their own thresholds.
\input{tables/appendix/consolidated_all}

\FloatBarrier
\subsection{Mixed provenance}
\label{app:mixed}
Four external benchmarks contain documents whose ideas are genuinely shared: a model writing a review from a person's notes (AI in Peer Review) or from a person's review (CoCoNUTS), a model imitating a named author from five of their passages (Epoch), and papers people wrote from a model's research idea (AI-Researcher). These have no correct binary label, so we compare their mean P(AI) with the same benchmark's clear arms (the full comparison, with a fifth mixed arm, GEN's AI editor adding details to a person's text, is in the repository). \model{} places them between the two: on AI in Peer Review, reviews written from a person's notes score 68.5, between model reviews from the paper (77.1) and human reviews polished by a model (4.1), while \twinmodel{} (99.3) and Pangram~4 (89.8) score them like fully model-written reviews.

\FloatBarrier
\subsection{OpAI-Bench: successive model edits of a human draft}
\label{app:opai}
OpAI-Bench \citep{Bsharat2026OperationGuidedPH} starts from 4,145 human drafts (v0) and applies eight successive model operations to each, so that by v8 a model has written almost every word while the ideas remain the person's. Every version is therefore human-ideated and every flag is a false positive. \model{} stays flat, flagging 0.3\% of the drafts and 0.2 to 0.6\% of every edited version, while \twinmodel{} rises to 26.3\% and Pangram~4 to 48.5\% as the model's share of the prose grows (\autoref{tab:opai:0}).
\input{tables/appx/opai}

\section{The evaluation sets we built}
\label{app:ours}
\FloatBarrier
\subsection{\collabLadder{}}
\label{app:ladder}
\collabLadder{} tests how \model{} responds as a model receives more of a source document's plan. For each of 500 test documents (237 human-written, 263 model-written), we write six prompts that reveal progressively more of the source (\autoref{tab:ladder_example}). Level 0 gives only the document type, subject and length, and each later level adds more: genre, audience, and tone (1); central theme (2); order of discourse roles (3); full outline (4); and an instruction to realize the outline exactly, adding nothing (5). One model writes all six versions of a source: GPT-5.6 Sol for about half the sources, GPT-5.6 Terra or Luna for the rest \citep{openai2026gpt56}. GPT-5.6 Sol, at high reasoning effort, writes the briefs for levels 0 to 4 from the source's outline; the level-5 prompt is a fixed template filled with the outline.
\input{tables/appx/ladder_example}
On human sources, the ideas shift from the model's (levels 0--1) to the person's (level 5); on model sources they are a model's at every level. \model{} follows the ideas: on human sources it flags 94.9\% of level-0 and 6.8\% of level-5 documents, and on model sources 96.6--99.2\% at every level. \twinmodel{} and Pangram~4 flag at least 86.5\% at every level of both (\autoref{tab:ladder_h:0} and~\autoref{tab:ladder_a:0}).
\input{tables/appx/t3h}
\input{tables/appx/t3a}
Each level carries the intended share of the source (\autoref{tab:ladder_sim}). On human sources, the share of the source's outline items a document keeps rises from 0.30 at level 0 to 0.99 at level 5 (0.05 for an unrelated source of the same format and topic), and its cosine similarity to the source rises from 0.52 to 0.87 (0.33 unrelated). AI sources follow the same curve.
\input{tables/appx/ladder_similarity}

\FloatBarrier
\subsection{\twicetold{}}
\label{app:aiip}
\twicetold{} is the reverse of \collabLadder{}: a model supplies the ideas and a person writes the prose. GPT-5.6 Sol \citep{openai2026gpt56} wrote 500-word stories from one-line genre prompts, and we extracted and paraphrased an outline of each with our pipeline. Seventeen writers (computer science graduate students and faculty) then wrote each story from its outline alone, without seeing the model's version (instructions in \S\ref{app:prompts}). They wrote in person rather than through crowdsourcing, to reduce the chance of AI assistance \citep{10.1145/3685527}. The 50 resulting stories have a model's ideas and a person's prose, so \model{} should flag them as AI. They average 837 words, against 502 for the model versions.

\subsubsection{Outline Adherence}
We analyze the faithfulness of the rewrite to the original outline to measure quality. On average, 94.1\% of the original outline is retained and 20.1\% of the rewrite-outline is extraneous. Authors rarely deviate from the chronology: 90.7\% of matched items are in order.

\subsubsection{Robustness to Renaming}
Character names and locations in \twicetold stories are largely decided by the model outlines. Model outlines exhibit strong tendencies towards certain names, including ``Mara'', names starting with ``El-'', and towns named ``Bellweather''. We analyze whether replacing character and location names with random ones sampled from the most popular U.S. baby names of the 2010s and from U.S. cities respectively affects the verdicts of our models. The mean P(AI) of \model{} decreases after renaming, from 79.8 to 69.4, while \twinmodel{}'s mean P(AI) decreases from 0.14 to 0.04. %

\subsubsection{Results}
\model{} flags 68.0\% of the rewrites and 96.0\% of the model originals they were written from using the global cut, while \twinmodel{}, EditLens-3B and Pangram~4 flag nearly every original but almost none of the human versions (\autoref{tab:twicetold}). All human and original versions fall into the Creative Writing format.
\input{tables/appx/twicetold_final}

\FloatBarrier
\subsection{\xLadder{}}
\label{app:multiling}
\xLadder{} tests whether idea detection carries over to other languages. \model{} is trained only on English, but it never reads the document itself: the extractor writes an English outline of a document in any language. We sample 10,160 human-written pages from FineWeb2 \citep{penedo2025fineweb}, all written before 2022, in 24 languages grouped by the size of their FineWeb2 share: high-resource (Chinese, French, German, Japanese, Russian, Spanish), mid-resource (Arabic, Bengali, Hindi, Indonesian, Persian, Thai, Ukrainian, Vietnamese) and low-resource (Amharic, Khmer, Malayalam, Marathi, Nepali, Sinhala, Swahili, Tamil, Urdu, Yoruba). For each page a model writes two versions in the page's language, following the \collabLadder{} protocol: level 0 from a one-line description of the document type and subject (model ideas), and level 5 from the page's full outline, adding nothing (human ideas). Level 0 and level 5 documents were generated by Gemini 3.7 Flash \citep{gemini37flash}.

The human pages are almost never flagged (0.1\%), \model{} detects 95.3\% of level-0 documents and flags 0.7\% of level-5 documents, and the level-0 detection rate holds across resource groups (94.2\%, 96.4\% and 95.4\% for high-, mid- and low-resource languages). Detectors that read the prose depend on the language: on low-resource level-0 documents \twinmodel{} detects 51.6\% and Pangram~4 58.0\% (\autoref{tab:multiling} shows four languages per group and the average over all 24; every language and the group averages are in the repository).
\input{tables/main_multiling}

\FloatBarrier
\subsection{Translation}
\label{app:translation}
To separate language from ideas, we translate a subset of \xLadder{} (50 pages per language for the human pages, level 0 and level 5; 3,594 documents) into English with Gemini 3.7 Flash and score both versions (\autoref{tab:translation}). \model{}'s verdicts barely move: level-0 TPR changes by under a point and the human FPR not at all, and its scores on the two versions correlate at 0.946 (Spearman). \twinmodel{}'s level-0 TPR rises from 74.7 to 98.2\%, all of it in low-resource languages, and its level-5 FPR rises by 13 points, because translated human ideas now read as model prose. Pangram~4 shows the same effect more strongly: its level-5 FPR goes from 51.3 to 97.0\%.
\input{tables/appx/translation}

\section{Robustness and generalization}
\label{app:robustness_all}
\FloatBarrier
\subsection{Pre-ChatGPT web text}
\label{app:c4}
Our in-domain human documents were crawled between 2024 and 2026, and some are likely AI-written despite Pangram's label. As a cleaner check of the false-positive rate we sample 10,000 pages from the April 2019 C4 snapshot \citep{c4dataset}, which predates ChatGPT, stratified by format. Every flag is a false positive: \model{} flags 0.01\%, and \twinmodel{}, Pangram~4 and EditLens-3B none.

\FloatBarrier
\subsection{Attacks and humanizers}
\label{app:attacks}
Eight external benchmarks pair clean model-written documents with versions that were humanized or adversarially rewritten, which keeps the ideas a model's. \autoref{tab:attacks_app} compares each detector's TPR on the clean and attacked arms: averaged over the seven evaluations with both, \model{} drops by 6.4 points, \twinmodel{} by 10.6, Pangram~4 by 4.9 and EditLens-3B by 42.6. The per-attack tables for MELD and DetectRL-X show which attacks cost the most: character-level attacks such as zero-width spaces and homoglyphs, which the extractor normalizes away, leave \model{} untouched, while paraphrasing costs it the most.
\input{tables/appx/attacks}
\input{tables/appx/meld_attacks}
\input{tables/appx/detectrl_attacks}

\FloatBarrier
\subsection{Do attacks preserve content?}
\label{app:attack_similarity}

An attack that changes a document's content can also change its ideas, so a lower TPR is not necessarily a failure of idea detection. We measure how much of each clean AI document survives its attack (\autoref{tab:attack_sim_doc:0} and~\autoref{tab:attack_sim_items:0}). Character-level attacks, polishing, condensing, and encoder paraphrasing barely change the document (cosine 0.98--0.99) or \model{}'s TPR. The attacks that lower TPR change the content: StealthGPT's humanized documents have cosine 0.873 with their originals, and 68.2\% fall below the 95th percentile for a different author's document on the same prompt. For DetectRL-X's decoder and seq2seq paraphrases, the documents \model{} still catches keep far more of the original than those it misses (cosine 0.993 and 0.979 vs.\ 0.853 and 0.768; 84\% and 81\% vs.\ 34\% and 21\% of outline items).

\input{tables/appx/attack_similarity_doc}
\input{tables/appx/attack_similarity}

\FloatBarrier
\subsection{Content-preserving transformations}
\label{app:robustness}

If \model{} reads ideas, transformations that keep a document's ideas should not change its verdict. On in-domain test documents we (i) score the raw extracted outline instead of its paraphrase, (ii) paraphrase the source document sentence by sentence before extraction, (iii) extract the outline with a different model family (GPT-5.6) and (iv) score a 500-word excerpt instead of the full document. The verdict survives the first three for 98.6\%, 98.6\% and 98.8\% of documents; excerpts lower the TPR from 96.5\% to 91.4\% (\autoref{tab:ablations} and \autoref{tab:t0:0} to \autoref{tab:t14:0}). Stripping the role labels or shuffling the items barely changes the AUROC (0.993), while the role sequence alone reaches only 0.601.
\input{tables/main_ablations}
\input{tables/appx/t0}
\input{tables/appx/t1}
\input{tables/appx/t4}
\input{tables/appx/t14}

\subsection{Format classification and extraction variance}
\label{app:stochasticity}
At test time a document passes through two stochastic steps before \model{} reads it: the format gate, which chooses the role vocabulary, and the outline extractor. Both run at the production settings (Gemini 3.7 Flash at default temperature; the extractor with six exemplars and high reasoning effort).

\paragraph{Format gate.} The gate is consistent with itself: the same format in all three of three runs for 90.9\% of 1,000 in-domain documents, and in all five of five runs for 93.2\% of the 400 out-of-domain documents below. On those 400 documents, 9.6\% of its runs place the document outside the eight formats with a role vocabulary.

\paragraph{Repeated extraction.} We extract each document five times under its recorded format and score every outline with \model{}: 33 in-domain test documents (the 33 of 42 sampled that completed all five extractions) and 400 out-of-domain documents from MELD, Saha et al., and DetectRL-X, half drawn at random and half near the decision threshold (\autoref{tab:stochasticity}). The outlines vary between runs (the number of items has a coefficient of variation of about 9\%), but the verdict rarely does: it never changes for an in-domain document and is unchanged for 94.0\% of the random out-of-domain documents, and no run flags any of their human-written documents as AI. Instability concentrates near the threshold, where only 52.5\% of documents keep their verdict in all five runs; \autoref{tab:stochasticity_flips} breaks this down by distance from the cut. We reran extraction but not the format gate, so these figures reflect extraction variance alone. Averaging several extractions would likely reduce it; we do not test this.
\input{tables/appx/stochasticity}
\input{tables/appx/stochasticity_flips}

\FloatBarrier
\subsection{Newer generators}
\label{app:unseen}
Models released after our corpus's June 2026 cutoff may produce ideas that older models did not. We prompt four of them, GPT-5.6 Sol \citep{openai2026gpt56}, GPT-6 Astra \citep{openai2026gpt6astra}, Claude Opus 5 \citep{anthropic2026opus5}, and Claude Fable 5.1 \citep{anthropic2026fable51}, with 250 writing prompts sampled from StoryScope (\autoref{tab:unseen:0}). At the global 1\% cut, \model{} detects 78.5\%, 62.8\%, 51.0\%, and 51.2\% of their stories (90.7, 74.8, 65.7, and 65.6\% at the per-format cut), against 97.0\% for the five older StoryScope generators. Pangram~4 flags 99.2--100\% of every new generator's stories. \twinmodel{} holds on the GPT models (98.8 and 83.2\%) but falls to 41.2\% on Opus 5 and 17.2\% on Fable 5.1, so the drop is not specific to reading ideas.
\input{tables/appx/test32}

\input{sections/app_feat_analysis}

\section{Prompts}
\label{app:prompts}
\input{sections/appendix_prompts}

\section{Where the remaining tables are}
\label{app:evals}

All other tables are in our GitHub repository (\url{https://github.com/RishanthRajendhran/IdeaLens/tree/main/paper}): every evaluation's full tables with all arms and length splits, every operating point (0.1, 0.5, 1, 2 and 5\% false-positive targets) under the global cut, the per-format 1\% cut on every arm (\S\ref{app:calibration}), every detector on every arm of every evaluation including the Qwen, logistic-regression and ModernBERT models, the similarity diagnostics, all prompts and role vocabularies, and the code.

%% file: sections/08_limitations.tex
\section{Limitations}
\label{app:limitations}

\paragraph{Noisy training labels.} We have no idea-authorship labels at scale, so we train on Pangram~3.3.2's prose-authorship labels and assume that a document's ideas and prose usually share an author. We keep only documents that Pangram labels fully human or fully AI, which excludes the AI-assisted cases where this assumption is weakest, but some labels are still wrong for our task: a person may have written up a model's ideas, or a model may have polished a person's. \model{} nonetheless departs sharply from Pangram wherever the two kinds of authorship differ (\S\ref{sec:results}). It is still less accurate on mixed-authorship documents than on conventional ones, with a higher FPR on human ideas in model prose and a lower TPR on model ideas in human prose. We do not measure how much of this gap comes from label noise, nor test ways to reduce it.

\paragraph{Long-form English documents only.} Our corpus contains only English documents of at least 500 words in eight long-form WebOrganizer formats, chosen so that each document carries enough ideas to detect. \model{} still works on shorter documents but is weaker on them in the conventional quadrants, mostly at detecting AI text (87.5\% vs.\ 90.4\% TPR; \autoref{tab:length_split}). Non-English documents reach \model{} only through English outlines, so its multilingual performance depends on the extractor (\S\ref{app:multiling}).

\begin{table}[!ht]
\caption{\textbf{Accuracy by document length} on the out-of-domain evaluations (\%, global 1\% cut). The training corpus has no documents under 500 words.}
\label{tab:length_split}
\begin{center}\small
\begin{tabular}{llrr}
\toprule
Quadrant & Rate & $<$500 words & $\geq$500 words \\
\midrule
Q1: human ideas, human prose & FPR & 0.7 & 0.2 \\
Q3: model ideas, model prose & TPR & 87.5 & 90.4 \\
Q4: human ideas, model prose & FPR & 4.2 & 5.2 \\
\bottomrule
\end{tabular}
\end{center}
\end{table}

\paragraph{Dependence on a proprietary extractor.} At inference, Gemini 3.7 Flash classifies the format and extracts the outline, at about \$0.018 per document in bulk (\S\ref{app:cost}). Swapping in other extractors barely changes \model{}'s verdicts (\autoref{tab:ablations}), but we tested only proprietary frontier models, whose cost and availability may limit adoption. Reading the raw document instead avoids the extractor but loses much of the sensitivity to idea authorship (\S\ref{sec:results}). Our released corpus pairs 1M documents with frontier-model outlines and could be used to train an open extractor; we leave this to future work.

\paragraph{Extraction variance.} Extraction is stochastic. Across five extractions of the same document, \model{}'s verdict is unchanged for 94\% of randomly drawn out-of-domain documents but for only 52.5\% of those near the decision threshold (\S\ref{app:stochasticity}). Averaging several extractions would likely reduce this variance; we do not test it.

\paragraph{Newer generators.} On stories from four models released after our corpus's cutoff, \model{} detects 51--79\% at the global cut, against 97\% for the older StoryScope generators (\S\ref{app:unseen}). \twinmodel{} also drops on the two Claude models, so the drop is not specific to reading ideas, but Pangram~4 does not drop. Keeping \model{} current will require training on newer generators.

\paragraph{Content-changing attacks.} \model{} is less robust than Pangram~4 to some paraphrasing attacks: it detects 77.6\% of DIPPER paraphrases in GEDE, against 95.1\% for Pangram~4 (\S\ref{sec:results}). Some of these attacks change content as well as style (\S\ref{app:attack_similarity}), so an idea detector should arguably be less certain about them, but we cannot fully separate that from a genuine weakness.

\paragraph{Little evaluation data with model ideas in human prose.} Existing benchmarks either lack this quadrant or approximate it with human edits of model drafts that keep most of the model's prose, so our only test is the 50 stories of \twicetold{}. The results there favor \model{} over prose detectors, but 50 stories cannot establish how well this generalizes; a larger dataset is future work.

\paragraph{Binary labels for mixed ideas.} \model{} predicts human or AI, but in the collaborative writing that motivates this work, ideas are often part human and part AI. Whose idea a given point was can sometimes be traced, for example from chat logs, but when people and models revise each other's ideas repeatedly, the granularity at which to assign authorship is unclear. Modeling mixed authorship needs data with controlled idea sources, which we leave to future work.

%% file: tables/appx/benchmarks.tex
\begin{table}[!ht]
\caption{\textbf{The 19 external benchmarks.} Documents scored in the arms we report, and the quadrants those arms fall in: HH human ideas and human prose, AA model ideas and model prose, HA human ideas and model prose, AH model ideas and human prose; ``attacked'' marks humanized or adversarially attacked model-written arms. \autoref{tab:labels} maps every arm to its label.}
\label{tab:benchmarks}
\begin{center}\scriptsize
\begin{tabularx}{\linewidth}{>{\raggedright\arraybackslash}X>{\raggedright\arraybackslash}Xrl}
\toprule
Benchmark & Domain & Documents & Quadrants \\
\midrule
AI Peer Review Detection \citep{ICLR2026_64592277} & peer reviews & 9,956 & HH, AA \\
AI in Peer Review \citep{saha2026policies} & peer reviews & 28,107 & HH, AA, HA (+ attacked) \\
DetectRL-X \citep{wu-etal-2026-detectrl-x} & academic, news, fiction, encyclopedia, web and SEO text (English rows) & 19,916 & HH, AA (+ attacked) \\
StoryScope \citep{russell2026storyscope} & short fiction & 8,291 & HH, AA \\
DetectionAI \citep{jabarian2025artificial} & fiction, news, product and restaurant reviews, blogs, r\'esum\'es & 17,340 & HH, AA (+ attacked) \\
OpAI-Bench \citep{Bsharat2026OperationGuidedPH} & essays and news articles, edited by successive model operations & 37,308 & HH, HA \\
Sem-Detect \citep{duarte2026semdetect} & peer reviews & 22,146 & HH, AA, HA \\
MELD \citep{li2026meldmultitaskequilibratedlearning} & news, books, Wikipedia, recipes, poetry, abstracts, Reddit, reviews & 14,713 & HH, AA (+ attacked) \\
LAMP \citep{LAMP} & fiction and creative nonfiction drafts & 1,057 & AA \\
ARB \citep{perrone2026arbmatchedauthorshiprewritingbenchmark} & web text, news and short stories & 22,726 & HH, AA, HA (+ attacked) \\
GEDE \citep{Gehring2025LimitsOL} & student essays & 10,037 & HH, AA, HA (+ attacked) \\
Perkins et al. \citep{perkins2024simple} & academic writing, with evasion techniques & 86 & AA (+ attacked) \\
HART \citep{Bao2025DecouplingCA} & news, essays, creative writing, arXiv papers (English rows) & 15,371 & HH, AA, HA (+ attacked) \\
Epoch author imitation \citep{lee2026ai} & blog posts, fiction and scientific prose by named authors & 792 & HH, AA \\
CoCoNUTS \citep{chen-etal-2026-coconuts} & peer reviews & 5,584 & HH, AA, HA \\
ELLIPSE \citep{crossley2023ellipse} & essays by English learners & 3,909 & HH \\
AI-Researcher \citep{si2025can,si2026the} & research idea proposals, papers written from them, expert reviews & 649 & HH, AA, HA \\
Academic Integrity \citep{Karr2026WhyAD} & paper abstracts & 2,312 & HH, HA \\
GEN (AI Writers/Editors) \citep{shan2026stylometric} & Wikipedia, WikiHow, arXiv, Reddit posts and stories & 21,187 & HH, AA, HA \\
\bottomrule
\end{tabularx}
\end{center}
\end{table}

%% file: tables/appx/roles.tex
\begin{table}[h]
\caption{\textbf{Role discovery by format.} Base roles proposed from the format description alone; candidate roles after labelling the items of 200 documents (100 human, 100 model-written) and inventing roles where none fit; the final vocabulary after consolidation, split into roles kept from the base set and roles invented during labelling.}
\label{tab:roles}
\begin{center}\small
\begin{tabular}{lrrrrr}
\toprule
Format & Base roles & Candidates & Final roles & kept from base & invented \\
\midrule
Academic Writing & 44 & 67 & 52 & 42 & 10 \\
Creative Writing & 30 & 84 & 50 & 30 & 20 \\
Knowledge Article & 33 & 70 & 47 & 31 & 16 \\
News Article & 33 & 54 & 41 & 33 & 8 \\
Nonfiction Writing & 40 & 51 & 42 & 37 & 5 \\
Personal About Page & 22 & 269 & 62 & 22 & 40 \\
Personal Blog & 37 & 46 & 43 & 36 & 7 \\
User Reviews & 21 & 126 & 45 & 20 & 25 \\
\bottomrule
\end{tabular}
\end{center}
\end{table}

%% file: tables/appx/roles_example.tex
\begin{table}[!ht]
\caption{\textbf{Eight of the 42 discourse roles discovered for nonfiction writing}, with the definitions the extractor receives. Each role also carries an assignment test, the roles it is distinguished from, and example excerpts; the full vocabularies of all eight formats are in the repository (the excerpts, which quote the training corpus, are released with the corpus).}
\label{tab:roles_example}
\begin{center}\scriptsize
\begin{tabular}{p{0.2\linewidth}p{0.72\linewidth}}
\toprule
Role & Definition \\
\midrule
Explanation & Content that makes a phenomenon intelligible by identifying causes, mechanisms, processes, reasons, or functional relationships. \\
Recommendation & Content that advocates a course of action or decision as desirable, preferable, or necessary. \\
Evidence & Content that supplies verifiable facts, records, observations, data, or sourced information as grounds for assessing another proposition. \\
Event Account & Content that recounts what happened, including relevant participants, actions, chronology, and immediate consequences. \\
Illustration & Content that clarifies or exemplifies a broader point through a particular case, anecdote, scenario, or analogy. \\
Framing & Content that establishes the scope, lens, premise, governing concern, or terms through which other material is to be understood. \\
Synthesis & Content that integrates claims, evidence, events, perspectives, or themes into a higher-order understanding or takeaway. \\
Instruction & Content that provides actionable steps, methods, conditions, or cautions for carrying out a task or reaching a goal. \\
\bottomrule
\end{tabular}
\end{center}
\end{table}

%% file: sections/extraction_benchmark.tex
\subsection{Outline extraction benchmark}
\label{app:extract_bench}

We chose the corpus extractors with a small benchmark (\autoref{tab:extract_bench}). Gemini 3.1 Pro and Gemini 3.7 Flash, both with six few-shot examples, extracted the corpus; Gemini 3.7 Flash extracts every out-of-domain evaluation document.

\paragraph{Setup.} The benchmark has 198 documents: eleven human-written and eleven AI-generated documents from each of eight formats, with authorship confirmed by both EditLens and Pangram~4. Reference outlines come from Claude Fable 5. We compare twelve configurations: GPT-5.6 Sol, Terra, and Luna zero-shot, and Gemini 3.1 Pro, 3.6 Flash \citep{gemini36flash}, and 3.7 Flash with zero, three, or six few-shot examples, each a full document with its reference outline drawn from a separate pool. The main score aligns candidate and reference items in order by embedding similarity, allowing merges, splits, omissions, and additions; the table caption defines the other metrics.

\paragraph{Results.} Every configuration captures nearly all reference content (coverage $\geq 0.986$); they differ mainly in granularity, that is, how finely they split content into items. Few-shot examples matter more than the choice of model: six examples raise alignment by 0.12--0.13 for every Gemini model. Gemini 3.1 Pro at six shots is within 0.03 of the best configuration (0.769 vs.\ 0.796 for GPT-5.6 Luna), and Gemini 3.7 Flash at six shots matches it (0.774) at about a third of the cost, making it the cheaper choice for the rest of the corpus.

\paragraph{Caveats.} Scores measure agreement with a single reference model, not correctness. Few-shot examples also increase phrasing copied from the document (from 2.6\% to 7.8\% of item words for 3.7 Flash), which the paraphrasing step then removes (\S\ref{app:probe}). Finally, no test directly swaps 3.1 Pro for 3.7 Flash; the extractor swap in \autoref{tab:ablations} uses GPT-5.6 extractors.

\input{tables/appendix/extract_bench}

%% file: tables/appendix/extract_bench.tex
\begin{table}[h]
\caption{\capExtractBench}
\label{tab:extract_bench}
\begin{center}\footnotesize
\setlength{\tabcolsep}{3pt}
\adjustbox{max width=\textwidth}{\begin{tabular}{@{}lrrrrrrrrr@{}}
\toprule
 & & \multicolumn{2}{c}{Coverage} & \multicolumn{2}{c}{Granularity} & & & & \\
\cmidrule(lr){3-4} \cmidrule(lr){5-6}
Configuration & Alignment & reference & candidate & agreement & split$-$merge & Items (cand.\,/\,ref.) & Role agreement & \$ per doc. & Judge rank\,/\,score \\
\midrule
GPT-5.6 Luna, 0-shot & 0.796 & 0.997 & 1.000 & 0.760 & +0.121 & 19.7\,/\,17.9 & 0.630 & 0.022 & 2.04\,/\,96.3 \\
GPT-5.6 Sol, 0-shot & 0.782 & 0.988 & 1.000 & 0.778 & +0.107 & 19.5\,/\,17.9 & 0.694 & 0.087 & 1.80\,/\,96.2 \\
\textbf{Gemini 3.7 Flash, 6-shot} & \textbf{0.774} & 0.998 & 0.999 & 0.787 & $-$0.130 & 16.0\,/\,18.0 & 0.646 & \textbf{0.017} & 4.76\,/\,90.9 \\
\textbf{Gemini 3.1 Pro, 6-shot} & \textbf{0.769} & 0.997 & 0.999 & 0.743 & $-$0.139 & 16.1\,/\,18.0 & 0.660 & \textbf{0.046} & 5.60\,/\,88.4 \\
GPT-5.6 Terra, 0-shot & 0.769 & 0.986 & 1.000 & 0.779 & $-$0.142 & 15.7\,/\,17.9 & 0.632 & 0.031 & 3.52\,/\,92.8 \\
Gemini 3.7 Flash, 3-shot & 0.763 & 1.000 & 0.999 & 0.796 & $-$0.094 & 16.4\,/\,18.0 & 0.659 & 0.013 & 4.79\,/\,90.8 \\
Gemini 3.1 Pro, 3-shot & 0.755 & 0.998 & 0.998 & 0.740 & $-$0.146 & 16.2\,/\,18.0 & 0.655 & 0.036 & 5.85\,/\,87.6 \\
Gemini 3.6 Flash, 6-shot & 0.751 & 0.999 & 1.000 & 0.744 & $-$0.220 & 14.9\,/\,18.0 & 0.618 & 0.033 & -- \\
Gemini 3.6 Flash, 3-shot & 0.739 & 0.998 & 1.000 & 0.762 & $-$0.175 & 15.5\,/\,18.0 & 0.621 & 0.026 & -- \\
Gemini 3.7 Flash, 0-shot & 0.657 & 0.997 & 0.999 & 0.698 & $-$0.348 & 13.5\,/\,18.0 & 0.599 & 0.007 & 7.65\,/\,81.5 \\
Gemini 3.1 Pro, 0-shot & 0.636 & 0.995 & 1.000 & 0.687 & $-$0.258 & 14.7\,/\,18.0 & 0.623 & 0.019 & -- \\
Gemini 3.6 Flash, 0-shot & 0.632 & 0.998 & 0.999 & 0.672 & $-$0.363 & 13.4\,/\,18.0 & 0.579 & 0.013 & -- \\
\bottomrule
\end{tabular}}
\end{center}
\end{table}

%% file: tables/appx/outline_stats.tex
\begin{table}[h]
\caption{\textbf{Outline statistics by format.} 1,000 training-corpus documents per format. Items per outline (median), distinct roles used per outline, mean words per item in the extracted and the paraphrased outline, and the share of the extracted outline's word 4-grams that survive paraphrasing (all formats: 4.2\%).}
\label{tab:outline_stats}
\begin{center}\small
\begin{tabular}{lrrrrr}
\toprule
 & & & \multicolumn{2}{c}{Words per item} & \\
\cmidrule(lr){4-5}
Format & Items & Roles used & extracted & paraphrased & 4-grams kept (\%) \\
\midrule
Academic Writing & 16 & 9.6 & 26.4 & 32.6 & 3.8 \\
Creative Writing & 14 & 9.6 & 24.0 & 29.0 & 1.9 \\
Knowledge Article & 14 & 7.8 & 27.0 & 33.3 & 3.8 \\
News Article & 12 & 7.7 & 28.5 & 34.6 & 9.1 \\
Nonfiction Writing & 13 & 8.4 & 27.0 & 33.0 & 3.4 \\
Personal About Page & 12 & 8.9 & 27.0 & 32.9 & 4.9 \\
Personal Blog & 13 & 8.4 & 25.3 & 30.6 & 3.1 \\
User Reviews & 15 & 7.9 & 25.1 & 31.6 & 3.5 \\
\bottomrule
\end{tabular}
\end{center}
\end{table}

%% file: tables/main_pangram_probe.tex
\begin{table}[t]
\caption{\capProbe}
\label{tab:probe}
\begin{center}\footnotesize
\setlength{\tabcolsep}{3.5pt}
\adjustbox{max width=\textwidth}{\begin{tabular}{@{}lllrrrrrrrrr@{}}
\toprule
 & & & & \multicolumn{4}{c}{human-labelled ($n$=500)} & \multicolumn{4}{c}{AI-labelled ($n$=500)} \\
\cmidrule(lr){5-8} \cmidrule(lr){9-12}
Detector & Input & AUC & & human & assisted & AI & P(AI) & human & assisted & AI & P(AI) \\
\midrule
Pangram 4 & document & 0.935 & mean & 98.7 & 0.2 & 1.1 & 1.2 & 31.7 & 0.0 & 68.3 & 68.3 \\
\rowcolor{cellgrey} & &  & median & 100.0 & 0.0 & 0.0 & 0.0 & 12.1 & 0.0 & 87.9 & 87.9 \\
Pangram 4 & raw outline & 0.878 & mean & 8.5 & 64.8 & 26.7 & 59.1 & 2.2 & 11.0 & 86.8 & 92.3 \\
\rowcolor{cellgrey} & &  & median & 0.0 & 76.7 & 7.7 & 50.1 & 0.0 & 0.0 & 100.0 & 100.0 \\
Pangram 4 & paraphrased outline & 0.733 & mean & 0.3 & 29.7 & 70.0 & 84.8 & 0.1 & 3.7 & 96.3 & 98.1 \\
\rowcolor{cellgrey} & &  & median & 0.0 & 7.7 & 91.3 & 95.2 & 0.0 & 0.0 & 100.0 & 100.0 \\
\midrule
\model & paraphrased outline & 0.995 & mean & -- & -- & -- & 4.9 & -- & -- & -- & 96.7 \\
\rowcolor{cellgrey} & &  & median & -- & -- & -- & 0.0 & -- & -- & -- & 99.7 \\
\bottomrule
\end{tabular}}
\end{center}
\end{table}

%% file: tables/appx/pangram_probe_similarity.tex
\begin{table}[h]
\caption{\textbf{Paraphrasing preserves meaning.} 1,000 held-out test documents (500 human-labelled, 500 AI-labelled), each with its raw extracted outline and its paraphrase, embedded with text-embedding-3-large; cosine between the raw and paraphrased units. $\dagger$ marks pairs of outlines, documents or items where paraphrases are of other raw units. The $n$-gram rows give the share of an outline's word $n$-grams found verbatim in the document.}
\label{tab:probe_sim:0}
\begin{center}\footnotesize
\setlength{\tabcolsep}{2.5pt}
\adjustbox{max width=\textwidth}{\begin{tabular}{@{}lrrrrrr@{}}
\toprule
\bf Measure & \multicolumn{2}{c}{\bf All} & \multicolumn{2}{c}{\bf Human} & \multicolumn{2}{c}{\bf AI}\\
& mean & med & mean & med & mean & med \\
\midrule
\multicolumn{7}{c}{\bf Outlines}\\
\midrule
raw vs paraphrased & 0.933 & 0.938 & 0.934 & 0.937 & 0.933 & 0.939 \\
$\dagger$raw vs paraphrased & 0.389 & 0.392 & 0.360 & 0.363 & 0.418 & 0.421 \\
\midrule
\multicolumn{7}{c}{\bf Items}\\
\midrule
raw vs paraphrased & 0.834 &0.837 & 0.845 & 0.851 & 0.822 & 0.826 \\
$\dagger$raw vs paraphrased (from same outline) & 0.436 & 0.436 & 0.418 & 0.420 & 0.453 & 0.452 \\
share of raw items whose closest paraphrased item is their own & 99.2 & 100.0 & 99.6 & 100.0 & 98.8 & 100.0 \\
\midrule
\multicolumn{7}{c}{\bf Documents}\\
\midrule
document vs raw outline & 0.794 & 0.804 & 0.790 & 0.802 & 0.798 & 0.806 \\
document vs paraphrased outline & 0.758 & 0.765 & 0.754 & 0.761 & 0.762 & 0.770 \\
$\dagger$document vs raw outline & 0.262 & 0.261 & 0.239 & 0.242 & 0.285 & 0.288 \\
$\dagger$document vs paraphrased outline & 0.262 & 0.261 & 0.239 & 0.243 & 0.285 & 0.289 \\
\midrule
\multicolumn{7}{c}{\bf N-gram overlap with document (\%)}\\
\midrule
raw, 3-grams & 12.3 & 11.0 & 10.3 & 9.2 & 14.2 & 13.1 \\
paraphrased, 3-grams & 3.5 & 2.8 & 3.6 &3.0 & 3.3 & 2.7 \\
raw outline, 5-grams & 3.6 & 2.6 & 3.0 & 2.1 & 4.3 & 3.3 \\
paraphrased, 5-grams & 0.8 & 0.3 & 0.9 & 0.3 & 0.8 & 0.3 \\
raw, 8-grams & 0.9 & 0.2 & 0.8 & 0.2 & 1.1 & 0.4 \\
paraphrased, 8-grams & 0.3 & 0.0 & 0.3 & 0.0 & 0.3 & 0.0 \\
\bottomrule
\end{tabular}}
\end{center}
\end{table}

%% file: tables/appx/oped_example.tex
\begin{table}[!ht]
\caption{\textbf{The outline of the op-ed in \autoref{fig:pipeline}.} All 17 items, as extracted (what \model{} reads at test time) and after paraphrasing (what it was trained on). The op-ed is classified as Nonfiction Writing.}
\label{tab:oped_example}
\begin{center}\tiny\setlength{\tabcolsep}{3pt}
\begin{tabularx}{\linewidth}{@{}r>{\raggedright\arraybackslash}p{0.15\linewidth}XX@{}}
\toprule
\# & Role & Extracted item & Paraphrased item \\
\midrule
1 & Event Account & The Treasury Department announced on Aug. 19 that it would double long-dated bond buybacks to at least \$4 billion per operation after 30-year yields hit a 19-year high, triggering an immediate yield decline that quickly reversed to higher levels. & Following 30-year yields reaching a 19-year peak, the Treasury Department stated on Aug. 19 that the size of extended maturity repurchases would double, reaching a minimum of \$4 billion each. This decision caused a sudden drop in yields, which soon rebounded to elevated positions. \\
2 & Evaluation & The market correctly identified the expanded buyback operation as an attempt at price management rather than liquidity support, representing a significant policy error. & Investors accurately interpreted the enlarged repurchase initiative as a mechanism for controlling prices instead of providing liquidity, marking a major misstep in government strategy. \\
3 & Rebuttal & Treasury's liquidity rationale is refuted by strong market participation and orderly trading, which lacked the genuine dysfunction seen during March 2020 or the 2022 UK gilt crisis. & The justification regarding liquidity provisions is undermined by stable trading conditions and robust investor involvement, demonstrating none of the severe disruptions witnessed during the 2022 UK gilt turmoil or in March 2020. \\
4 & Evidence & Underlying economic data--including 3\% to 4\% inflation, 4.1\% unemployment, a 6\% peacetime deficit, \$40 trillion in national debt, and interest costs exceeding defense spending--justifies higher long-term yields. & Elevated rates on extended maturities are warranted by fundamental economic indicators, encompassing a national debt of \$40 trillion, a peacetime deficit of 6\%, joblessness at 4.1\%, inflation running between 3\% and 4\%, alongside debt servicing expenses that surpass military expenditures. \\
5 & Interpretation & Current bond yield levels remain historically accommodative relative to nominal economic growth, indicating the market was not acting punitively but barely adjusting to fiscal realities. & When compared to nominal economic expansion, contemporary rates on government securities are still quite lenient by historical standards, suggesting that investors were merely adapting to budgetary facts rather than imposing punitive measures. \\
6 & Explanation & Because politicians consistently avoid entitlement reform, the long-term bond yield serves as the only remaining mechanism capable of compelling democracies to confront unsustainable spending. & With elected officials routinely dodging structural changes to safety net programs, extended maturity rates function as the final available tool to force democratic nations into addressing unmanageable financial obligations. \\
7 & Problem & Suppressing yields masks fiscal urgency and subsidizes political procrastination, guaranteeing that necessary spending adjustments will be postponed at far greater ultimate cost. & Artificially depressing rates conceals the severity of budgetary crises and rewards legislative delays, ensuring that essential expenditure modifications are deferred and ultimately become substantially more expensive. \\
8 & Background & The historical precedent of 1942 to 1951 yield caps shows how debt management operations can slide into inflationary price control and prolonged financial repression. & Previous implementation of rate ceilings between 1942 and 1951 demonstrates that strategies for handling national obligations can easily deteriorate into extended periods of capital restriction and inflationary market manipulation. \\
9 & Event Account & Treasury officials subsequently signaled that buyback operations could expand indefinitely and utilize the Treasury General Account, establishing a dangerous dynamic of testing official resolve against market pressures. & Representatives from the Treasury later indicated that repurchasing activities might grow without limits by drawing upon the Treasury General Account, creating a hazardous environment where investor forces continually challenge administrative determination. \\
\bottomrule
\end{tabularx}
\end{center}
\end{table}
\begin{table}[!ht]
\caption*{\textbf{\autoref{tab:oped_example}, continued} (items 10--17).}
\begin{center}\tiny\setlength{\tabcolsep}{3pt}
\begin{tabularx}{\linewidth}{@{}r>{\raggedright\arraybackslash}p{0.15\linewidth}XX@{}}
\toprule
\# & Role & Extracted item & Paraphrased item \\
\midrule
10 & Problem & Replacing long-term bonds with short-term bills acts as backdoor monetary easing during elevated inflation, while conducting interventions near election cycles jeopardizes the Treasury market's long-term credibility. & Exchanging extended duration securities for shorter obligations functions as a covert form of economic stimulation amid high inflation, and timing these actions close to voting periods threatens the enduring reliability of the Treasury ecosystem. \\
11 & Perspective & The author highlights past warnings on fiscal policy, noting that federal transfer payments have grown to consume 70\% of outlays and criticizing past failures to lock in low borrowing rates. & Emphasizing previous cautions regarding budgetary choices, the writer points out that government redistribution initiatives currently account for 70\% of expenditures, while finding fault with earlier missed opportunities to secure inexpensive financing. \\
12 & Stakes & Projected interest expenses reaching 4.5\% of GDP guarantee that entitlement restructuring will eventually occur, either through deliberate political reform or abrupt market-imposed correction. & With anticipated financing costs approaching 4.5\% of GDP, structural modifications to safety net programs are inevitable, arriving either via intentional legislative action or through sudden, investor-driven forces. \\
13 & Rebuttal & The author rejects the defense of buybacks as routine tools, arguing that their irregular timing, doubled scale, and open-ended nature confirm they were deployed to suppress prices rather than resolve plumbing issues. & Dismissing claims that repurchase initiatives are standard administrative mechanisms, the writer contends that their unlimited scope, increased magnitude, and unusual scheduling prove these measures were utilized to depress valuations instead of addressing structural trading frictions. \\
14 & Recommendation & Policymakers should restrict buybacks to regular liquidity operations, accept market-clearing yields, and address the primary deficit through phased, generational entitlement reforms. & Lawmakers are advised to limit repurchase activities to standard maintenance of market fluidity, tolerate rates dictated by investor demand, and resolve the primary deficit utilizing gradual, multi-generational adjustments to public benefit programs. \\
15 & Claim & Implementing a credible fiscal reform package would lower long-term yields far more effectively than any scale of debt buybacks. & Executing an authentic budgetary overhaul would reduce extended borrowing rates much more successfully than repurchasing government obligations at any volume. \\
16 & Synthesis & Concludes that government efforts to defend artificial prices against economic fundamentals inevitably fail, urging Washington to heed rather than suppress the bond market's pricing signals. & The piece finishes by asserting that state attempts to uphold unnatural valuations in defiance of core economic realities are destined to collapse, advising Washington to interpret rather than stifle the indicators provided by investor yield requirements. \\
17 & Contributor Record & Biographical attribution identifying the author as chairman and CEO of Duquesne Family Office LLC. & A biographical note specifies that the writer serves as the CEO and chairman of Duquesne Family Office LLC. \\
\bottomrule
\end{tabularx}
\end{center}
\end{table}

%% file: tables/appx/corpus.tex
\begin{table}[h]
\caption{\textbf{The corpus}: 1,001,145 web documents in eight formats. Validation (29,975, checkpoint selection only) omitted. Calibration is human-only, 10,000 per format. Median words are for the train split.}
\label{tab:corpus}
\begin{center}\small
\adjustbox{max width=\textwidth}{\begin{tabular}{lrrrrrr}
\toprule
Format & Train $n$ & Train \% AI & Calibration $n$ & Test $n$ & Test \% AI & Median words \\
\midrule
Academic Writing & 71,656 & 58.7 & 10,000 & 6,264 & 50.0 & 1,051 \\
Creative Writing & 30,751 & 47.0 & 10,000 & 6,324 & 50.5 & 857 \\
Knowledge Article & 189,448 & 55.0 & 10,000 & 6,264 & 50.0 & 956 \\
News Article & 99,303 & 58.0 & 10,000 & 5,152 & 48.8 & 625 \\
Nonfiction Writing & 211,267 & 56.9 & 10,000 & 6,264 & 50.0 & 928 \\
Personal About Page & 32,082 & 41.8 & 10,000 & 5,966 & 48.0 & 652 \\
Personal Blog & 143,477 & 56.1 & 10,000 & 6,336 & 50.2 & 762 \\
User Reviews & 64,317 & 60.3 & 10,000 & 6,299 & 50.0 & 813 \\
\midrule
Total & 842,301 & 56.0 & 80,000 & 48,869 & 49.7 & 823 \\
\bottomrule
\end{tabular}}
\end{center}
\end{table}

%% file: tables/appx/thresholds.tex
\begin{table}[!ht]
\caption{\textbf{How each detector's scores become verdicts.} Calibrated cuts are the 1\% quantile of the 80,000 human documents in our calibration split, pooled over formats (\S\ref{app:calibration}); the cut column gives its value on the detector's own score scale. Pangram~4 returns the fractions of a text it judges AI, AI-assisted and human; we weight AI-assisted text by one half.}
\label{tab:thresholds}
\begin{center}\scriptsize
\begin{tabular}{llr}
\toprule
Detector & Threshold & Cut \\
\midrule
\multicolumn{3}{l}{\emph{Our detectors: flagged as AI when P(human) falls below the cut}} \\
\hspace{1em}\model{} (outline) & calibrated, global 1\% & 0.137 \\
\hspace{1em}\model{} (document) & calibrated, global 1\% & 0.087 \\
\hspace{1em}\twinmodel{} & calibrated, global 1\% & 0.605 \\
\hspace{1em}\modelModernBERT{} (outline) & calibrated, global 1\% & 0.055 \\
\hspace{1em}\modelModernBERT{} (document) & calibrated, global 1\% & 0.046 \\
\hspace{1em}\twinmodelModernBERT{} & calibrated, global 1\% & 0.188 \\
\hspace{1em}\modelQwen{} (outline) & calibrated, global 1\% & 0.052 \\
\hspace{1em}\modelQwen{} (document) & calibrated, global 1\% & 0.016 \\
\hspace{1em}\modelLR{} & calibrated, global 1\% & 0.060 \\
\multicolumn{3}{l}{\emph{Baselines without a published threshold: flagged as AI when the score exceeds the cut}} \\
\hspace{1em}EditLens-Llama-3B \citep{thai2026editlens} & calibrated, global 1\% & 0.828 \\
\hspace{1em}EditLens-RoBERTa \citep{thai2026editlens} & calibrated, global 1\% & 0.905 \\
\hspace{1em}Likelihood \citep{gehrmann-etal-2019-gltr} & calibrated, global 1\% & -1.33 \\
\hspace{1em}Log-rank \citep{mitchell2023detectgpt} & calibrated, global 1\% & -0.573 \\
\hspace{1em}Rank \citep{gehrmann-etal-2019-gltr} & calibrated, global 1\% & -8.82 \\
\hspace{1em}Entropy \citep{gehrmann-etal-2019-gltr} & calibrated, global 1\% & -1.41 \\
\hspace{1em}LRR \citep{su-etal-2023-detectllm} & calibrated, global 1\% & 2.36 \\
\multicolumn{3}{l}{\emph{Baselines at their own published threshold}} \\
\hspace{1em}Binoculars \citep{hans2024spotting} & own & -- \\
\hspace{1em}Fast-DetectGPT \citep{bao2024fastdetectgpt} & own & -- \\
\hspace{1em}MELD \citep{li2026meldmultitaskequilibratedlearning} & own & -- \\
\hspace{1em}Desklib \citep{desklib2025aitextdetector} & own & -- \\
\hspace{1em}Desklib-academic \citep{desklib2025aitextdetector} & own & -- \\
\hspace{1em}MAGE \citep{li-etal-2024-mage} & own & -- \\
\hspace{1em}OpenAI-RoBERTa-base \citep{solaiman2019release} & own & -- \\
\hspace{1em}OpenAI-RoBERTa-large \citep{solaiman2019release} & own & -- \\
\hspace{1em}RADAR \citep{hu2023radar} & own & -- \\
\multicolumn{3}{l}{\emph{Commercial}} \\
\hspace{1em}Pangram 4 \citep{glickenhaus2026pangram4technicalreport} & AI if AI $+\frac{1}{2}$ AI-assisted $\geq 0.5$ & -- \\
\bottomrule
\end{tabular}
\end{center}
\end{table}

%% file: tables/appx/cost_build.tex
\begin{table}[!ht]
\caption{\textbf{One-time costs} (US dollars). API calls at batch rates, computed from our token logs; Tinker training metered by the provider. An invoice check on part of the corpus run implies the two corpus rows may total up to \$78,437.}
\label{tab:cost}
\begin{center}\small
\adjustbox{max width=\textwidth}{\begin{tabular}{llrr}
\toprule
Stage & Model & Scale & Cost \\
\midrule
Role discovery & GPT-5.6 Sol & 1,600 documents & \$116 \\
Outline extraction & Gemini 3.7 Flash (613k), 3.1 Pro (388k) & 1,001,145 documents & \$40,255 \\
Outline paraphrase & Gemini 3.1 Pro & 1,001,145 documents & \$26,173 \\
Training \model{} & Nemotron LoRA, Tinker & 804M tokens & \$354 \\
Training \twinmodel{} & Nemotron LoRA, Tinker & 1,613M tokens & \$710 \\
\midrule
Total & & & \$67,608 \\
\bottomrule
\end{tabular}}
\end{center}
\end{table}

%% file: tables/appx/cost_doc.tex
\begin{table}[!ht]
\caption{\textbf{Cost of scoring one new document} (US dollars, mean tokens per document at the corpus's mean length of 1,049 words). The gate and the extractor are Gemini 3.7 Flash; extraction tokens are from the 614,454 extractions of the corpus build, where \emph{input} includes the six in-context examples and \emph{output} includes the extractor's thinking. \emph{Standard} and \emph{Batch} price every token at the standard and batch API rates (batch is half of standard); \emph{Batch, cached} also bills the share of the extraction input that hit the prompt cache in our corpus run (88\%, the in-context examples shared by all documents of a format) at the cached rate, a tenth of the input rate. Scoring runs \model{} and \twinmodel{} (Nemotron LoRA) on Tinker, which bills the prompt only, twice per document (one forward pass per answer token); \twinmodel{} reads the document itself and needs neither the gate nor the extractor. Gemini rates as published for 2026 (they double on 1 January 2027); Tinker rates carry a limited-time 50\% discount on list.}
\label{tab:cost_doc}
\begin{center}\small\setlength{\tabcolsep}{4pt}
\adjustbox{max width=\textwidth}{\begin{tabular}{lrrrrr}
\toprule
Step & Input tokens & Output tokens & Standard & Batch & Batch, cached \\
\midrule
Format gate & 2,647 & 98 & \$0.0024 & \$0.0012 & \$0.0012 \\
Outline extraction & 45,522 & 6,765 & \$0.060 & \$0.030 & \$0.016 \\
\model{} scoring & 1,212 & 0 & \$0.0002 & \$0.0002 & \$0.0002 \\
\midrule
\textbf{\model{}, per document} & & & \textbf{\$0.062} & \textbf{\$0.031} & \textbf{\$0.018} \\
\model{}, per 1,000 documents & & & \$62 & \$31 & \$18 \\
\midrule
\twinmodel{}, per document & 2,732 & 0 & \$0.0005 & \$0.0005 & \$0.0005 \\
\bottomrule
\end{tabular}}
\end{center}
\end{table}

%% file: tables/main_q1.tex
\begingroup\footnotesize\setlength{\tabcolsep}{3pt}\setlength{\LTcapwidth}{\textwidth}\captionsetup{font=normalsize}
\begin{longtable}{@{}>{\raggedright\arraybackslash}p{0.35\textwidth}rrrrrr@{}}
\caption{\capQone}
\label{tab:q1} \\
\toprule
 & & \multicolumn{2}{c}{\model} &  &  &  \\
\cmidrule(lr){3-4}
Evaluation and arm & $n$ & outline & document & \twinmodel & Pangram 4 & EditLens \\
\midrule
\endfirsthead
\multicolumn{7}{@{}l}{\normalsize \autoref{tab:q1}, continued} \\
\toprule
 & & \multicolumn{2}{c}{\model} &  &  &  \\
\cmidrule(lr){3-4}
Evaluation and arm & $n$ & outline & document & \twinmodel & Pangram 4 & EditLens \\
\midrule
\endhead
\midrule
\multicolumn{7}{r@{}}{\emph{continued on the next page}} \\
\endfoot
\bottomrule
\endlastfoot
\textbf{Mean accuracy, the other three quadrants} & & \textbf{84.6} & \textbf{63.9} & \textbf{49.9} & \textbf{49.7} & \textbf{51.3} \\
\textbf{Mean accuracy, this quadrant (19 evaluations)} & & \textbf{99.4} & \textbf{99.7} & \textbf{99.3} & \textbf{99.7} & \textbf{99.8} \\
\midrule
\textbf{IntelLabs reviews}\hspace{0.5em}{\footnotesize human reviews} & 1,622 & 99.9\,/\,99.6 & 100.0 & 100.0 & 100.0 & 99.9 \\
\hspace{1em}{\scriptsize\color{muted} $<$\,500 words} & {\scriptsize\color{muted} 1,212} & {\scriptsize\color{muted} 99.9\,/\,99.5} & {\scriptsize\color{muted} 100.0} & {\scriptsize\color{muted} 100.0} & {\scriptsize\color{muted} 100.0} & {\scriptsize\color{muted} 99.9} \\
\hspace{1em}{\scriptsize\color{muted} $\geq$\,500 words} & {\scriptsize\color{muted} 410} & {\scriptsize\color{muted} 100.0\,/\,99.8} & {\scriptsize\color{muted} 100.0} & {\scriptsize\color{muted} 100.0} & {\scriptsize\color{muted} 100.0} & {\scriptsize\color{muted} 100.0} \\
\addlinespace[2pt]
\textbf{Saha et al.}\hspace{0.5em}{\footnotesize H: human review} & 2,621 & 100.0\,/\,99.7 & 100.0 & 100.0 & 100.0 & 100.0 \\
\addlinespace[2pt]
\textbf{DetectRL-X}\hspace{0.5em}{\footnotesize human} & 7,169 & 95.2\,/\,94.6 & 95.9 & 88.7 & 96.7 & 99.0 \\
\hspace{1em}{\scriptsize\color{muted} $<$\,500 words} & {\scriptsize\color{muted} 5,895} & {\scriptsize\color{muted} 94.3} & {\scriptsize\color{muted} 95.0} & {\scriptsize\color{muted} 86.4} & {\scriptsize\color{muted} 96.0} & {\scriptsize\color{muted} 98.8} \\
\hspace{1em}{\scriptsize\color{muted} $\geq$\,500 words} & {\scriptsize\color{muted} 1,274} & {\scriptsize\color{muted} 99.5} & {\scriptsize\color{muted} 99.8} & {\scriptsize\color{muted} 99.7} & {\scriptsize\color{muted} 100.0} & {\scriptsize\color{muted} 100.0} \\
\addlinespace[2pt]
\textbf{StoryScope}\hspace{0.5em}{\footnotesize human} & 1,345 & 99.9\,/\,99.3 & 100.0 & 100.0 & 100.0 & 100.0 \\
\addlinespace[2pt]
\textbf{\xLadder{}}\hspace{0.5em}{\footnotesize human pages} & 10,160 & 99.9\,/\,99.9 & 100.0 & 100.0 & 100.0 & 100.0 \\
\addlinespace[2pt]
\textbf{DetectionAI}\hspace{0.5em}{\footnotesize human} & 1,927 & 100.0\,/\,99.4 & 100.0 & 99.9 & 100.0 & 100.0 \\
\hspace{1em}{\scriptsize\color{muted} $<$\,500 words} & {\scriptsize\color{muted} 658} & {\scriptsize\color{muted} 100.0\,/\,98.5} & {\scriptsize\color{muted} 100.0} & {\scriptsize\color{muted} 99.8} & {\scriptsize\color{muted} 100.0} & {\scriptsize\color{muted} 100.0} \\
\hspace{1em}{\scriptsize\color{muted} $\geq$\,500 words} & {\scriptsize\color{muted} 1,269} & {\scriptsize\color{muted} 100.0\,/\,99.9} & {\scriptsize\color{muted} 100.0} & {\scriptsize\color{muted} 100.0} & {\scriptsize\color{muted} 100.0} & {\scriptsize\color{muted} 100.0} \\
\addlinespace[2pt]
\textbf{OpAI-Bench}\hspace{0.5em}{\footnotesize v0: the human draft} & 4,145 & 99.7\,/\,99.2 & 100.0 & 100.0 & 100.0 & 100.0 \\
\hspace{1em}{\scriptsize\color{muted} $<$\,500 words} & {\scriptsize\color{muted} 2,883} & {\scriptsize\color{muted} 99.6} & {\scriptsize\color{muted} 100.0} & {\scriptsize\color{muted} 100.0} & {\scriptsize\color{muted} 100.0} & {\scriptsize\color{muted} 100.0} \\
\hspace{1em}{\scriptsize\color{muted} $\geq$\,500 words} & {\scriptsize\color{muted} 1,262} & {\scriptsize\color{muted} 99.8} & {\scriptsize\color{muted} 100.0} & {\scriptsize\color{muted} 100.0} & {\scriptsize\color{muted} 100.0} & {\scriptsize\color{muted} 100.0} \\
\addlinespace[2pt]
\textbf{Sem-Detect}\hspace{0.5em}{\footnotesize human reviews} & 3,061 & 99.8\,/\,99.3 & 100.0 & 100.0 & 100.0 & 100.0 \\
\hspace{1em}{\scriptsize\color{muted} $<$\,500 words} & {\scriptsize\color{muted} 2,007} & {\scriptsize\color{muted} 99.8\,/\,99.1} & {\scriptsize\color{muted} 100.0} & {\scriptsize\color{muted} 100.0} & {\scriptsize\color{muted} 100.0} & {\scriptsize\color{muted} 100.0} \\
\hspace{1em}{\scriptsize\color{muted} $\geq$\,500 words} & {\scriptsize\color{muted} 1,054} & {\scriptsize\color{muted} 100.0\,/\,99.8} & {\scriptsize\color{muted} 100.0} & {\scriptsize\color{muted} 100.0} & {\scriptsize\color{muted} 100.0} & {\scriptsize\color{muted} 100.0} \\
\addlinespace[2pt]
\textbf{MELD}\hspace{0.5em}{\footnotesize human} & 3,918 & 99.8\,/\,99.6 & 100.0 & 100.0 & 100.0 & 100.0 \\
\hspace{1em}{\scriptsize\color{muted} $<$\,500 words} & {\scriptsize\color{muted} 3,493} & {\scriptsize\color{muted} 99.8} & {\scriptsize\color{muted} 100.0} & {\scriptsize\color{muted} 100.0} & {\scriptsize\color{muted} 100.0} & {\scriptsize\color{muted} 100.0} \\
\hspace{1em}{\scriptsize\color{muted} $\geq$\,500 words} & {\scriptsize\color{muted} 425} & {\scriptsize\color{muted} 100.0} & {\scriptsize\color{muted} 100.0} & {\scriptsize\color{muted} 100.0} & {\scriptsize\color{muted} 100.0} & {\scriptsize\color{muted} 100.0} \\
\addlinespace[2pt]
\textbf{ARB}\hspace{0.5em}{\footnotesize human} & 1,752 & 99.5\,/\,98.1 & 100.0 & 99.9 & 100.0 & 99.9 \\
\hspace{1em}{\scriptsize\color{muted} $<$\,500 words} & {\scriptsize\color{muted} 1,747} & {\scriptsize\color{muted} 99.5} & {\scriptsize\color{muted} 100.0} & {\scriptsize\color{muted} 99.9} & {\scriptsize\color{muted} 100.0} & {\scriptsize\color{muted} 99.9} \\
\hspace{1em}{\scriptsize\color{muted} $\geq$\,500 words} & {\scriptsize\color{muted} 5} & {\scriptsize\color{muted} 100.0} & {\scriptsize\color{muted} 100.0} & {\scriptsize\color{muted} 100.0} & {\scriptsize\color{muted} 100.0} & {\scriptsize\color{muted} 100.0} \\
\addlinespace[2pt]
\textbf{GEDE$^{\dagger}$}\hspace{0.5em}{\footnotesize human} & 915 & 99.9\,/\,99.7 & 100.0 & 100.0 & 100.0 & 100.0 \\
\addlinespace[2pt]
\textbf{HART}\hspace{0.5em}{\footnotesize human} & 3,922 & 99.4\,/\,98.3 & 99.8 & 99.6 & 99.9 & 99.7 \\
\hspace{1em}{\scriptsize\color{muted} $<$\,500 words} & {\scriptsize\color{muted} 3,466} & {\scriptsize\color{muted} 99.3} & {\scriptsize\color{muted} 99.9} & {\scriptsize\color{muted} 99.7} & {\scriptsize\color{muted} 99.9} & {\scriptsize\color{muted} 99.7} \\
\hspace{1em}{\scriptsize\color{muted} $\geq$\,500 words} & {\scriptsize\color{muted} 456} & {\scriptsize\color{muted} 99.6} & {\scriptsize\color{muted} 99.8} & {\scriptsize\color{muted} 99.6} & {\scriptsize\color{muted} 100.0} & {\scriptsize\color{muted} 99.6} \\
\addlinespace[2pt]
\textbf{Epoch imitation}\hspace{0.5em}{\footnotesize genuine passages} & 495 & 100.0\,/\,99.2 & 100.0 & 100.0 & 100.0 & 100.0 \\
\hspace{1em}{\scriptsize\color{muted} $<$\,500 words} & {\scriptsize\color{muted} 245} & {\scriptsize\color{muted} 100.0\,/\,98.4} & {\scriptsize\color{muted} 100.0} & {\scriptsize\color{muted} 100.0} & {\scriptsize\color{muted} 100.0} & {\scriptsize\color{muted} 100.0} \\
\hspace{1em}{\scriptsize\color{muted} $\geq$\,500 words} & {\scriptsize\color{muted} 250} & {\scriptsize\color{muted} 100.0\,/\,100.0} & {\scriptsize\color{muted} 100.0} & {\scriptsize\color{muted} 100.0} & {\scriptsize\color{muted} 100.0} & {\scriptsize\color{muted} 100.0} \\
\addlinespace[2pt]
\textbf{CoCoNUTS}\hspace{0.5em}{\footnotesize human review} & 1,194 & 100.0\,/\,99.9 & 100.0 & 100.0 & 100.0 & 100.0 \\
\hspace{1em}{\scriptsize\color{muted} $<$\,500 words} & {\scriptsize\color{muted} 735} & {\scriptsize\color{muted} 100.0\,/\,99.9} & {\scriptsize\color{muted} 100.0} & {\scriptsize\color{muted} 100.0} & {\scriptsize\color{muted} 100.0} & {\scriptsize\color{muted} 100.0} \\
\hspace{1em}{\scriptsize\color{muted} $\geq$\,500 words} & {\scriptsize\color{muted} 459} & {\scriptsize\color{muted} 100.0\,/\,100.0} & {\scriptsize\color{muted} 100.0} & {\scriptsize\color{muted} 100.0} & {\scriptsize\color{muted} 100.0} & {\scriptsize\color{muted} 100.0} \\
\addlinespace[2pt]
\textbf{ELLIPSE}\hspace{0.5em}{\footnotesize all essays} & 3,909 & 99.6\,/\,97.8 & 100.0 & 100.0 & 100.0 & 100.0 \\
\hspace{1em}{\scriptsize\color{muted} $<$\,500 words} & {\scriptsize\color{muted} 2,757} & {\scriptsize\color{muted} 99.6} & {\scriptsize\color{muted} 100.0} & {\scriptsize\color{muted} 100.0} & {\scriptsize\color{muted} 100.0} & {\scriptsize\color{muted} 100.0} \\
\hspace{1em}{\scriptsize\color{muted} $\geq$\,500 words} & {\scriptsize\color{muted} 1,152} & {\scriptsize\color{muted} 99.7} & {\scriptsize\color{muted} 100.0} & {\scriptsize\color{muted} 100.0} & {\scriptsize\color{muted} 100.0} & {\scriptsize\color{muted} 100.0} \\
\addlinespace[2pt]
\textbf{AI-Researcher}\hspace{0.5em}{\footnotesize expert reviews} & 504 & 97.4\,/\,87.9 & 97.8 & 97.8 & 97.6 & 96.8 \\
\hspace{1em}{\scriptsize\color{muted} $<$\,500 words} & {\scriptsize\color{muted} 476} & {\scriptsize\color{muted} 97.3} & {\scriptsize\color{muted} 97.7} & {\scriptsize\color{muted} 97.7} & {\scriptsize\color{muted} 97.7} & {\scriptsize\color{muted} 96.6} \\
\hspace{1em}{\scriptsize\color{muted} $\geq$\,500 words} & {\scriptsize\color{muted} 28} & {\scriptsize\color{muted} 100.0} & {\scriptsize\color{muted} 100.0} & {\scriptsize\color{muted} 100.0} & {\scriptsize\color{muted} 96.4} & {\scriptsize\color{muted} 100.0} \\
\addlinespace[2pt]
\textbf{AcademicIntegrity}\hspace{0.5em}{\footnotesize original} & 289 & 99.7\,/\,99.0 & 100.0 & 100.0 & 99.7 & 100.0 \\
\hspace{1em}{\scriptsize\color{muted} $<$\,500 words} & {\scriptsize\color{muted} 287} & {\scriptsize\color{muted} 99.7} & {\scriptsize\color{muted} 100.0} & {\scriptsize\color{muted} 100.0} & {\scriptsize\color{muted} 99.7} & {\scriptsize\color{muted} 100.0} \\
\hspace{1em}{\scriptsize\color{muted} $\geq$\,500 words} & {\scriptsize\color{muted} 2} & {\scriptsize\color{muted} 100.0} & {\scriptsize\color{muted} 100.0} & {\scriptsize\color{muted} 100.0} & {\scriptsize\color{muted} 100.0} & {\scriptsize\color{muted} 100.0} \\
\addlinespace[2pt]
\textbf{GEN}\hspace{0.5em}{\footnotesize human} & 3,055 & 99.5\,/\,98.7 & 100.0 & 100.0 & 100.0 & 100.0 \\
\hspace{1em}{\scriptsize\color{muted} $<$\,500 words} & {\scriptsize\color{muted} 2,027} & {\scriptsize\color{muted} 99.7} & {\scriptsize\color{muted} 100.0} & {\scriptsize\color{muted} 100.0} & {\scriptsize\color{muted} 100.0} & {\scriptsize\color{muted} 100.0} \\
\hspace{1em}{\scriptsize\color{muted} $\geq$\,500 words} & {\scriptsize\color{muted} 1,028} & {\scriptsize\color{muted} 99.1} & {\scriptsize\color{muted} 100.0} & {\scriptsize\color{muted} 100.0} & {\scriptsize\color{muted} 100.0} & {\scriptsize\color{muted} 100.0} \\
\addlinespace[2pt]
\textbf{C4}\hspace{0.5em}{\footnotesize web pages} & 9,998 & 100.0 & 100.0 & 100.0 & 100.0 & 100.0 \\
\addlinespace[2pt]
\end{longtable}
\endgroup

%% file: tables/main_q3.tex
\begingroup\footnotesize\setlength{\tabcolsep}{3pt}\setlength{\LTcapwidth}{\textwidth}\captionsetup{font=normalsize}
\begin{longtable}{@{}>{\raggedright\arraybackslash}p{0.33\textwidth}rrrrrr@{}}
\caption{\capQthree}
\label{tab:q3} \\
\toprule
 & & \multicolumn{2}{c}{\model} &  &  &  \\
\cmidrule(lr){3-4}
Evaluation and arm & $n$ & outline & document & \twinmodel & Pangram 4 & EditLens \\
\midrule
\endfirsthead
\multicolumn{7}{@{}l}{\normalsize \autoref{tab:q3}, continued} \\
\toprule
 & & \multicolumn{2}{c}{\model} &  &  &  \\
\cmidrule(lr){3-4}
Evaluation and arm & $n$ & outline & document & \twinmodel & Pangram 4 & EditLens \\
\midrule
\endhead
\midrule
\multicolumn{7}{r@{}}{\emph{continued on the next page}} \\
\endfoot
\bottomrule
\endlastfoot
\textbf{Mean accuracy, the other three quadrants} & & \textbf{87.4} & \textbf{65.1} & \textbf{50.0} & \textbf{50.5} & \textbf{57.3} \\
\textbf{Mean accuracy, this quadrant (19 evaluations)} & & \textbf{91.1} & \textbf{96.1} & \textbf{98.9} & \textbf{97.3} & \textbf{81.8} \\
\midrule
\textbf{IntelLabs reviews}\hspace{0.5em}{\footnotesize model reviews} & 8,334 & 73.8\,/\,85.3 & 84.7 & 98.6 & 94.7 & 94.5 \\
\hspace{1em}{\scriptsize\color{muted} $<$\,500 words} & {\scriptsize\color{muted} 6,575} & {\scriptsize\color{muted} 70.1\,/\,82.6} & {\scriptsize\color{muted} 82.1} & {\scriptsize\color{muted} 98.3} & {\scriptsize\color{muted} 93.8} & {\scriptsize\color{muted} 94.1} \\
\hspace{1em}{\scriptsize\color{muted} $\geq$\,500 words} & {\scriptsize\color{muted} 1,759} & {\scriptsize\color{muted} 87.8\,/\,95.7} & {\scriptsize\color{muted} 94.7} & {\scriptsize\color{muted} 100.0} & {\scriptsize\color{muted} 97.8} & {\scriptsize\color{muted} 96.1} \\
\addlinespace[2pt]
\multicolumn{7}{@{}l@{}}{\textbf{Saha et al.}\hspace{0.5em}{\footnotesize\color{muted} peer reviews, assistance levels}} \\
\hspace{1em}{\footnotesize AI-BP} & 4,920 & 64.8\,/\,77.7 & 82.9 & 100.0 & 97.8 & 88.3 \\
\hspace{2em}{\scriptsize\color{muted} $<$\,500 words} & {\scriptsize\color{muted} 1,744} & {\scriptsize\color{muted} 69.4\,/\,81.9} & {\scriptsize\color{muted} 81.5} & {\scriptsize\color{muted} 100.0} & {\scriptsize\color{muted} 97.9} & {\scriptsize\color{muted} 98.2} \\
\hspace{2em}{\scriptsize\color{muted} $\geq$\,500 words} & {\scriptsize\color{muted} 3,176} & {\scriptsize\color{muted} 62.2\,/\,75.5} & {\scriptsize\color{muted} 83.7} & {\scriptsize\color{muted} 100.0} & {\scriptsize\color{muted} 97.8} & {\scriptsize\color{muted} 82.8} \\
\hspace{1em}{\footnotesize AI-EP} & 4,530 & 70.9\,/\,81.7 & 84.8 & 100.0 & 98.5 & 91.5 \\
\hspace{2em}{\scriptsize\color{muted} $<$\,500 words} & {\scriptsize\color{muted} 1,688} & {\scriptsize\color{muted} 73.7\,/\,83.8} & {\scriptsize\color{muted} 83.2} & {\scriptsize\color{muted} 100.0} & {\scriptsize\color{muted} 98.3} & {\scriptsize\color{muted} 98.8} \\
\hspace{2em}{\scriptsize\color{muted} $\geq$\,500 words} & {\scriptsize\color{muted} 2,842} & {\scriptsize\color{muted} 69.3\,/\,80.5} & {\scriptsize\color{muted} 85.8} & {\scriptsize\color{muted} 100.0} & {\scriptsize\color{muted} 98.6} & {\scriptsize\color{muted} 87.2} \\
\hspace{1em}{\footnotesize AI-BP, humanized \textit{(attacked)}} & 200 & 86.0\,/\,93.0 & 87.0 & 99.0 & 78.5 & 0.0 \\
\addlinespace[2pt]
\multicolumn{7}{@{}l@{}}{\textbf{DetectRL-X}\hspace{0.5em}{\footnotesize\color{muted} English rows}} \\
\hspace{1em}{\footnotesize clean generations} & 7,451 & 87.3\,/\,83.4 & 98.6 & 100.0 & 99.7 & 92.9 \\
\hspace{2em}{\scriptsize\color{muted} $<$\,500 words} & {\scriptsize\color{muted} 5,984} & {\scriptsize\color{muted} 85.5} & {\scriptsize\color{muted} 98.2} & {\scriptsize\color{muted} 100.0} & {\scriptsize\color{muted} 99.7} & {\scriptsize\color{muted} 95.4} \\
\hspace{2em}{\scriptsize\color{muted} $\geq$\,500 words} & {\scriptsize\color{muted} 1,467} & {\scriptsize\color{muted} 94.9} & {\scriptsize\color{muted} 100.0} & {\scriptsize\color{muted} 100.0} & {\scriptsize\color{muted} 99.9} & {\scriptsize\color{muted} 83.0} \\
\hspace{1em}{\footnotesize eleven attacks, pooled \textit{(attacked)}} & 5,296 & 83.8\,/\,80.4 & 93.1 & 97.1 & 97.9 & 84.1 \\
\hspace{2em}{\scriptsize\color{muted} $<$\,500 words} & {\scriptsize\color{muted} 4,016} & {\scriptsize\color{muted} 80.8} & {\scriptsize\color{muted} 92.2} & {\scriptsize\color{muted} 96.3} & {\scriptsize\color{muted} 97.4} & {\scriptsize\color{muted} 83.8} \\
\hspace{2em}{\scriptsize\color{muted} $\geq$\,500 words} & {\scriptsize\color{muted} 1,280} & {\scriptsize\color{muted} 93.0} & {\scriptsize\color{muted} 95.7} & {\scriptsize\color{muted} 99.5} & {\scriptsize\color{muted} 99.5} & {\scriptsize\color{muted} 85.3} \\
\addlinespace[2pt]
\textbf{StoryScope}\hspace{0.5em}{\footnotesize five generators} & 6,902 & 97.0\,/\,98.5 & 99.8 & 99.8 & 100.0 & 21.6 \\
\hspace{1em}{\scriptsize\color{muted} $<$\,500 words} & {\scriptsize\color{muted} 29} & {\scriptsize\color{muted} 72.4\,/\,82.8} & {\scriptsize\color{muted} 89.7} & {\scriptsize\color{muted} 89.7} & {\scriptsize\color{muted} 100.0} & {\scriptsize\color{muted} 65.5} \\
\hspace{1em}{\scriptsize\color{muted} $\geq$\,500 words} & {\scriptsize\color{muted} 6,882} & {\scriptsize\color{muted} 97.1\,/\,98.5} & {\scriptsize\color{muted} 99.9} & {\scriptsize\color{muted} 99.8} & {\scriptsize\color{muted} 100.0} & {\scriptsize\color{muted} 21.5} \\
\addlinespace[2pt]
\textbf{\xLadder{}}\hspace{0.5em}{\footnotesize level 0: topic-only} & 10,152 & 95.3\,/\,95.0 & 81.4 & 82.5 & 85.4 & 15.4 \\
\hspace{1em}{\scriptsize\color{muted} $<$\,500 words} & {\scriptsize\color{muted} 940} & {\scriptsize\color{muted} 96.3} & {\scriptsize\color{muted} 86.4} & {\scriptsize\color{muted} 87.7} & {\scriptsize\color{muted} 90.9} & {\scriptsize\color{muted} 27.2} \\
\hspace{1em}{\scriptsize\color{muted} $\geq$\,500 words} & {\scriptsize\color{muted} 7,369} & {\scriptsize\color{muted} 95.3} & {\scriptsize\color{muted} 79.9} & {\scriptsize\color{muted} 81.0} & {\scriptsize\color{muted} 84.5} & {\scriptsize\color{muted} 14.6} \\
\addlinespace[2pt]
\multicolumn{7}{@{}l@{}}{\textbf{DetectionAI}\hspace{0.5em}{\footnotesize\color{muted} essays, with a humanizer}} \\
\hspace{1em}{\footnotesize four generators} & 7,707 & 90.3\,/\,93.7 & 93.7 & 99.7 & 99.8 & 94.7 \\
\hspace{2em}{\scriptsize\color{muted} $<$\,500 words} & {\scriptsize\color{muted} 2,634} & {\scriptsize\color{muted} 76.5\,/\,83.8} & {\scriptsize\color{muted} 81.5} & {\scriptsize\color{muted} 99.2} & {\scriptsize\color{muted} 99.5} & {\scriptsize\color{muted} 92.0} \\
\hspace{2em}{\scriptsize\color{muted} $\geq$\,500 words} & {\scriptsize\color{muted} 5,073} & {\scriptsize\color{muted} 97.5\,/\,98.8} & {\scriptsize\color{muted} 100.0} & {\scriptsize\color{muted} 100.0} & {\scriptsize\color{muted} 100.0} & {\scriptsize\color{muted} 96.0} \\
\hspace{1em}{\footnotesize humanized \textit{(attacked)}} & 7,706 & 57.9\,/\,69.3 & 32.4 & 64.1 & 94.2 & 4.9 \\
\hspace{2em}{\scriptsize\color{muted} $<$\,500 words} & {\scriptsize\color{muted} 2,917} & {\scriptsize\color{muted} 47.4\,/\,56.7} & {\scriptsize\color{muted} 27.0} & {\scriptsize\color{muted} 64.5} & {\scriptsize\color{muted} 84.9} & {\scriptsize\color{muted} 11.0} \\
\hspace{2em}{\scriptsize\color{muted} $\geq$\,500 words} & {\scriptsize\color{muted} 4,789} & {\scriptsize\color{muted} 64.2\,/\,77.0} & {\scriptsize\color{muted} 35.7} & {\scriptsize\color{muted} 63.8} & {\scriptsize\color{muted} 99.9} & {\scriptsize\color{muted} 1.2} \\
\addlinespace[2pt]
\textbf{Sem-Detect}\hspace{0.5em}{\footnotesize model reviews} & 6,767 & 71.4\,/\,85.7 & 95.9 & 100.0 & 75.7 & 65.2 \\
\hspace{1em}{\scriptsize\color{muted} $<$\,500 words} & {\scriptsize\color{muted} 1,488} & {\scriptsize\color{muted} 91.7\,/\,97.3} & {\scriptsize\color{muted} 98.4} & {\scriptsize\color{muted} 100.0} & {\scriptsize\color{muted} 80.0} & {\scriptsize\color{muted} 98.4} \\
\hspace{1em}{\scriptsize\color{muted} $\geq$\,500 words} & {\scriptsize\color{muted} 5,279} & {\scriptsize\color{muted} 65.7\,/\,82.4} & {\scriptsize\color{muted} 95.2} & {\scriptsize\color{muted} 100.0} & {\scriptsize\color{muted} 74.6} & {\scriptsize\color{muted} 55.9} \\
\addlinespace[2pt]
\multicolumn{7}{@{}l@{}}{\textbf{MELD}\hspace{0.5em}{\footnotesize\color{muted} long-form, four generators}} \\
\hspace{1em}{\footnotesize clean generations} & 1,546 & 88.1\,/\,88.5 & 97.0 & 99.9 & 100.0 & 98.0 \\
\hspace{2em}{\scriptsize\color{muted} $<$\,500 words} & {\scriptsize\color{muted} 884} & {\scriptsize\color{muted} 85.6} & {\scriptsize\color{muted} 96.0} & {\scriptsize\color{muted} 99.8} & {\scriptsize\color{muted} 100.0} & {\scriptsize\color{muted} 97.4} \\
\hspace{2em}{\scriptsize\color{muted} $\geq$\,500 words} & {\scriptsize\color{muted} 662} & {\scriptsize\color{muted} 91.4} & {\scriptsize\color{muted} 98.3} & {\scriptsize\color{muted} 100.0} & {\scriptsize\color{muted} 100.0} & {\scriptsize\color{muted} 98.8} \\
\hspace{1em}{\footnotesize six attacks, pooled \textit{(attacked)}} & 9,249 & 84.2\,/\,84.9 & 71.6 & 72.9 & 100.0 & 53.1 \\
\hspace{2em}{\scriptsize\color{muted} $<$\,500 words} & {\scriptsize\color{muted} 5,290} & {\scriptsize\color{muted} 79.9} & {\scriptsize\color{muted} 68.8} & {\scriptsize\color{muted} 70.5} & {\scriptsize\color{muted} 100.0} & {\scriptsize\color{muted} 52.7} \\
\hspace{2em}{\scriptsize\color{muted} $\geq$\,500 words} & {\scriptsize\color{muted} 3,959} & {\scriptsize\color{muted} 89.9} & {\scriptsize\color{muted} 75.5} & {\scriptsize\color{muted} 76.2} & {\scriptsize\color{muted} 100.0} & {\scriptsize\color{muted} 53.6} \\
\addlinespace[2pt]
\textbf{LAMP$^{\dagger}$}\hspace{0.5em}{\footnotesize model draft} & 1,057 & 98.3\,/\,99.1 & 100.0 & 100.0 & 100.0 & 99.8 \\
\addlinespace[2pt]
\multicolumn{7}{@{}l@{}}{\textbf{ARB$^{\dagger}$}\hspace{0.5em}{\footnotesize\color{muted} matched rewriting benchmark}} \\
\hspace{1em}{\footnotesize free generation} & 6,904 & 97.9\,/\,96.8 & 97.3 & 100.0 & 99.9 & 98.4 \\
\hspace{1em}{\footnotesize generation, rewritten \textit{(attacked)}} & 7,043 & 97.5\,/\,96.1 & 96.5 & 99.9 & 99.9 & 98.1 \\
\addlinespace[2pt]
\multicolumn{7}{@{}l@{}}{\textbf{GEDE$^{\dagger}$}\hspace{0.5em}{\footnotesize\color{muted} student essays, contribution levels}} \\
\hspace{1em}{\footnotesize from the assignment} & 1,870 & 99.4\,/\,97.3 & 99.9 & 100.0 & 100.0 & 100.0 \\
\hspace{1em}{\footnotesize paraphrase attacks \textit{(attacked)}} & 3,604 & 88.7\,/\,81.4 & 76.6 & 98.6 & 97.6 & 60.7 \\
\addlinespace[2pt]
\multicolumn{7}{@{}l@{}}{\textbf{HART}\hspace{0.5em}{\footnotesize\color{muted} English rows}} \\
\hspace{1em}{\footnotesize clean generations} & 3,953 & 96.5\,/\,95.4 & 97.7 & 100.0 & 100.0 & 96.1 \\
\hspace{2em}{\scriptsize\color{muted} $<$\,500 words} & {\scriptsize\color{muted} 3,444} & {\scriptsize\color{muted} 96.3} & {\scriptsize\color{muted} 97.4} & {\scriptsize\color{muted} 100.0} & {\scriptsize\color{muted} 100.0} & {\scriptsize\color{muted} 96.7} \\
\hspace{2em}{\scriptsize\color{muted} $\geq$\,500 words} & {\scriptsize\color{muted} 509} & {\scriptsize\color{muted} 98.0} & {\scriptsize\color{muted} 99.6} & {\scriptsize\color{muted} 100.0} & {\scriptsize\color{muted} 100.0} & {\scriptsize\color{muted} 91.6} \\
\hspace{1em}{\footnotesize humanized \textit{(attacked)}} & 3,592 & 84.6\,/\,84.3 & 82.0 & 93.5 & 94.9 & 71.0 \\
\hspace{2em}{\scriptsize\color{muted} $<$\,500 words} & {\scriptsize\color{muted} 3,165} & {\scriptsize\color{muted} 82.8} & {\scriptsize\color{muted} 80.1} & {\scriptsize\color{muted} 92.7} & {\scriptsize\color{muted} 94.2} & {\scriptsize\color{muted} 71.6} \\
\hspace{2em}{\scriptsize\color{muted} $\geq$\,500 words} & {\scriptsize\color{muted} 427} & {\scriptsize\color{muted} 97.4} & {\scriptsize\color{muted} 96.7} & {\scriptsize\color{muted} 100.0} & {\scriptsize\color{muted} 100.0} & {\scriptsize\color{muted} 66.5} \\
\addlinespace[2pt]
\textbf{Epoch imitation}\hspace{0.5em}{\footnotesize model, topic prompt} & 297 & 97.6\,/\,98.7 & 100.0 & 100.0 & 100.0 & 93.9 \\
\hspace{1em}{\scriptsize\color{muted} $<$\,500 words} & {\scriptsize\color{muted} 120} & {\scriptsize\color{muted} 95.8\,/\,97.5} & {\scriptsize\color{muted} 100.0} & {\scriptsize\color{muted} 100.0} & {\scriptsize\color{muted} 100.0} & {\scriptsize\color{muted} 99.2} \\
\hspace{1em}{\scriptsize\color{muted} $\geq$\,500 words} & {\scriptsize\color{muted} 177} & {\scriptsize\color{muted} 98.9\,/\,99.4} & {\scriptsize\color{muted} 100.0} & {\scriptsize\color{muted} 100.0} & {\scriptsize\color{muted} 100.0} & {\scriptsize\color{muted} 90.4} \\
\addlinespace[2pt]
\textbf{CoCoNUTS}\hspace{0.5em}{\footnotesize model-generated} & 1,400 & 92.1\,/\,97.1 & 98.9 & 100.0 & 96.3 & 99.0 \\
\hspace{1em}{\scriptsize\color{muted} $<$\,500 words} & {\scriptsize\color{muted} 728} & {\scriptsize\color{muted} 93.4\,/\,98.2} & {\scriptsize\color{muted} 98.9} & {\scriptsize\color{muted} 100.0} & {\scriptsize\color{muted} 96.3} & {\scriptsize\color{muted} 99.9} \\
\hspace{1em}{\scriptsize\color{muted} $\geq$\,500 words} & {\scriptsize\color{muted} 672} & {\scriptsize\color{muted} 90.6\,/\,96.0} & {\scriptsize\color{muted} 98.8} & {\scriptsize\color{muted} 100.0} & {\scriptsize\color{muted} 96.3} & {\scriptsize\color{muted} 98.1} \\
\addlinespace[2pt]
\textbf{AI-Researcher}\hspace{0.5em}{\footnotesize proposals from model ideas} & 97 & 95.9\,/\,100.0 & 100.0 & 100.0 & 100.0 & 93.8 \\
\addlinespace[2pt]
\textbf{GEN}\hspace{0.5em}{\footnotesize model-written} & 4,046 & 95.8\,/\,96.5 & 98.4 & 99.7 & 99.8 & 70.9 \\
\hspace{1em}{\scriptsize\color{muted} $<$\,500 words} & {\scriptsize\color{muted} 314} & {\scriptsize\color{muted} 81.2} & {\scriptsize\color{muted} 83.1} & {\scriptsize\color{muted} 96.2} & {\scriptsize\color{muted} 98.1} & {\scriptsize\color{muted} 49.7} \\
\hspace{1em}{\scriptsize\color{muted} $\geq$\,500 words} & {\scriptsize\color{muted} 3,732} & {\scriptsize\color{muted} 97.0} & {\scriptsize\color{muted} 99.7} & {\scriptsize\color{muted} 99.9} & {\scriptsize\color{muted} 100.0} & {\scriptsize\color{muted} 72.7} \\
\addlinespace[2pt]
\textbf{\aiipset{}}\hspace{0.5em}{\footnotesize model originals} & 50 & 96.0 & 100.0 & 100.0 & 100.0 & 90.0 \\
\addlinespace[2pt]
\textbf{\collabLadder{}, human sources}\hspace{0.5em}{\footnotesize levels 0--1: topic-only or genre, audience and tone} & 474 & 93.0\,/\,93.5 & 98.5 & 99.4 & 100.0 & 63.9 \\
\addlinespace[2pt]
\textbf{\collabLadder{}, AI sources}\hspace{0.5em}{\footnotesize levels 0--5} & 1,578 & 98.3\,/\,98.6 & 99.9 & 100.0 & 99.9 & 76.2 \\
\addlinespace[2pt]
\end{longtable}
\endgroup

%% file: tables/main_q4.tex
\begingroup\footnotesize\setlength{\tabcolsep}{3pt}\setlength{\LTcapwidth}{\textwidth}\captionsetup{font=normalsize}
\begin{longtable}{@{}>{\raggedright\arraybackslash}p{0.35\textwidth}rrrrrr@{}}
\caption{\capQfour}
\label{tab:q4} \\
\toprule
 & & \multicolumn{2}{c}{\model} &  &  &  \\
\cmidrule(lr){3-4}
Evaluation and arm & $n$ & outline & document & \twinmodel & Pangram 4 & EditLens \\
\midrule
\endfirsthead
\multicolumn{7}{@{}l}{\normalsize \autoref{tab:q4}, continued} \\
\toprule
 & & \multicolumn{2}{c}{\model} &  &  &  \\
\cmidrule(lr){3-4}
Evaluation and arm & $n$ & outline & document & \twinmodel & Pangram 4 & EditLens \\
\midrule
\endhead
\midrule
\multicolumn{7}{r@{}}{\emph{continued on the next page}} \\
\endfoot
\bottomrule
\endlastfoot
\textbf{Mean accuracy, the other three quadrants} & & \textbf{86.2} & \textbf{67.2} & \textbf{66.1} & \textbf{68.3} & \textbf{60.5} \\
\textbf{Mean accuracy, this quadrant (12 evaluations)} & & \textbf{94.7} & \textbf{89.6} & \textbf{50.7} & \textbf{43.8} & \textbf{72.1} \\
\midrule
\multicolumn{7}{@{}l@{}}{\textbf{Saha et al.}\hspace{0.5em}{\footnotesize\color{muted} peer reviews, assistance levels}} \\
\hspace{1em}{\footnotesize H-AI: human review, polished} & 15,118 & 99.4\,/\,97.7 & 99.8 & 52.1 & 41.3 & 62.1 \\
\hspace{2em}{\scriptsize\color{muted} $<$\,500 words} & {\scriptsize\color{muted} 13,284} & {\scriptsize\color{muted} 99.3\,/\,97.6} & {\scriptsize\color{muted} 99.8} & {\scriptsize\color{muted} 47.9} & {\scriptsize\color{muted} 36.7} & {\scriptsize\color{muted} 60.0} \\
\hspace{2em}{\scriptsize\color{muted} $\geq$\,500 words} & {\scriptsize\color{muted} 1,834} & {\scriptsize\color{muted} 99.5\,/\,98.5} & {\scriptsize\color{muted} 99.8} & {\scriptsize\color{muted} 82.6} & {\scriptsize\color{muted} 74.4} & {\scriptsize\color{muted} 77.3} \\
\hspace{1em}{\footnotesize H-AI, humanized \textit{(humanized)}} & 718 & 98.5\,/\,96.2 & 98.9 & 31.1 & 28.3 & 99.4 \\
\addlinespace[2pt]
\textbf{\xLadder{}}\hspace{0.5em}{\footnotesize level 5: the person's outline} & 10,116 & 99.3\,/\,99.0 & 96.6 & 55.5 & 38.1 & 98.4 \\
\hspace{1em}{\scriptsize\color{muted} $<$\,500 words} & {\scriptsize\color{muted} 446} & {\scriptsize\color{muted} 99.8} & {\scriptsize\color{muted} 99.6} & {\scriptsize\color{muted} 62.6} & {\scriptsize\color{muted} 36.5} & {\scriptsize\color{muted} 92.8} \\
\hspace{1em}{\scriptsize\color{muted} $\geq$\,500 words} & {\scriptsize\color{muted} 7,830} & {\scriptsize\color{muted} 99.2} & {\scriptsize\color{muted} 97.1} & {\scriptsize\color{muted} 57.9} & {\scriptsize\color{muted} 42.6} & {\scriptsize\color{muted} 98.4} \\
\addlinespace[2pt]
\textbf{OpAI-Bench}\hspace{0.5em}{\footnotesize v1--v8: model-rewritten} & 33,163 & 99.6\,/\,98.8 & 100.0 & 87.9 & 83.8 & 99.1 \\
\hspace{1em}{\scriptsize\color{muted} $<$\,500 words} & {\scriptsize\color{muted} 22,303} & {\scriptsize\color{muted} 99.5} & {\scriptsize\color{muted} 100.0} & {\scriptsize\color{muted} 87.7} & {\scriptsize\color{muted} 80.4} & {\scriptsize\color{muted} 98.9} \\
\hspace{1em}{\scriptsize\color{muted} $\geq$\,500 words} & {\scriptsize\color{muted} 10,860} & {\scriptsize\color{muted} 99.7} & {\scriptsize\color{muted} 100.0} & {\scriptsize\color{muted} 88.2} & {\scriptsize\color{muted} 90.8} & {\scriptsize\color{muted} 99.5} \\
\addlinespace[2pt]
\textbf{Sem-Detect}\hspace{0.5em}{\footnotesize human review, rewritten} & 12,318 & 91.8\,/\,84.8 & 88.2 & 35.3 & 22.3 & 34.0 \\
\hspace{1em}{\scriptsize\color{muted} $<$\,500 words} & {\scriptsize\color{muted} 7,903} & {\scriptsize\color{muted} 92.3\,/\,86.0} & {\scriptsize\color{muted} 91.4} & {\scriptsize\color{muted} 31.0} & {\scriptsize\color{muted} 18.3} & {\scriptsize\color{muted} 29.2} \\
\hspace{1em}{\scriptsize\color{muted} $\geq$\,500 words} & {\scriptsize\color{muted} 4,415} & {\scriptsize\color{muted} 90.8\,/\,82.6} & {\scriptsize\color{muted} 82.4} & {\scriptsize\color{muted} 43.2} & {\scriptsize\color{muted} 29.4} & {\scriptsize\color{muted} 42.6} \\
\addlinespace[2pt]
\textbf{ARB$^{\dagger}$}\hspace{0.5em}{\footnotesize human article, rewritten} & 7,027 & 98.1\,/\,95.2 & 99.5 & 59.7 & 46.7 & 96.7 \\
\addlinespace[2pt]
\textbf{GEDE$^{\dagger}$}\hspace{0.5em}{\footnotesize essay improved or rewritten} & 3,648 & 98.4\,/\,97.6 & 98.9 & 38.5 & 25.5 & 71.4 \\
\addlinespace[2pt]
\textbf{HART}\hspace{0.5em}{\footnotesize human text, rephrased} & 3,904 & 95.1\,/\,92.9 & 95.1 & 55.5 & 57.2 & 79.9 \\
\hspace{1em}{\scriptsize\color{muted} $<$\,500 words} & {\scriptsize\color{muted} 3,510} & {\scriptsize\color{muted} 94.7} & {\scriptsize\color{muted} 94.7} & {\scriptsize\color{muted} 52.4} & {\scriptsize\color{muted} 55.0} & {\scriptsize\color{muted} 78.4} \\
\hspace{1em}{\scriptsize\color{muted} $\geq$\,500 words} & {\scriptsize\color{muted} 394} & {\scriptsize\color{muted} 98.7} & {\scriptsize\color{muted} 98.7} & {\scriptsize\color{muted} 83.5} & {\scriptsize\color{muted} 76.4} & {\scriptsize\color{muted} 93.4} \\
\addlinespace[2pt]
\multicolumn{7}{@{}l@{}}{\textbf{CoCoNUTS}\hspace{0.5em}{\footnotesize\color{muted} peer reviews, collaboration modes}} \\
\hspace{1em}{\footnotesize human review, polished} & 1,497 & 87.8\,/\,80.9 & 88.0 & 39.0 & 38.8 & 37.1 \\
\hspace{2em}{\scriptsize\color{muted} $<$\,500 words} & {\scriptsize\color{muted} 1,153} & {\scriptsize\color{muted} 86.4\,/\,78.9} & {\scriptsize\color{muted} 86.4} & {\scriptsize\color{muted} 31.5} & {\scriptsize\color{muted} 31.2} & {\scriptsize\color{muted} 30.3} \\
\hspace{2em}{\scriptsize\color{muted} $\geq$\,500 words} & {\scriptsize\color{muted} 344} & {\scriptsize\color{muted} 92.7\,/\,87.5} & {\scriptsize\color{muted} 93.3} & {\scriptsize\color{muted} 64.2} & {\scriptsize\color{muted} 64.2} & {\scriptsize\color{muted} 59.9} \\
\hspace{1em}{\footnotesize human review, translated} & 1,493 & 99.9\,/\,99.5 & 100.0 & 99.5 & 99.9 & 98.7 \\
\hspace{2em}{\scriptsize\color{muted} $<$\,500 words} & {\scriptsize\color{muted} 1,139} & {\scriptsize\color{muted} 99.9\,/\,99.4} & {\scriptsize\color{muted} 100.0} & {\scriptsize\color{muted} 99.4} & {\scriptsize\color{muted} 99.8} & {\scriptsize\color{muted} 98.2} \\
\hspace{2em}{\scriptsize\color{muted} $\geq$\,500 words} & {\scriptsize\color{muted} 354} & {\scriptsize\color{muted} 100.0\,/\,100.0} & {\scriptsize\color{muted} 100.0} & {\scriptsize\color{muted} 100.0} & {\scriptsize\color{muted} 100.0} & {\scriptsize\color{muted} 100.0} \\
\addlinespace[2pt]
\textbf{AI-Researcher}\hspace{0.5em}{\footnotesize proposals from expert ideas} & 48 & 89.6\,/\,81.2 & 87.5 & 85.4 & 87.5 & 75.0 \\
\hspace{1em}{\scriptsize\color{muted} $<$\,500 words} & {\scriptsize\color{muted} 1} & {\scriptsize\color{muted} 100.0} & {\scriptsize\color{muted} 100.0} & {\scriptsize\color{muted} 100.0} & {\scriptsize\color{muted} 100.0} & {\scriptsize\color{muted} 0.0} \\
\hspace{1em}{\scriptsize\color{muted} $\geq$\,500 words} & {\scriptsize\color{muted} 47} & {\scriptsize\color{muted} 89.4} & {\scriptsize\color{muted} 87.2} & {\scriptsize\color{muted} 85.1} & {\scriptsize\color{muted} 87.2} & {\scriptsize\color{muted} 76.6} \\
\addlinespace[2pt]
\multicolumn{7}{@{}l@{}}{\textbf{AcademicIntegrity}\hspace{0.5em}{\footnotesize\color{muted} pre-2016 abstracts}} \\
\hspace{1em}{\footnotesize refined or regenerated} & 867 & 91.8\,/\,83.9 & 96.3 & 16.1 & 3.6 & 30.6 \\
\hspace{1em}{\footnotesize humanized \textit{(humanized)}} & 1,156 & 99.1\,/\,97.7 & 99.9 & 99.3 & 91.0 & 99.6 \\
\hspace{2em}{\scriptsize\color{muted} $<$\,500 words} & {\scriptsize\color{muted} 1,145} & {\scriptsize\color{muted} 99.1} & {\scriptsize\color{muted} 99.9} & {\scriptsize\color{muted} 99.3} & {\scriptsize\color{muted} 91.0} & {\scriptsize\color{muted} 99.6} \\
\hspace{2em}{\scriptsize\color{muted} $\geq$\,500 words} & {\scriptsize\color{muted} 11} & {\scriptsize\color{muted} 100.0} & {\scriptsize\color{muted} 100.0} & {\scriptsize\color{muted} 100.0} & {\scriptsize\color{muted} 90.9} & {\scriptsize\color{muted} 100.0} \\
\addlinespace[2pt]
\textbf{GEN}\hspace{0.5em}{\footnotesize model-edited, seven categories} & 13,964 & 86.0\,/\,83.0 & 73.7 & 40.1 & 43.2 & 66.3 \\
\hspace{1em}{\scriptsize\color{muted} $<$\,500 words} & {\scriptsize\color{muted} 10,420} & {\scriptsize\color{muted} 86.5} & {\scriptsize\color{muted} 76.3} & {\scriptsize\color{muted} 40.4} & {\scriptsize\color{muted} 37.9} & {\scriptsize\color{muted} 64.9} \\
\hspace{1em}{\scriptsize\color{muted} $\geq$\,500 words} & {\scriptsize\color{muted} 3,666} & {\scriptsize\color{muted} 84.4} & {\scriptsize\color{muted} 66.1} & {\scriptsize\color{muted} 39.2} & {\scriptsize\color{muted} 58.2} & {\scriptsize\color{muted} 70.3} \\
\addlinespace[2pt]
\textbf{\collabLadder{}, human sources}\hspace{0.5em}{\footnotesize level 5: the person's outline} & 237 & 93.2\,/\,92.0 & 45.6 & 13.5 & 7.6 & 84.0 \\
\addlinespace[2pt]
\end{longtable}
\endgroup

%% file: tables/appendix/q2_full.tex
\begin{table}[h]
\caption{\textbf{Model ideas, human prose: accuracy (\%).} \twicetold{}: 50 stories people wrote from scratch from the outline of a model's story (\S\ref{sec:experiments}), so every flag is a correct detection. Human-edited model drafts (LAMP, HART's human-edited arm) keep most of the model's prose and are excluded. The top two rows give each detector's mean accuracy in this quadrant and in the other three, so no detector can look strong by always predicting one label. Outline column: global 1\% cut\,/\,per-format cut. All arms are in \S\ref{app:evals}.}
\label{tab:q2}
\begin{center}\footnotesize
\setlength{\tabcolsep}{3pt}
\adjustbox{max width=\textwidth}{\begin{tabular}{@{}lrrrrrr@{}}
\toprule
 & & \multicolumn{2}{c}{\model} &  &  &  \\
\cmidrule(lr){3-4}
Evaluation and arm & $n$ & outline & document & \twinmodel & Pangram 4 & EditLens \\
\midrule
\textbf{Mean accuracy, the other three quadrants} & & \textbf{95.1} & \textbf{95.1} & \textbf{83.0} & \textbf{80.3} & \textbf{84.6} \\
\textbf{Mean accuracy, this quadrant (1 evaluation)} & & \textbf{68.0} & \textbf{6.0} & \textbf{0.0} & \textbf{8.0} & \textbf{0.0} \\
\midrule
\textbf{\aiipset{}}\hspace{0.5em}{\footnotesize rewrites} & 50 & 68.0\,/\,80.0 & 6.0 & 0.0 & 8.0 & 0.0 \\
\addlinespace[2pt]
\bottomrule
\end{tabular}}
\end{center}
\end{table}

%% file: tables/appendix/consolidated_all.tex
\begin{table}[t]
\caption{\textbf{Every detector, every quadrant.} Each cell is the mean over evaluations of the pooled rate at the global 1\% cut (``own'': the detector's own threshold). Q1: human ideas, human prose; Q2: model ideas, human prose; Q3: model ideas, model prose (attacked arms separate); Q4: human ideas, model prose. A superscript gives the number of evaluations when a detector covers fewer than the column. An idea detector flags Q2 and Q3 but not Q1 or Q4; a prose detector flags Q3 and Q4.}
\label{tab:cons}
\begin{center}\small
\adjustbox{max width=\textwidth}{\begin{tabular}{lrrrrr}
\toprule
Detector & \shortstack{Q1\\FPR, 19 evals} & \shortstack{Q4\\FPR, 12 evals} & \shortstack{Q3\\TPR, 19 evals} & \shortstack{Q3 attacked\\TPR, 8 evals} & \shortstack{Q2\\TPR, 1 eval} \\
\midrule
\multicolumn{6}{l}{\emph{Ours}} \\
\model{} (outline) & 0.6 & 5.3 & 91.1 & 84.5 & 68.0 \\
\model{} (document) & 0.3 & 10.4 & 96.1 & 79.8 & 6.0 \\
\twinmodel{} & 0.7 & 49.3 & 98.9 & 90.6 & 0.0 \\
\modelModernBERT{} (outline) & 1.5 & 6.5 & 79.1 & 73.9 & 28.0 \\
\modelModernBERT{} (document) & 0.2$^{17}$ & 9.4$^{11}$ & 89.1$^{17}$ & 55.5 & -- \\
\twinmodelModernBERT{} & 1.0$^{18}$ & 38.1$^{11}$ & 99.1$^{18}$ & 76.2 & 0.0 \\
\modelQwen{} (outline) & 0.9$^{18}$ & 4.7$^{11}$ & 84.5$^{16}$ & 76.8 & -- \\
\modelQwen{} (document) & 0.3$^{18}$ & 5.5$^{11}$ & 89.9$^{16}$ & 64.9 & -- \\
\modelLR{} & 1.1$^{18}$ & 1.9$^{11}$ & 22.1$^{16}$ & 27.4 & -- \\
\multicolumn{6}{l}{\emph{Baselines}} \\
Pangram 4 (own threshold) & 0.3 & 56.2 & 97.3 & 94.9 & 8.0 \\
EditLens-Llama-3B & 0.2 & 27.9 & 81.8 & 58.3 & 0.0 \\
Binoculars (own threshold) & 1.0 & 5.8 & 40.3 & 29.1 & 0.0 \\
EditLens-RoBERTa & 0.2 & 13.3 & 59.3 & 45.4 & 0.0 \\
MELD (own threshold) & 8.2 & 77.6 & 98.2 & 83.2 & 22.0 \\
Desklib-academic (own threshold) & 38.8 & 89.1 & 95.1 & 94.7 & 84.0 \\
Desklib (own threshold) & 14.6 & 81.6 & 88.8 & 85.2 & 40.0 \\
Entropy & 2.8 & 3.5 & 2.5 & 5.2 & 0.0 \\
Fast-DetectGPT (own threshold) & 23.0 & 50.5 & 75.3 & 67.5 & 10.0 \\
Likelihood & 2.6 & 3.3 & 4.4 & 5.6 & 0.0 \\
Log-rank & 2.6 & 3.2 & 5.3 & 6.2 & 0.0 \\
LRR & 2.2 & 2.3 & 9.1 & 7.4 & 0.0 \\
MAGE (own threshold) & 27.0 & 59.3 & 62.1 & 62.5 & 14.0 \\
OpenAI-RoBERTa-base (own threshold) & 4.4 & 7.5 & 11.1 & 17.4 & 0.0 \\
OpenAI-RoBERTa-large (own threshold) & 3.5 & 5.5 & 6.0 & 11.2 & 0.0 \\
RADAR (own threshold) & 17.4 & 28.8 & 34.7 & 40.6 & 8.0 \\
Rank & 3.4 & 5.4 & 8.2 & 9.8 & 0.0 \\
\bottomrule
\end{tabular}}
\end{center}
\end{table}

%% file: tables/appx/opai.tex
\begin{table}[h]
\caption{\textbf{OpAI-Bench}: main rows.}
\label{tab:opai:0}
\begin{center}\footnotesize
\setlength{\tabcolsep}{2.5pt}
\adjustbox{max width=\textwidth}{\begin{tabular}{@{}llrrrrrr@{}}
\toprule
 &  &  & \multicolumn{2}{c}{\model} &  &  &  \\
\cmidrule(lr){4-5}
Arm & metric & n & outline & document & \twinmodel & EditLens & Pangram 4 \\
\midrule
v0: the human seed & FPR & 4,145 & 0.3 & 0.0 & 0.0 & 0.0 & 0.0 \\
\textbf{v1 to v8: AI-edited versions} &  &  &  &  &  &  &  \\
\hspace{1em}mean over 8 versions & FPR & 33,163 & 0.4 & 0.0 & 12.1 & 0.9 & 16.2 \\
\hspace{1em}median over 8 versions & FPR &  & 0.4 & 0.0 & 10.1 & 0.3 & 4.7 \\
\hspace{1em}min over 8 versions & FPR &  & 0.2 (v1) & 0.0 (v1) & 0.0 (v1) & 0.0 (v2) & 0.0 (v1) \\
\hspace{1em}max over 8 versions & FPR &  & 0.6 (v5) & 0.0 (v5) & 26.3 (v6) & 3.7 (v6) & 48.5 (v8) \\
\bottomrule
\end{tabular}}
\end{center}
\end{table}

%% file: tables/appx/ladder_example.tex
\begin{table}[!ht]
\caption{\textbf{What the writing model is given at each level of \collabLadder{}}, for one human-written source: a 744-word personal blog post reviewing a restaurant visit. Every level also asks for a complete, original document of about the source's length; levels 0 to 4 leave everything not specified to the model. The full prompts are in \S\ref{app:prompts} and the repository.}
\label{tab:ladder_example}
\begin{center}\scriptsize
\begin{tabularx}{\linewidth}{r>{\raggedright\arraybackslash}X}
\toprule
Level & What the model receives \\
\midrule
0 & Write a dining and lifestyle blog post reviewing a restaurant visit. \\
1 & Write a first-person dining and lifestyle blog post that reviews a restaurant visit for food and travel enthusiasts in a casual, conversational, and evaluative tone, using dish-by-dish critiques, descriptions of the venue's ambiance, and links to related blog posts. \emph{Focus broadly on a seasonal dining and travel experience in a scenic destination.} \\
2 & Level 1's description, plus the source's global theme: \emph{A winter dining excursion to the refurbished Hydro Majestic's Winter Garden restaurant features a cozy, picturesque atmosphere overlooking Megalong Valley and a mixed experience with a gourmet set menu.} \\
3 & Level 2, plus the order of the source's roles: \emph{Arrange the post so that it begins with an attention hook and purpose framing, continues with two separate event accounts, then moves through a description, a recommendation, a reference aid, and five separate evaluations before ending with a synthesis, social rapport, and navigation.} \\
4 & Level 2, plus the source's full outline, one line per item (15 items), for example: \emph{Attention hook: Contrast people who dislike winter cold with those who welcome the season's fireplaces and substantial comfort food.} \\
5 & The source's description, themes and full outline (15 items), with an instruction to realise every item in order and add nothing, for example: \emph{1. [ATTENTION HOOK] The author opens by contrasting people who dread winter's cold with those who embrace the season for cozy fireplaces and hearty comfort food.} \\
\bottomrule
\end{tabularx}
\end{center}
\end{table}

%% file: tables/appx/t3h.tex
\begin{table}[h]
\caption{\textbf{\collabLadder{}, human sources}: main rows.}
\label{tab:ladder_h:0}
\begin{center}\footnotesize
\setlength{\tabcolsep}{2.5pt}
\adjustbox{max width=\textwidth}{\begin{tabular}{@{}llrrrrrr@{}}
\toprule
 &  &  & \multicolumn{2}{c}{\model} &  &  &  \\
\cmidrule(lr){4-5}
Arm & metric & n & outline & document & \twinmodel & EditLens & Pangram 4 \\
\midrule
source (the person's document) & FPR & 237 & 3.4 & 3.4 & 2.1 & 2.5 & 0.4 \\
level 0 (topic-only) & TPR & 237 & 94.9 & 98.7 & 99.2 & 68.4 & 100.0 \\
level 1 (genre, audience and tone) & TPR & 237 & 91.1 & 98.3 & 99.6 & 59.5 & 100.0 \\
level 2 (themes) & mixed, AI-leaning (fire rate) & 237 & 86.1 & 97.0 & 98.7 & 57.0 & 100.0 \\
level 3 (themes and structure) & mixed, AI-leaning (fire rate) & 237 & 78.9 & 97.5 & 99.6 & 54.0 & 99.6 \\
level 4 (themes and full outline) & mixed & 237 & 23.6 & 75.1 & 94.9 & 29.5 & 98.3 \\
level 5 (the source's outline) & FPR & 237 & 6.8 & 54.4 & 86.5 & 16.0 & 92.4 \\
AUC, levels 0-1 vs source &  &  & 0.991 & 0.998 & 0.999 & 0.980 & 0.998 \\
\bottomrule
\end{tabular}}
\end{center}
\end{table}

%% file: tables/appx/t3a.tex
\begin{table}[h]
\caption{\textbf{\collabLadder{}, AI sources}: main rows.}
\label{tab:ladder_a:0}
\begin{center}\footnotesize
\setlength{\tabcolsep}{2.5pt}
\adjustbox{max width=\textwidth}{\begin{tabular}{@{}llrrrrrr@{}}
\toprule
 &  &  & \multicolumn{2}{c}{\model} &  &  &  \\
\cmidrule(lr){4-5}
Arm & metric & n & outline & document & \twinmodel & EditLens & Pangram 4 \\
\midrule
source (a model's document) & TPR & 263 & 98.1 & 98.5 & 100.0 & 77.2 & 73.4 \\
level 0 (topic-only) & TPR & 263 & 98.1 & 99.6 & 100.0 & 74.9 & 100.0 \\
level 1 (genre, audience and tone) & TPR & 263 & 98.5 & 100.0 & 100.0 & 72.6 & 100.0 \\
level 2 (themes) & TPR & 263 & 99.2 & 100.0 & 100.0 & 75.7 & 100.0 \\
level 3 (themes and structure) & TPR & 263 & 96.6 & 100.0 & 100.0 & 75.7 & 100.0 \\
level 4 (themes and full outline) & TPR & 263 & 98.9 & 100.0 & 100.0 & 78.3 & 100.0 \\
level 5 (the source's outline) & TPR & 263 & 98.5 & 100.0 & 100.0 & 79.8 & 99.2 \\
\bottomrule
\end{tabular}}
\end{center}
\end{table}

%% file: tables/appx/ladder_similarity.tex
\begin{table}[h]
\caption{\textbf{\collabLadder{}: how much of the source each level carries.} \emph{Doc cosine}: text-embedding-3-large similarity between each level's document and its source. \emph{Items kept}: share of the source's outline items whose best match in the level's outline has cosine $\geq \tau = 0.471$, the 95th percentile against unrelated sources of the same format and topic. \emph{Rises}: share of sources whose kept share rises from the previous level (ties count half). Reference rows pair each source with an unrelated one.}
\label{tab:ladder_sim}
\begin{center}\small
\adjustbox{max width=\textwidth}{\begin{tabular}{llrrrrr}
\toprule
 & & \multicolumn{2}{c}{Human sources ($n$ = 237)} & \multicolumn{2}{c}{AI sources ($n$ = 263)} & \\
\cmidrule(lr){3-4}\cmidrule(lr){5-6}
Level & Given to the writer & doc cosine & items kept & doc cosine & items kept & rises (\%, human / AI) \\
\midrule
0 & type + broad subject & 0.518 & 0.301 & 0.537 & 0.348 & -- \\
1 & very high-level topic & 0.572 & 0.450 & 0.589 & 0.471 & 72 / 67 \\
2 & global themes & 0.746 & 0.831 & 0.788 & 0.874 & 94 / 93 \\
3 & themes + role sequence & 0.754 & 0.835 & 0.792 & 0.885 & 51 / 54 \\
4 & themes + full outline & 0.842 & 0.982 & 0.865 & 0.997 & 85 / 78 \\
5 & full outline, realise exactly & 0.868 & 0.990 & 0.889 & 0.999 & 51 / 51 \\
\midrule
\multicolumn{2}{l}{\emph{unrelated source, same format and topic}} & 0.330 & 0.047 & 0.348 & 0.064 & \\
\multicolumn{2}{l}{\emph{unrelated source, same format}} & 0.235 & 0.008 & 0.290 & 0.018 & \\
\bottomrule
\end{tabular}}
\end{center}
\end{table}

%% file: tables/appx/twicetold_final.tex
\begin{table}[!ht]
\caption{\textbf{\twicetold{}: flagged as AI (\%), 1\% global cut} on the 50 stories people wrote from an AI story's outline and the 50 original AI model stories. Evaluating on \twicetold tests whether a detector finds model ideas in human prose, while the model originals are the control.}
\label{tab:twicetold}
\begin{center}\small
\begin{tabular}{lll}
\toprule
Detector & \twicetold & Model originals \\
\midrule
\model{} (outline) & 68.0 (34/50) & 96.0 (48/50) \\
\model{} (document) & 6.0 (3/50) & 100.0 (50/50) \\
\twinmodel{} & 0.0 (0/50) & 100.0 (50/50) \\
EditLens-3B & 0.0 (0/50) & 90.0 (45/50) \\
Pangram 4 & 8.0 (4/50) & 100.0 (50/50) \\
\bottomrule
\end{tabular}
\end{center}
\end{table}

\begin{table}[!ht]
\caption{\textbf{\twicetold Correlations:} Among human rewrites, \twinmodel has a more strongly negative Spearman correlation with Jaccard unigram overlap than \model. For the original AI stories, \model has a weak negative correlation, while the \twinmodel correlation is near zero.}
\label{tab:twicetold-correlations}
\begin{center}\small
\begin{tabular}{lcccc}
\toprule
& \multicolumn{2}{c}{\model} & \multicolumn{2}{c}{\twinmodel} \\
\midrule
Author & Length & Jacc-1g & Length & Jacc-1g \\
\midrule
Human       &  0.07 & -0.52 &  0.60 & -0.65 \\
GPT 5.6 Sol & -0.11 & -0.28 & -0.18 & -0.02 \\
\bottomrule
\end{tabular}
\end{center}
\end{table}

%% file: tables/main_multiling.tex
\begin{table}[t]
\caption{\capMultiling}
\label{tab:multiling}
\begin{center}\footnotesize
\setlength{\tabcolsep}{4pt}
\adjustbox{max width=\textwidth}{\begin{tabular}{@{}lrrrrrrrrrr@{}}
\toprule
 & & \multicolumn{3}{c}{\model} & \multicolumn{3}{c}{\twinmodel} & \multicolumn{3}{c}{Pangram 4} \\
\cmidrule(lr){3-5} \cmidrule(lr){6-8} \cmidrule(lr){9-11}
Language & $n$ & human & level 0 & level 5 & human & level 0 & level 5 & human & level 0 & level 5 \\
\midrule
\multicolumn{11}{l}{\emph{high-resource}} \\
\hspace{1em}German & 704 & 0.0 & 93.0 & 0.3 & 0.0 & 99.6 & 81.1 & 0.0 & 100.0 & 98.3 \\
\hspace{1em}Spanish & 733 & 0.0 & 98.4 & 0.8 & 0.0 & 100.0 & 89.9 & 0.0 & 100.0 & 96.7 \\
\hspace{1em}Chinese & 614 & 0.0 & 92.2 & 1.3 & 0.0 & 99.0 & 88.5 & 0.0 & 99.8 & 98.4 \\
\hspace{1em}Japanese & 729 & 0.0 & 95.2 & 0.4 & 0.0 & 99.9 & 61.7 & 0.0 & 100.0 & 97.8 \\
\multicolumn{11}{l}{\emph{mid-resource}} \\
\hspace{1em}Arabic & 734 & 0.0 & 97.3 & 0.1 & 0.0 & 98.8 & 67.2 & 0.0 & 100.0 & 89.4 \\
\hspace{1em}Bengali & 250 & 0.0 & 97.6 & 0.4 & 0.0 & 76.6 & 2.9 & 0.0 & 89.1 & 28.4 \\
\hspace{1em}Persian & 250 & 0.0 & 96.8 & 1.2 & 0.0 & 99.2 & 45.2 & 0.0 & 100.0 & 70.6 \\
\hspace{1em}Thai & 250 & 0.0 & 98.0 & 0.8 & 0.0 & 98.4 & 19.2 & 0.0 & 100.0 & 68.0 \\
\multicolumn{11}{l}{\emph{low-resource}} \\
\hspace{1em}Tamil & 719 & 0.0 & 95.8 & 0.1 & 0.0 & 67.1 & 1.6 & 0.0 & 69.0 & 0.0 \\
\hspace{1em}Sinhala & 250 & 0.0 & 97.2 & 0.4 & 0.0 & 2.8 & 0.0 & 0.0 & 0.4 & 0.0 \\
\hspace{1em}Yoruba & 250 & 1.2 & 88.0 & 2.4 & 0.0 & 36.8 & 0.0 & 0.0 & 3.2 & 0.0 \\
\hspace{1em}Amharic & 250 & 0.0 & 96.0 & 0.8 & 0.0 & 1.2 & 0.0 & 0.0 & 0.0 & 0.0 \\
\midrule
All 24 languages & 10,160 & 0.1 & 95.3 & 0.7 & 0.0 & 82.5 & 44.5 & 0.0 & 85.4 & 61.9 \\
\bottomrule
\end{tabular}}
\end{center}
\end{table}

%% file: tables/appx/translation.tex
\begin{table}[h]
\caption{\textbf{Translation into English.} 50 sources per language in each of 24 languages, for the human arm and ladder levels 0 and 5 (3,594 documents translated of 3,600). Rates (\%) on the original $\to$ the English translation, global 1\% cut; Pangram~4 at its own threshold. Spearman correlation of original and translated scores over all pairs: \model{} 0.946, \twinmodel{} 0.878.}
\label{tab:translation}
\begin{center}\small
\begin{tabular}{llrrr}
\toprule
Arm & Rate & \model{} & \twinmodel{} & Pangram~4 \\
\midrule
Human sources & FPR & 0.25 $\to$ 0.25 & 0.00 $\to$ 0.25 & 0.00 $\to$ 2.17 \\
Level 0 (model ideas) & TPR & 95.1 $\to$ 94.2 & 74.7 $\to$ 98.2 & 78.5 $\to$ 100.0 \\
Level 5 (human ideas, model prose) & FPR & 1.1 $\to$ 0.8 & 34.9 $\to$ 47.7 & 51.3 $\to$ 97.0 \\
\bottomrule
\end{tabular}
\end{center}
\end{table}

%% file: tables/appx/attacks.tex
\begin{table}[h]
\caption{\textbf{Attacks and humanizers.} TPR (\%) at the global 1\% cut on each evaluation's unattacked model-written rows (clean) and on its attacked or humanized rows; $n$ is the attacked rows. Pangram~4 at its own threshold. The mean is over the evaluations that have both a clean and an attacked arm.}
\label{tab:attacks_app}
\begin{center}\scriptsize\resizebox{\textwidth}{!}{\begin{tabular}{lrrrrrrrrr}
\toprule
 & & \multicolumn{2}{c}{\model{}} & \multicolumn{2}{c}{\twinmodel{}} & \multicolumn{2}{c}{Pangram~4} & \multicolumn{2}{c}{EditLens-3B} \\
\cmidrule(lr){3-4}\cmidrule(lr){5-6}\cmidrule(lr){7-8}\cmidrule(lr){9-10}
Evaluation & $n$ & clean & attacked & clean & attacked & clean & attacked & clean & attacked \\
\midrule
Saha et al.\ (humanizer) & 200 & 67.7 & 86.0 & 100.0 & 99.0 & 98.1 & 78.5 & 89.8 & 0.0 \\
DetectRL-X & 5,296 & 87.3 & 83.8 & 100.0 & 97.1 & 99.7 & 97.9 & 92.9 & 84.1 \\
DetectionAI (StealthGPT) & 7,706 & 90.3 & 57.9 & 99.7 & 64.1 & 99.8 & 94.2 & 94.7 & 4.9 \\
MELD & 9,249 & 88.1 & 84.2 & 99.9 & 72.9 & 100.0 & 100.0 & 98.0 & 53.1 \\
ARB (llm2l) & 7,043 & 97.9 & 97.5 & 100.0 & 99.9 & 99.9 & 99.9 & 98.4 & 98.1 \\
GEDE & 3,604 & 99.4 & 88.7 & 100.0 & 98.6 & 100.0 & 97.6 & 100.0 & 60.7 \\
Perkins et al.\ (6 techniques) & 86 & -- & 93.0 & -- & 100.0 & -- & 96.5 & -- & 94.2 \\
HART (humanizers) & 3,592 & 96.5 & 84.6 & 100.0 & 93.5 & 100.0 & 94.9 & 96.1 & 71.0 \\
\midrule
Mean over the 7 paired evaluations & & 89.6 & 83.2 & 99.9 & 89.3 & 99.7 & 94.7 & 95.7 & 53.1 \\
Drop & & \multicolumn{2}{c}{$-6.4$} & \multicolumn{2}{c}{$-10.6$} & \multicolumn{2}{c}{$-4.9$} & \multicolumn{2}{c}{$-42.6$} \\
\bottomrule
\end{tabular}}
\end{center}
\end{table}

%% file: tables/appx/meld_attacks.tex
\begin{table}[h]
\caption{\textbf{MELD, per attack}: by attack. Fire rate \% at the 1\% cut.}
\label{tab:meld_attacks:3}
\begin{center}\footnotesize
\setlength{\tabcolsep}{2.5pt}
\adjustbox{max width=\textwidth}{\begin{tabular}{@{}lrrrrrrr@{}}
\toprule
 &  &  & \multicolumn{2}{c}{\model} &  &  &  \\
\cmidrule(lr){4-5}
Attack & Arm & n & outline & document & \twinmodel & EditLens & Pangram 4 \\
\midrule
homoglyph & ai\_adversarial (6 attacks) & 1,533 & 86.5 & 64.9 & 48.8 & 4.0 & 100.0 \\
number & ai\_adversarial (6 attacks) & 1,553 & 84.4 & 95.6 & 99.4 & 96.1 & 100.0 \\
synonym & ai\_adversarial (6 attacks) & 1,543 & 76.3 & 71.8 & 91.7 & 43.6 & 100.0 \\
upper\_lower & ai\_adversarial (6 attacks) & 1,564 & 86.4 & 97.3 & 97.8 & 82.4 & 100.0 \\
whitespace & ai\_adversarial (6 attacks) & 1,556 & 85.5 & 97.5 & 96.9 & 89.6 & 100.0 \\
zero\_width\_space & ai\_adversarial (6 attacks) & 1,500 & 86.4 & 0.1 & 0.0 & 0.0 & 100.0 \\
\bottomrule
\end{tabular}}
\end{center}
\end{table}

%% file: tables/appx/detectrl_attacks.tex
\begin{table}[h]
\caption{\textbf{DetectRL-X, per attack}: by attack. Fire rate \% at the 1\% cut.}
\label{tab:detectrl_attacks:4}
\begin{center}\footnotesize
\setlength{\tabcolsep}{2.5pt}
\adjustbox{max width=\textwidth}{\begin{tabular}{@{}lrrrrrrr@{}}
\toprule
 &  &  & \multicolumn{2}{c}{\model} &  &  &  \\
\cmidrule(lr){4-5}
Attack & Arm & n & outline & document & \twinmodel & EditLens & Pangram 4 \\
\midrule
ai\_backtranslation & English · ai\_adversarial & 487 & 87.3 & 90.3 & 97.3 & 66.3 & 99.8 \\
ai\_character\_deletion & English · ai\_adversarial & 496 & 86.9 & 98.0 & 99.8 & 89.1 & 99.8 \\
ai\_character\_insertion & English · ai\_adversarial & 493 & 86.0 & 98.0 & 99.8 & 89.0 & 100.0 \\
ai\_character\_substitution & English · ai\_adversarial & 482 & 86.3 & 98.5 & 100.0 & 91.1 & 99.8 \\
ai\_condensing & English · ai\_adversarial & 497 & 83.7 & 96.8 & 99.0 & 90.5 & 99.8 \\
ai\_decoder\_paraphrasing & English · ai\_adversarial & 486 & 59.7 & 64.2 & 74.5 & 62.8 & 80.2 \\
ai\_encoder\_paraphrasing & English · ai\_adversarial & 491 & 85.7 & 96.3 & 99.4 & 75.4 & 100.0 \\
ai\_expanding & English · ai\_adversarial & 499 & 92.4 & 99.2 & 100.0 & 96.6 & 100.0 \\
ai\_polishing & English · ai\_adversarial & 497 & 88.3 & 98.2 & 100.0 & 91.5 & 99.8 \\
ai\_seq2seq\_paraphrasing & English · ai\_adversarial & 372 & 76.9 & 82.5 & 97.8 & 80.1 & 98.1 \\
ai\_zero\_width\_insertion & English · ai\_adversarial & 496 & 86.1 & 98.6 & 100.0 & 91.3 & 99.6 \\
\bottomrule
\end{tabular}}
\end{center}
\end{table}

%% file: tables/appx/attack_similarity_doc.tex
\begin{table}[h]
\caption{\textbf{Content preservation under attack (documents).} text-embedding-3-large cosine between each clean AI document and its attacked version, and \model{} detection. Same-topic reference: another author's document on the same prompt or source (DetectionAI: another generator, mean 0.769, 95th pct.\ 0.901; DetectRL-X: the paired human document, mean 0.828, 95th pct.\ 0.942). ``Still caught'' and ``missed'' split the attacked documents whose clean version \model{} caught.}
\label{tab:attack_sim_doc:0}
\begin{center}\footnotesize
\setlength{\tabcolsep}{2.5pt}
\adjustbox{max width=\textwidth}{\begin{tabular}{@{}lrrrrrrrrr@{}}
\toprule
Set & Attack & pairs & doc cosine (mean) & below the reference's 95th pct & IdeaLens TPR clean & attacked & cosine, still caught & cosine, missed & Spearman(cosine, P(AI)) \\
\midrule
DetectionAI & StealthGPT humanizer & 7,705 & 0.873 & 68.2\% & 90.3 & 57.8 & 0.879 (n 4369) & 0.862 (n 2587) & +0.10 \\
DetectRL-X & seq2seq paraphrasing & 550 & 0.914 & 34.2\% & 82.9 & 60.7 & 0.979 (n 319) & 0.768 (n 137) & +0.57 \\
DetectRL-X & backtranslation & 550 & 0.945 & 24.0\% & 82.9 & 81.5 & 0.955 (n 430) & 0.808 (n 26) & +0.10 \\
DetectRL-X & decoder paraphrasing & 550 & 0.946 & 32.4\% & 82.9 & 58.4 & 0.993 (n 304) & 0.853 (n 152) & +0.60 \\
DetectRL-X & encoder paraphrasing & 550 & 0.982 & 10.5\% & 82.9 & 83.1 & 0.984 (n 428) & 0.972 (n 28) & +0.06 \\
DetectRL-X & expanding & 550 & 0.982 & 10.4\% & 82.9 & 88.5 & 0.982 (n 440) & 0.994 (n 16) & -0.17 \\
DetectRL-X & character substitution & 550 & 0.988 & 2.9\% & 82.9 & 83.3 & 0.989 (n 428) & 0.983 (n 28) & +0.09 \\
DetectRL-X & condensing & 550 & 0.990 & 1.5\% & 82.9 & 83.6 & 0.991 (n 437) & 0.982 (n 19) & +0.09 \\
DetectRL-X & character deletion & 550 & 0.992 & 1.1\% & 82.9 & 83.6 & 0.992 (n 436) & 0.991 (n 20) & +0.08 \\
DetectRL-X & polishing & 549 & 0.992 & 1.1\% & 82.9 & 84.7 & 0.992 (n 435) & 0.985 (n 20) & +0.06 \\
DetectRL-X & character insertion & 550 & 0.992 & 0.4\% & 82.9 & 84.5 & 0.993 (n 439) & 0.988 (n 17) & +0.05 \\
DetectRL-X & zero width insertion & 550 & 0.993 & 0.0\% & 82.9 & 83.5 & 0.994 (n 434) & 0.992 (n 22) & +0.07 \\
\bottomrule
\end{tabular}}
\end{center}
\end{table}

%% file: tables/appx/attack_similarity.tex
\begin{table}[h]
\caption{\textbf{Content preservation under attack, outline items}: content preserved by humanizers and adversarial attacks, and IdeaLens detection.}
\label{tab:attack_sim_items:0}
\begin{center}\footnotesize
\setlength{\tabcolsep}{2.5pt}
\adjustbox{max width=\textwidth}{\begin{tabular}{@{}lrrrrrrrrrrr@{}}
\toprule
Set & Attack & pairs & doc cosine & items kept & items traceable & items attacked / clean & IdeaLens TPR clean & IdeaLens TPR attacked & kept, caught & kept, missed & Spearman(kept, P(AI)) \\
\midrule
DetectionAI & StealthGPT humanizer & 7,705 & 0.873 & 0.430 & 0.563 & 0.80 & 90.3 & 57.8 & 0.440 (n 4369) & 0.373 (n 2587) & +0.10 \\
DetectRL-X & seq2seq paraphrasing & 550 & 0.914 & 0.620 & 0.609 & 1.12 & 82.9 & 60.7 & 0.807 (n 319) & 0.208 (n 137) & +0.64 \\
DetectRL-X & decoder paraphrasing & 550 & 0.946 & 0.657 & 0.664 & 1.00 & 82.9 & 58.4 & 0.840 (n 304) & 0.335 (n 152) & +0.63 \\
DetectRL-X & backtranslation & 550 & 0.945 & 0.773 & 0.769 & 1.02 & 82.9 & 81.5 & 0.784 (n 430) & 0.615 (n 26) & +0.11 \\
DetectRL-X & expanding & 550 & 0.982 & 0.782 & 0.741 & 1.11 & 82.9 & 88.5 & 0.786 (n 440) & 0.798 (n 16) & -0.02 \\
DetectRL-X & condensing & 550 & 0.990 & 0.793 & 0.801 & 1.00 & 82.9 & 83.6 & 0.801 (n 437) & 0.777 (n 19) & +0.15 \\
DetectRL-X & encoder paraphrasing & 550 & 0.982 & 0.825 & 0.822 & 1.02 & 82.9 & 83.1 & 0.837 (n 428) & 0.810 (n 28) & +0.10 \\
DetectRL-X & polishing & 549 & 0.992 & 0.830 & 0.818 & 1.03 & 82.9 & 84.7 & 0.841 (n 435) & 0.819 (n 20) & +0.13 \\
DetectRL-X & character substitution & 550 & 0.988 & 0.831 & 0.826 & 1.02 & 82.9 & 83.3 & 0.844 (n 428) & 0.840 (n 28) & +0.16 \\
DetectRL-X & character insertion & 550 & 0.992 & 0.835 & 0.833 & 1.02 & 82.9 & 84.5 & 0.846 (n 439) & 0.823 (n 17) & +0.10 \\
DetectRL-X & character deletion & 550 & 0.992 & 0.842 & 0.841 & 1.02 & 82.9 & 83.6 & 0.854 (n 436) & 0.793 (n 20) & +0.13 \\
DetectRL-X & zero width insertion & 550 & 0.993 & 0.844 & 0.836 & 1.03 & 82.9 & 83.5 & 0.854 (n 434) & 0.780 (n 22) & +0.09 \\
\bottomrule
\end{tabular}}
\end{center}
\end{table}

%% file: tables/main_ablations.tex
\begin{table}[t]
\caption{\capAblations}
\label{tab:ablations}
\begin{center}\scriptsize
\setlength{\tabcolsep}{4pt}
\resizebox{\textwidth}{!}{\begin{tabular}{lrlrl}
\toprule
 & \multicolumn{2}{c}{\model{}} & \multicolumn{2}{c}{\modelModernBERT{}} \\
\cmidrule(lr){2-3} \cmidrule(lr){4-5}
 & AUC & TPR / FPR & AUC & TPR / FPR \\
\midrule
\multicolumn{5}{l}{\emph{Held-out test split (48,869 documents, paraphrased outline, global 1\% cut): TPR / realised FPR}} \\
in-domain test & -- & 95.8 / 2.6 & -- & 86.9 / 2.0 \\
\midrule
\multicolumn{5}{l}{\emph{What the outline carries (2,500 test documents, inference-time transforms, variant-matched 1\% cut): AUC, TPR / FPR}} \\
full outline (roles + content) & 0.995 & 94.7 / 2.2 & 0.990 & 87.5 / 2.0 \\
content only (roles stripped) & 0.993 & 93.2 / 2.3 & 0.986 & 83.8 / 2.2 \\
shuffled item order & 0.993 & 90.5 / 1.9 & 0.986 & 82.6 / 2.0 \\
roles only (content stripped) & 0.601 & 2.4 / 0.4 & 0.603 & 2.7 / 0.5 \\
roles-only model, trained (48,869 docs) & -- & -- & 0.913 & 39.0 / 1.5 \\
items scored alone, logit-mean pooled & 0.972 & 60.1 / 1.2 & 0.932 & 47.4 / 1.3 \\
\midrule
\multicolumn{5}{l}{\emph{Pipeline invariance (\% of verdicts unchanged at the global 1\% cut): all / AI / human}} \\
paraphrased vs raw outline ($n$=10,000) & \multicolumn{2}{c}{98.6 / 97.9 / 99.2} & \multicolumn{2}{c}{96.7 / 94.0 / 99.4} \\
source paraphrased before extraction ($n$=2,494) & \multicolumn{2}{c}{98.6 / 97.9 / 99.2} & \multicolumn{2}{c}{96.6 / 94.2 / 99.2} \\
extractor swapped ($n$=1,874) & \multicolumn{2}{c}{98.8 / 98.3 / 99.3} & \multicolumn{2}{c}{96.0 / 93.0 / 99.1} \\
\midrule
\multicolumn{5}{l}{\emph{Excerpt vs full document (raw outline): TPR / FPR}} \\
full document & -- & 96.5 / 3.1 & -- & 88.6 / 2.8 \\
random 500-word excerpt & -- & 91.4 / 2.8 & -- & 71.6 / 2.0 \\
\bottomrule
\end{tabular}}
\end{center}
\end{table}

%% file: tables/appx/t0.tex
\begin{table}[H]
\caption{\textbf{Raw versus paraphrased outline}: main rows.}
\label{tab:t0:0}
\begin{center}\footnotesize
\setlength{\tabcolsep}{2.5pt}
\adjustbox{max width=\textwidth}{\begin{tabular}{@{}llrr@{}}
\toprule
 &  &  & \multicolumn{1}{c}{\model} \\
\cmidrule(lr){4-4}
Rows & metric & n & \% unchanged \\
\midrule
all & all & 10,000 & 98.6 \\
AI & AI & 4,993 & 97.9 \\
human & human & 5,007 & 99.2 \\
\bottomrule
\end{tabular}}
\end{center}
\end{table}

%% file: tables/appx/t1.tex
\begin{table}[H]
\caption{\textbf{Source document paraphrased before extraction}: main rows.}
\label{tab:t1:0}
\begin{center}\footnotesize
\setlength{\tabcolsep}{2.5pt}
\adjustbox{max width=\textwidth}{\begin{tabular}{@{}llrr@{}}
\toprule
 &  &  & \multicolumn{1}{c}{\model} \\
\cmidrule(lr){4-4}
Rows & metric & n & \% unchanged \\
\midrule
all & all & 2,494 & 98.6 \\
AI & AI & 1,294 & 97.9 \\
human & human & 1,200 & 99.2 \\
\bottomrule
\end{tabular}}
\end{center}
\end{table}

%% file: tables/appx/t4.tex
\begin{table}[H]
\caption{\textbf{A different outline extractor}: main rows.}
\label{tab:t4:0}
\begin{center}\footnotesize
\setlength{\tabcolsep}{2.5pt}
\adjustbox{max width=\textwidth}{\begin{tabular}{@{}llrr@{}}
\toprule
 &  &  & \multicolumn{1}{c}{\model} \\
\cmidrule(lr){4-4}
Rows & metric & n & \% unchanged \\
\midrule
all & all & 1,874 & 98.8 \\
AI & AI & 963 & 98.3 \\
human & human & 911 & 99.3 \\
\bottomrule
\end{tabular}}
\end{center}
\end{table}

%% file: tables/appx/t14.tex
\begin{table}[H]
\caption{\textbf{500-word excerpts}: main rows.}
\label{tab:t14:0}
\begin{center}\footnotesize
\setlength{\tabcolsep}{2.5pt}
\adjustbox{max width=\textwidth}{\begin{tabular}{@{}llrr@{}}
\toprule
 &  &  & \multicolumn{1}{c}{\model} \\
\cmidrule(lr){4-4}
Arm & metric & n & outline \\
\midrule
full document, AI & TPR & 1,196 & 96.5 \\
500-word excerpt, AI & TPR & 2,059 & 91.4 \\
full document, human & human (FPR) & 1,153 & 3.1 \\
500-word excerpt, human & human (FPR) & 2,045 & 2.8 \\
AUC, full documents &  &  & 0.995 \\
AUC, excerpts &  &  & 0.992 \\
\bottomrule
\end{tabular}}
\end{center}
\end{table}

%% file: tables/appx/stochasticity.tex
\begin{table}[H]
\caption{\textbf{Repeated extraction.} Each document is extracted five times with the production settings and scored by \model{}. Columns: share (\%) of documents with the same verdict in all five runs, mean pairwise verdict agreement, and mean per-document SD of P(AI). In-domain: the 33 of 42 sampled test documents with all five extractions. Out-of-domain documents are drawn at random or near the cut (P(human) within 0.05 of it; 177 of the 200 are model-written).}
\label{tab:stochasticity}
\begin{center}\small
\adjustbox{max width=\textwidth}{\begin{tabular}{llrrrr}
\toprule
Set & Draw & $n$ & same verdict, 5 runs & pairwise agreement & SD of P(AI) \\
\midrule
In-domain test split & random & 33 & 100.0 & 100.0 & 0.003 \\
DetectRL-X & near the cut & 60 & 65.0 & 84.0 & 0.100 \\
DetectRL-X & random & 60 & 91.7 & 95.7 & 0.037 \\
MELD & near the cut & 80 & 32.5 & 65.8 & 0.118 \\
MELD & random & 80 & 93.8 & 97.3 & 0.018 \\
Saha & near the cut & 60 & 66.7 & 83.7 & 0.064 \\
Saha & random & 60 & 96.7 & 98.3 & 0.007 \\
\midrule
All three OOD sets & random & 200 & 94.0 & 97.1 & 0.021 \\
All three OOD sets & near the cut & 200 & 52.5 & 76.6 & 0.096 \\
\bottomrule
\end{tabular}}
\end{center}
\end{table}

%% file: tables/appx/stochasticity_flips.tex
\begin{table}[H]
\caption{\textbf{Verdict changes by distance from the cut} (400 out-of-domain documents, five extractions each). Distance is the mean over the five runs of the logit of P(human), minus the logit of the cut; a change is any run that disagrees with another.}
\label{tab:stochasticity_flips}
\begin{center}\small
\begin{tabular}{lrr}
\toprule
Logit distance from the cut & documents & verdict changes (\%) \\
\midrule
below $-2$ & 133 & 3.0 \\
$[-2, -0.5)$ & 94 & 57.4 \\
$[-0.5, +0.5)$ & 29 & 100.0 \\
$[+0.5, +2)$ & 23 & 78.3 \\
above $+2$ & 121 & 1.7 \\
\bottomrule
\end{tabular}
\end{center}
\end{table}

%% file: tables/appx/test32.tex
\begin{table}[h]
\caption{\textbf{StoryScope, newer generators}: main rows.}
\label{tab:unseen:0}
\begin{center}\footnotesize
\setlength{\tabcolsep}{2.5pt}
\adjustbox{max width=\textwidth}{\begin{tabular}{@{}llrrrrrr@{}}
\toprule
 &  &  & \multicolumn{2}{c}{\model} &  &  &  \\
\cmidrule(lr){4-5}
Arm & metric & n & outline & document & \twinmodel & EditLens & Pangram 4 \\
\midrule
gpt-5.6-sol (unseen) & TPR & 246 & 78.5\,/\,90.7 & 98.8\,/\,100.0 & 98.8\,/\,99.6 & 1.2\,/\,81.7 & 100.0 \\
gpt-6-astra (unseen) & TPR & 250 & 62.8\,/\,74.8 & 91.6\,/\,99.2 & 83.2\,/\,96.4 & 0.0\,/\,67.6 & 100.0 \\
claude-fable-5-1 (unseen) & TPR & 244 & 51.2\,/\,65.6 & 27.9\,/\,79.1 & 17.2\,/\,35.2 & 0.0\,/\,7.4 & 99.2 \\
claude-opus-5 (unseen) & TPR & 245 & 51.0\,/\,65.7 & 49.4\,/\,84.9 & 41.2\,/\,64.1 & 0.0\,/\,24.9 & 99.2 \\
\bottomrule
\end{tabular}}
\end{center}
\end{table}

%% file: sections/app_feat_analysis.tex
\section{Feature Analysis Details}
\label{app:feat-analysis}

\subsection{Feature discovery}
\label{app:feat-discovery}
We discover features from the outline items of in-domain test documents, separately for each of nine groups: the eight formats (\autoref{tab:corpus}) and one format-agnostic group. For each group, the contrast set is the 5,000 items \model{} scores as most AI-like and the 5,000 it scores as most human-like, balanced across outline roles. GPT-5.6 Sol \citep{openai2026gpt56} (\textit{max} effort) reads each contrast set and proposes features that separate its two sides, giving for each a name, a description, the direction it points (human or AI), its value type (\autoref{tab:feat-output-type}), and how to assign it to an item. This yields 207 features, 23 per group on average. GPT-6 Astra \citep{openai2026gpt6astra} (\textit{extra-high} effort) then merges duplicates and keeps the features proposed across several groups, leaving 79 (see \autoref{tab:feat-list-human} and \autoref{tab:feat-list-ai} for examples); \autoref{tab:feat-count-per-format} shows how many each group contributed.

\subsection{Feature assignment}
\label{app:feat-assignment}
We use a held-out set of documents for feature value assignment. We sample 50 documents from each (format, Pangram silver label) pair, yielding a balanced set of 800 documents from 16 groups. Documents which contain at least one item that was used during the feature discovery process were excluded from this selection process, to guarantee that we are doing this experiment on a held-out set. Then, we assign values for all 79 features on the 13,616 items from these 800 documents, by prompting Gemini 3.8 Flash \citep{gemini38flash} with \textit{medium} effort to rate each item on each of the features, according to each feature's definition and output scale. We use these feature assignment results for the quantitative analysis in \S\ref{app:feat-verification}. 

To verify the effectiveness of this method, we also perform the identical feature assignment task using GPT-5.6 Luna \citep{openai2026gpt56} with \textit{medium} effort. Given the two assignment results, we compute agreement metrics and present them in \autoref{tab:feat-assign-model-comp}. Both raters show a similar number of features firing per item, and a similar distribution of \featscore{} ($r=0.85$). Gwet's AC1 \citep{gwet2014handbook} indicates a substantial agreement, while Krippendorff's $\alpha$ \citep{krippendorff2018content} does not. This is explained by both raters assigning zero so often, while AC1 gives more credit for these matches but $\alpha$ treats them as less informative about raters' agreements. Because individual items naturally have non-zero values for only a small subset of features, we prioritize the AC1 metric and interpret this result as indicating moderate to high agreement.

\subsection{Verifying the legitimacy of our discovered features}
\label{app:feat-verification}
\begin{figure}[t]
\centering
\includegraphics[width=1.00\textwidth]{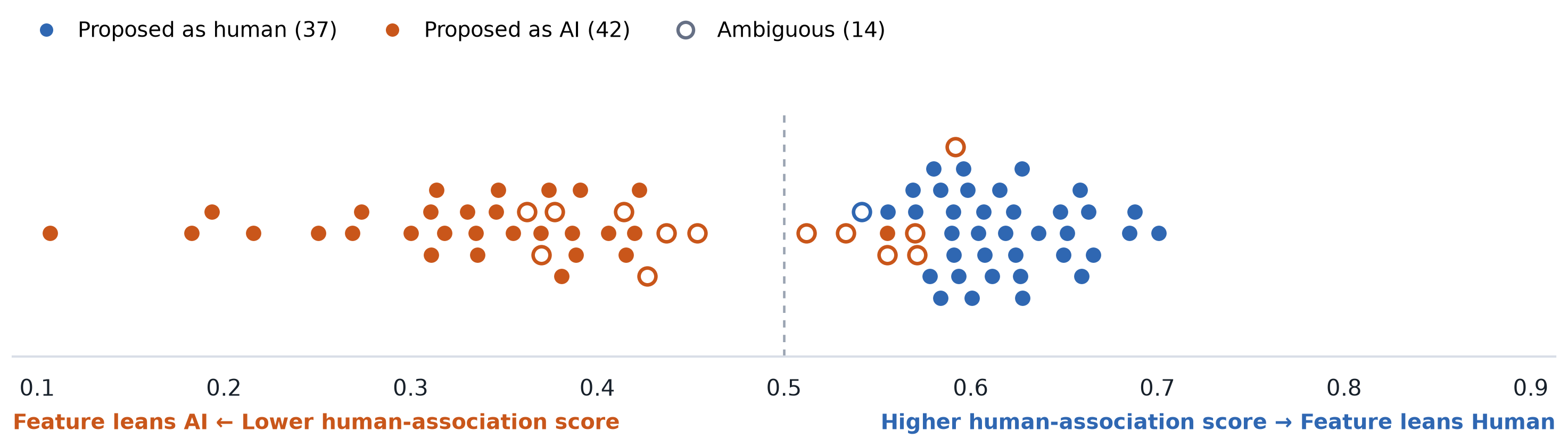}
\caption{\textbf{Distribution of \featscores.} Each point represents one of 79 features; blue and orange indicate features proposed as human- and AI-associated, respectively. The score is the chance that, within the same format, an item where the feature fires receives a higher detector human-likeness score than an item where it does not fire. A score of 0.5 indicates no association. Hollow points have 95\% bootstrap intervals that include 0.5; vertical placement is for visibility only.}
\label{fig:feat-score-dist}
\end{figure}

\begin{figure}[t]
\centering
\includegraphics[width=1.00\textwidth]{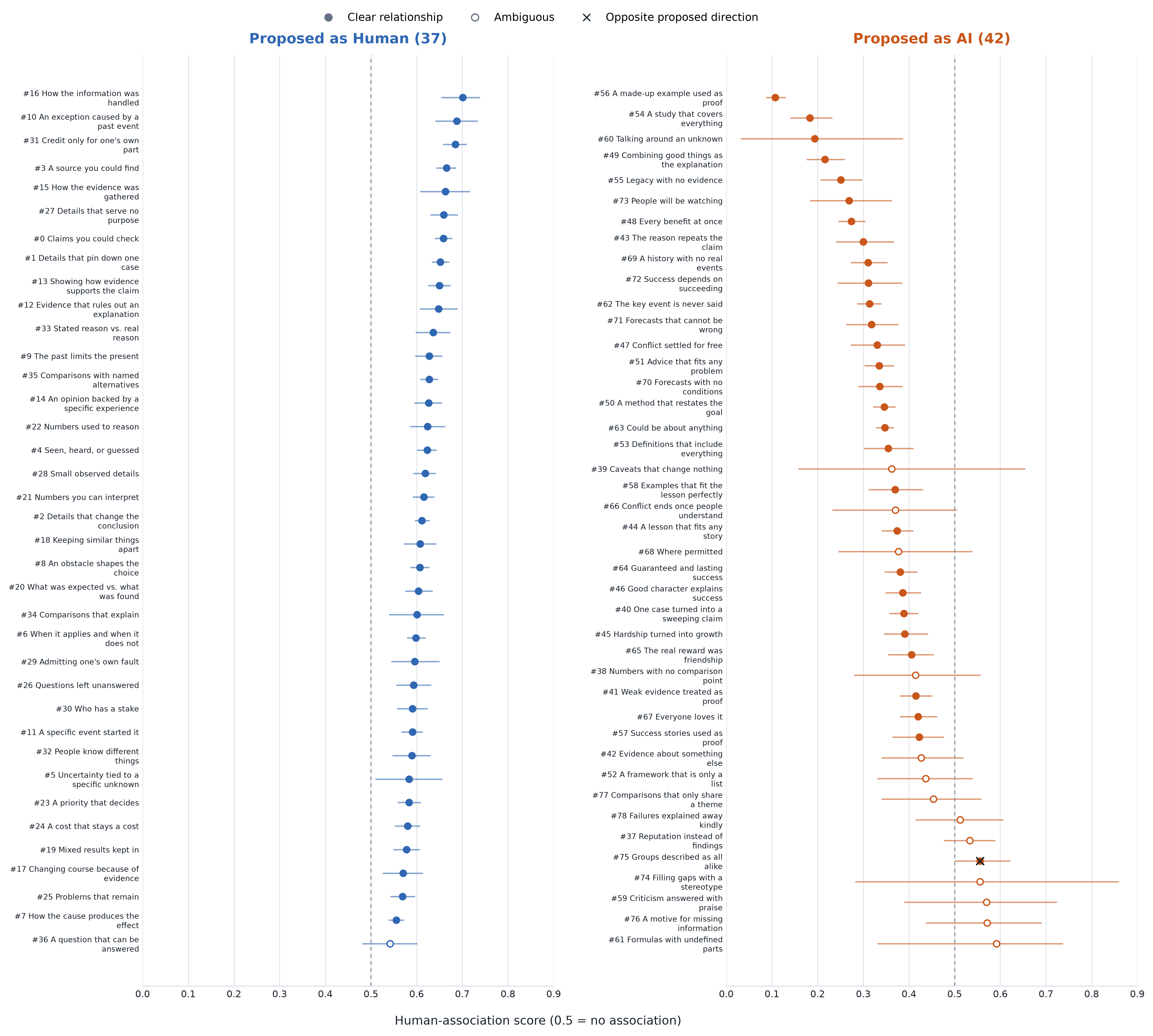}
\caption{\textbf{\featscores\ for individual features.} Points show feature scores and horizontal lines show their 95\% bootstrap intervals, grouped by proposed direction. Human-proposed features are ordered from highest to lowest score, and AI-proposed features from lowest to highest, placing each panel's strongest associations with its proposed direction at the top. Scores above 0.5 lean human; scores below 0.5 lean AI. Hollow points have intervals that include 0.5, and the cross mark indicates the feature whose observed lean opposes its proposed direction.}
\label{fig:feat-score-forest}
\end{figure}

\begin{figure}[t]
\centering
\includegraphics[width=1.00\textwidth]{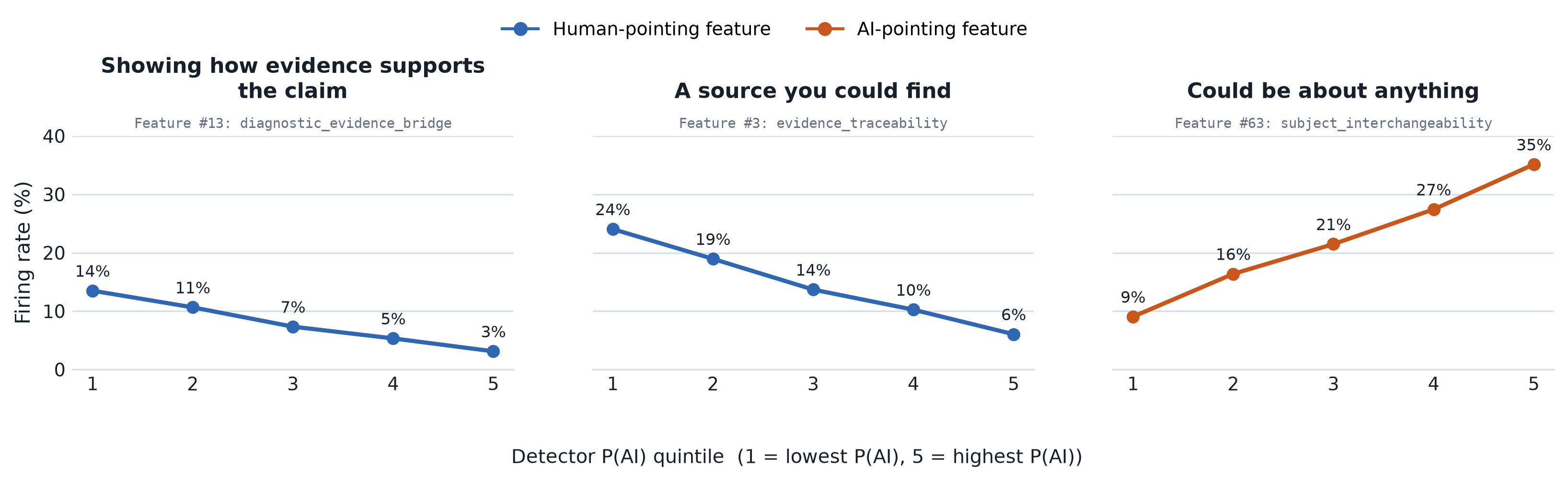}
\caption{\textbf{Feature firing across detector P(AI) quintiles.} Items were ranked by detector P(AI) within each format and divided into five groups, from lowest P(AI) (1) to highest P(AI) (5). Each point shows the share of items in a group where the indicated feature fires. Firing falls across the quintiles for the two human-proposed features (\#13 and \#3) and rises for the AI-proposed feature (\#63).}
\label{fig:feat-quintile-fire}
\end{figure}

We conduct quantitative analysis to verify whether our features align with their proposed directions, and whether they explain our model's behavior. We first assign values for all 79 features on 13,616 held-out items. To quantify the extent of a certain feature pointing towards human ideas, we use \featscore, the chance that an item firing\footnote{The definition of a feature \textit{firing} on an item depends on the type of the feature, and is provided in \autoref{tab:feat-output-type}.} on that feature is detected as more human-like by \model, than an item not firing on that feature.\footnote{A \featscore\ near 1 marks a strongly human-leaning feature, near 0 an AI-leaning one, and 0.5 is neutral.} We find that 64 of the 79 features clearly align with their proposed direction, while only one feature shows a conflict (\autoref{tab:feat-align-summary}, \autoref{fig:feat-score-dist}, \autoref{fig:feat-score-forest}). We also observe that features whose proposed direction was AI tend to fire more on items that \model\ detected as more AI-like, and vice versa (\autoref{fig:feat-quintile-fire}). This is consistent with \model\ relying on these features when making decisions.

\input{tables/feature/diff_summary} %
\input{tables/feature/assignment_model_comparison} %
\input{tables/feature/direction_alignment} %
\input{tables/feature/format_feature_counts} %
\input{tables/feature/output_type} %
\input{tables/feature/feature_results_human_top10} %
\input{tables/feature/feature_results_ai_top10} %

%% file: tables/feature/diff_summary.tex
\providecommand{\diffsummaryfeat}[2]{\path{#1}~(\textsc{#2})}

\begin{table*}[t]
  \centering
  \caption{\textbf{Summary of the four feature families.} Confirmed features have 95\% bootstrap intervals entirely on their proposed side of 0.5. Mean \featscores\ are reported for human- and AI-proposed features. Examples are the two strongest direction-aligned features firing on at least 100 items.}
  \label{tab:feat-diff-summary}
  \small
  \setlength{\tabcolsep}{4pt}
  \renewcommand{\arraystretch}{1.15}
  \begin{tabularx}{\textwidth}{@{}
      >{\raggedright\arraybackslash}p{0.23\textwidth}
      >{\centering\arraybackslash}p{0.12\textwidth}
      >{\centering\arraybackslash}p{0.16\textwidth}
      >{\raggedright\arraybackslash}X@{}}
    \toprule
    \textbf{Difference} &
    \textbf{Confirmed} &
    \textbf{Mean \featscore} &
    \textbf{Strongest examples} \\
    \midrule

    \textbf{Decisive vs. Universal Details} &
    9/9 &
    H: 0.64\newline A: 0.36 &
    \diffsummaryfeat{uninstantiated_change_account}{AI};
    \diffsummaryfeat{interchangeable_remedy_bundle}{AI} \\

    \addlinespace
    \textbf{Functional vs. Apparent Specificity} &
    12/16 &
    H: 0.60\newline A: 0.35 &
    \diffsummaryfeat{nonselective_analytic_agenda}{AI};
    \diffsummaryfeat{abstract_complementarity_as_mechanism}{AI} \\

    \addlinespace
    \textbf{Bounded vs. Overextended Claims} &
    10/10 &
    H: 0.62\newline A: 0.35 &
    \diffsummaryfeat{unearned_legacy_magnitude}{AI};
    \diffsummaryfeat{unbounded_future_extension}{AI} \\

    \addlinespace
    \textbf{Preserved vs. Erased Costs} &
    12/13 &
    H: 0.61\newline A: 0.38 &
    \diffsummaryfeat{guaranteed_independent_cobenefits}{AI};
    \diffsummaryfeat{contingent_exception}{Human} \\

    \bottomrule
  \end{tabularx}
\end{table*}

%% file: tables/feature/assignment_model_comparison.tex
\begin{table*}[t]
  \centering
  \caption{\textbf{Comparison of independent feature assignments on the same 13,616 items from 800 documents.} An aligned feature has a 95\% bootstrap interval entirely on its proposed side of the \featscore's neutral value, 0.5. A valid feature for Krippendorff's $\alpha$ is one that both models rated above zero on at least 100 items.}
  \label{tab:feat-assign-model-comp}
  \small
  \setlength{\tabcolsep}{7pt}
  \renewcommand{\arraystretch}{1.16}
  \begin{tabularx}{\textwidth}{@{}
      >{\raggedright\arraybackslash}X
      >{\centering\arraybackslash}p{0.18\textwidth}
      >{\centering\arraybackslash}p{0.18\textwidth}@{}}
    \toprule
    \textbf{Statistic} & \textbf{Gemini 3.8 Flash} & \textbf{GPT-5.6 Luna} \\
    \midrule
    Features assessed & 79 & 79 \\
    Mean features firing per item & 4.54 & 5.39 \\
    Features aligned with proposed direction & 64/79 & 60/79 \\
    \quad Human-proposed features aligned & 36/37 & 32/37 \\
    \quad AI-proposed features aligned & 28/42 & 28/42 \\
    \addlinespace
    \multicolumn{3}{@{}l}{\emph{Agreement in feature-level associations}} \\
    Same side of 0.5 for the \featscore &
      \multicolumn{2}{c@{}}{73/77 features (95\%)} \\
    Same confidence-interval-based observed lean &
      \multicolumn{2}{c@{}}{66/79 features (84\%)} \\
    Features with confidently opposite leans &
      \multicolumn{2}{c@{}}{0/79} \\
    Correlation of \featscores &
      \multicolumn{2}{c@{}}{$r=0.85$ (77 features)} \\
    \addlinespace
    \multicolumn{3}{@{}l}{\emph{Agreement in same-item feature ratings}} \\
    Exact rating match &
      \multicolumn{2}{c@{}}{0.900} \\
    Ratings jointly equal to zero &
      \multicolumn{2}{c@{}}{0.856} \\
    Mean Gwet's AC1, all features &
      \multicolumn{2}{c@{}}{0.966 (79 features)} \\
    Median Gwet's AC1, all features &
      \multicolumn{2}{c@{}}{0.990 (79 features)} \\
    Mean usable Krippendorff's $\alpha$ &
      \multicolumn{2}{c@{}}{0.480 (on 61 valid features)} \\
    Median usable Krippendorff's $\alpha$ &
      \multicolumn{2}{c@{}}{0.479 (on 61 valid features)} \\
    \bottomrule
  \end{tabularx}

  \vspace{2pt}
  \begin{minipage}{\textwidth}
    \footnotesize\emph{Note.} Two GPT-assigned features never fire, so their \featscores\ are undefined; comparisons of \featscores\ use the other 77. The observed lean is human, AI, or unclear according to whether the 95\% interval lies above, below, or across 0.5. AC1 includes common zero ratings, whereas $\alpha$ more strictly measures whether both models distinguish the same items. The comparison therefore provides stronger evidence for the aggregate feature-direction pattern than for interchangeable item-level ratings.
  \end{minipage}
\end{table*}

%% file: tables/feature/direction_alignment.tex
\begin{table}[t]
  \centering
  \caption{\textbf{Observed directions of the 79 features after assignment.} A feature is aligned when its 95\% bootstrap interval for \featscore\ lies entirely on its proposed side of 0.5 (above for human, below for AI), opposite when it lies entirely on the other side, and ambiguous when it includes 0.5.}
  \label{tab:feat-align-summary}
  \small
  \setlength{\tabcolsep}{3pt}
  \begin{tabular}{@{}lrrrr@{}}
    \toprule
    \textbf{Proposed} & \textbf{Number} & \textbf{Aligned} &
    \textbf{Ambiguous} & \textbf{Opposite} \\
    \midrule
    Human & 37 & 36 & 1 & 0 \\
    AI    & 42 & 28 & 13 & 1 \\
    \midrule
    Total & 79 & 64 & 14 & 1 \\
    \bottomrule
  \end{tabular}
\end{table}

%% file: tables/feature/format_feature_counts.tex
\begin{table*}[t]
  \centering
  \caption{\textbf{Numbers of consolidated features contributed by each discovery group.} A feature may originate from multiple groups, so row totals exceed the 79 unique features.}
  \label{tab:feat-count-per-format}
  \small
  \setlength{\tabcolsep}{4pt}
  \renewcommand{\arraystretch}{1.15}
  \adjustbox{max width=\textwidth}{\begin{tabular}{@{}lccccccccc@{}}
    \toprule
    & \textbf{\shortstack{Mixed\\group}} &
      \shortstack{Personal\\blogs} &
      \shortstack{Creative\\writing} &
      \shortstack{Personal\\about pages} &
      News &
      \shortstack{User\\reviews} &
      Nonfiction &
      \shortstack{Academic\\writing} &
      \shortstack{Knowledge\\articles} \\
    \midrule
    Human & \textbf{12} & 15 & 15 & 14 & 13 & 15 & 12 & 11 & 12 \\
    AI    & \textbf{8}  & 8  & 7  & 9  & 8  & 9  & 12 & 12 & 12 \\
    \bottomrule
  \end{tabular}}
\end{table*}

%% file: tables/feature/output_type.tex
\begin{table}[t]
  \centering
  \caption{\textbf{Output types of the 79 consolidated item-level features.} A feature fires on an item when its assigned value meets the rule in the final column; fractional outputs may take any real value in $[0,1]$.}
  \label{tab:feat-output-type}
  \small
  \setlength{\tabcolsep}{3pt}
  \renewcommand{\arraystretch}{1.15}
  \begin{tabular}{@{}lccc@{}}
    \toprule
    \textbf{Output type} &
    \shortstack{\textbf{Number of}\\\textbf{features}} &
    \textbf{Possible values} &
    \textbf{Fires when} \\
    \midrule
    \texttt{binary\_0\_1} & 5 & $0, 1$ & $x=1$ \\
    \texttt{ordinal\_0\_3} & 73 & $0, 1, 2, 3$ & $x\geq 2$ \\
    \texttt{fraction\_0\_1} & 1 & $x\in[0,1]$ & $x>0.5$ \\
    \bottomrule
  \end{tabular}
\end{table}

%% file: tables/feature/feature_results_human_top10.tex
  \begingroup
  \begin{table}[t]
  \centering
  \scriptsize
  \setlength{\tabcolsep}{2pt}
  \renewcommand{\arraystretch}{1.08}
    \caption{\textbf{Top 10 features proposed as human-pointing} with a fire rate greater than 1.5\%, ordered from highest to lowest \featscore.}
    \label{tab:feat-list-human}
    \begin{adjustbox}{max width=\linewidth,max totalheight=0.88\textheight}
    \begin{tabular}{@{}r
      >{\raggedright\arraybackslash}p{0.22\linewidth}
      >{\raggedright\arraybackslash}p{0.24\linewidth}
      cc
      r
      cc@{}}
    \toprule
    \textbf{No.} & \textbf{Plain-language feature} & \textbf{Raw feature} &
    \textbf{Proposed} & \textbf{Observed} & \textbf{Fire rate} &
    \textbf{Score} & \textbf{95\% interval} \\
    \midrule
    31 & Credit only for one's own part & \path{bounded_credit_allocation} & Human & Human & 6.71\% & 0.68 & [0.66, 0.71] \\
    3 & A source you could find & \path{evidence_traceability} & Human & Human & 14.59\% & 0.67 & [0.64, 0.69] \\
    27 & Details that serve no purpose & \path{incidental_contingency} & Human & Human & 3.93\% & 0.66 & [0.63, 0.69] \\
    0 & Claims you could check & \path{claim_adjudicability} & Human & Human & 84.78\% & 0.66 & [0.64, 0.68] \\
    1 & Details that pin down one case & \path{contingent_factual_anchoring} & Human & Human & 56.44\% & 0.65 & [0.63, 0.67] \\
    13 & Showing how evidence supports the claim & \path{diagnostic_evidence_bridge} & Human & Human & 7.98\% & 0.65 & [0.63, 0.67] \\
    33 & Stated reason vs. real reason & \path{public_private_motive_divergence} & Human & Human & 1.51\% & 0.64 & [0.60, 0.67] \\
    9 & The past limits the present & \path{inherited_constraint_dependence} & Human & Human & 3.87\% & 0.63 & [0.60, 0.66] \\
    35 & Comparisons with named alternatives & \path{comparison_baseline_resolution} & Human & Human & 9.61\% & 0.63 & [0.61, 0.65] \\
    14 & An opinion backed by a specific experience & \path{situated_experiential_warrant} & Human & Human & 10.43\% & 0.63 & [0.59, 0.66] \\
    \bottomrule
    \end{tabular}
    \end{adjustbox}
  \end{table}
  \endgroup

%% file: tables/feature/feature_results_ai_top10.tex
  \begingroup
  \begin{table}[t]
  \centering
  \scriptsize
  \setlength{\tabcolsep}{2pt}
  \renewcommand{\arraystretch}{1.08}
    \caption{\textbf{Top 10 features proposed as AI-pointing} with a fire rate greater than 1.5\%, ordered from lowest to highest \featscore.}
    \label{tab:feat-list-ai}
    \begin{adjustbox}{max width=\linewidth,max totalheight=0.88\textheight}
    \begin{tabular}{@{}r
      >{\raggedright\arraybackslash}p{0.22\linewidth}
      >{\raggedright\arraybackslash}p{0.24\linewidth}
      cc
      r
      cc@{}}
    \toprule
    \textbf{No.} & \textbf{Plain-language feature} & \textbf{Raw feature} &
    \textbf{Proposed} & \textbf{Observed} & \textbf{Fire rate} &
    \textbf{Score} & \textbf{95\% interval} \\
    \midrule
    48 & Every benefit at once & \path{guaranteed_independent_cobenefits} & AI & AI & 2.44\% & 0.27 & [0.24, 0.30] \\
    69 & A history with no real events & \path{uninstantiated_change_account} & AI & AI & 1.61\% & 0.31 & [0.27, 0.35] \\
    62 & The key event is never said & \path{central_semantic_undercommitment} & AI & AI & 5.21\% & 0.31 & [0.29, 0.34] \\
    51 & Advice that fits any problem & \path{interchangeable_remedy_bundle} & AI & AI & 2.43\% & 0.33 & [0.30, 0.37] \\
    50 & A method that restates the goal & \path{goals_substituted_for_operations} & AI & AI & 4.45\% & 0.35 & [0.32, 0.37] \\
    63 & Could be about anything & \path{subject_interchangeability} & AI & AI & 21.90\% & 0.35 & [0.33, 0.37] \\
    53 & Definitions that include everything & \path{boundary_free_construct} & AI & AI & 1.56\% & 0.35 & [0.30, 0.41] \\
    44 & A lesson that fits any story & \path{generic_symbolic_lesson} & AI & AI & 3.47\% & 0.37 & [0.34, 0.41] \\
    64 & Guaranteed and lasting success & \path{contingent_outcome_assurance} & AI & AI & 2.35\% & 0.38 & [0.35, 0.42] \\
    46 & Good character explains success & \path{moralized_causal_shortcut} & AI & AI & 2.06\% & 0.39 & [0.35, 0.43] \\
    \bottomrule
    \end{tabular}
    \end{adjustbox}
  \end{table}
  \endgroup

%% file: sections/appendix_prompts.tex
\newtcolorbox{pbox}[1]{enhanced, breakable, colback=white, colframe=black!35, boxrule=0.5pt, arc=1.5pt,
  left=4pt, right=4pt, top=2pt, bottom=2pt, fonttitle=\small\bfseries, coltitle=black, colbacktitle=black!6,
  title={#1}, before skip=6pt, after skip=6pt, fontupper=\small}
Each box condenses one prompt to its instructions and inputs; placeholders are in \texttt{\{\{braces\}\}}. The full prompts, their output schemas and the extraction exemplars' reference outlines are in the repository (\url{https://github.com/RishanthRajendhran/IdeaLens/tree/main/paper/prompts}); the exemplar documents, which come from the training corpus, are released with the corpus.

\subsection{Outline pipeline}
\begin{pbox}{Format gate (Gemini 3.7 Flash)}
Classify the document into one of WebOrganizer's 24 format categories (letter choices A--X), reading the whole document. Only the eight long-form formats with a role vocabulary continue; the rest are out of scope.\\
\textit{Input:} \texttt{\{\{DOCUMENT\}\}}.
\end{pbox}
\begin{pbox}{Role discovery, step 1: base role proposal}
From general knowledge of \texttt{\{\{FORMAT\}\}} and its description, propose an inventory of the recurring functions that substantive, rhetorical and bounded structural items perform in such documents. No example documents are shown. Roles must be functions, not topics or surface formatting, must recur across subtypes and subject areas (a topic list is given as a generality check only), and each comes with a definition, an assignment test and its nearest functional distinction.
\end{pbox}
\begin{pbox}{Role discovery, step 2: outline and role assignment (200 documents per format)}
Extract a flat, ordered outline of the document's latent plan (what the writer chose to say and in what order, abstracted from wording). Then assign each item one primary role, preferring a base role whenever it fits and inventing a new candidate role only when every base role would misstate the item's function. Roles are decided in context: what the item contributes and why it appears at that point.
\end{pbox}
\begin{pbox}{Role discovery, step 3: consolidation}
Given every base and invented candidate with its definition, assignment test, usage count and real example items, merge candidates that name the same function, drop rare or topic-specific ones, and return a parsimonious final vocabulary that is cross-topic and usable with one label per item.
\end{pbox}
\begin{pbox}{Labelled outline extraction (Gemini 3.1 Pro or 3.7 Flash, six exemplars)}
Infer the document's latent composition and express it as a flat, ordered, role-labelled outline from which a competent writer could reproduce a substantively, rhetorically and structurally equivalent document. Steps: filter extraction artefacts; identify the content commitments; segment them into items at a canonical granularity; write each item's content in the item's own words, keeping load-bearing names, numbers, dates and quotations; assign one role from the format's vocabulary (or \texttt{Other}); describe the document type and its global themes.\\
\textit{Inputs:} \texttt{\{\{FORMAT\}\}}, its description, \texttt{\{\{ROLES\_BLOCK\}\}}, six complete documents with their reference outlines, \texttt{\{\{DOCUMENT\}\}}.
\end{pbox}
\begin{pbox}{Paraphrasing (Gemini 3.1 Pro; the source document is never shown)}
Given only the role-labelled outline, rewrite every field in concise, neutral, self-contained plan-note language, reconstructed from its meaning rather than edited, so that paraphrases of the same document converge on the same canonical wording; share no run of more than a few consecutive words with the input except protected material (proper nouns, numbers, dates, technical terms, quoted text, and items marked verbatim). Keep the JSON structure, every theme and item, their ids, roles and order unchanged, and never add, remove, strengthen, weaken or move a proposition. \textit{Input:} the outline as JSON; returns the rewritten JSON.
\end{pbox}
\begin{pbox}{Classifier input (IdeaLens)}
System: \textit{``Given a role-labelled outline of a document, answer with one word: human if the source document was human-written, ai if it was AI-generated.''} User: the outline as one line per item, \texttt{[Role] content}. The score is the probability of the \texttt{ai} token. \twinmodel{} uses the same template with the raw document and \textit{``Given a document, \ldots''}.
\end{pbox}

\subsection{Evaluation construction}
\begin{pbox}{\collabLadder{} and \xLadder{}: briefs (one call per source)}
From the source's outline, write five matched, ordinary-sounding requests for the same document type that reveal progressively more of the source: a minimal topic-only line naming only the document type and broad subject (level 0); the type with its genre, audience and tone (level 1); plus the global themes (level 2); themes and structure (level 3); and the full outline as a starting point (level 4). Treat the source as data; never copy its wording. \textit{Inputs:} \texttt{\{\{DOCUMENT\_DESCRIPTION\}\}}, \texttt{\{\{GLOBAL\_THEMES\}\}}, \texttt{\{\{FULL\_OUTLINE\}\}}; \xLadder{} adds \texttt{\{\{OUTPUT\_LANGUAGE\}\}} and \texttt{\{\{TARGET\_LENGTH\}\}}. (GPT-5.6 Sol for \collabLadder{}, Gemini 3.7 Flash for \xLadder{}.)
\end{pbox}
\begin{pbox}{\collabLadder{} and \xLadder{}: generation from a brief (levels 0 to 4)}
System prompt: ``You are a skilled writer.''; the user message is the brief itself. (\collabLadder{}: GPT-5.6 Sol, Terra or Luna, one per source; \xLadder{}: Gemini 3.7 Flash, level 0 only, in the page's language.)
\end{pbox}
\begin{pbox}{Level 5: realise the person's outline, adding nothing}
Write the complete document the outline represents, realising every item in the given order as natural prose. The description and global themes are context only. Add nothing the outline does not contain: no new introduction, conclusion, framing or section. \xLadder{} adds one line naming the output language.
\end{pbox}
\begin{pbox}{Translation into English (translation study)}
Translate every marked segment \texttt{[S1]}\ldots\texttt{[Sn]} exactly once and in order; nothing added, removed or improved; meaning first, then structure, then fluency; never add or remove a negation; keep names, numbers, URLs and English text.
\end{pbox}

\subsection{Feature Analysis}
\begin{pbox}{Feature Discovery}
Each item is one atomic idea extracted from a source document. Below are \texttt{\{\{N\}\}} HUMAN-source items and \texttt{\{\{N\}\}} AI-source items. The two sides are role-matched: every role name contributes the same number of items to each side. Differences in role usage are therefore zero by construction and cannot be a feature. Read them as contrast sets, then propose discriminating features of individual items. Propose features that separate a single human-source item from a single AI-source item.\\
\textit{Inputs:} \texttt{\{\{FORMAT\}\}}, \texttt{\{\{HUMAN-SOURCE ITEMS\}\}}, \texttt{\{\{AI-SOURCE ITEMS\}\}}
\end{pbox}
\begin{pbox}{Feature Consolidation}
You are consolidating a large pool of candidate features for a human-vs-AI text classifier. The features were proposed independently by reading individual items of document outlines from 9 sources: eight document formats separately and one cross-format agnostic pass. Your job is to merge the pool into a clean, deduplicated, canonical feature bank. Every canonical feature must be grounded in the raw features it subsumes.\\
\textit{Input:} \texttt{\{\{FEATURES\}\}}
\end{pbox}
\begin{pbox}{Feature Assignment}
You will be given a numbered set of items. Rate each item on each of the following features, strictly per each feature's definition and output scale. Give each item exactly the ratings it would get if it were the only item you were shown. Judge an item only from its own format, role label, and text. They are listed in random order and may come from different documents. Do not use one item to explain, complete, support, or contradict another, and do not consider how items relate to, follow, or repeat each other.\\
\textit{Inputs:} \texttt{\{\{FEATURES\}\}}, \texttt{\{\{ITEMS\}\}}
\end{pbox}

\subsection{\twicetold Curation}
\begin{pbox}{TwiceTold Instructions}
Please write a 500-word story that follows the given outline as closely as possible. The outline includes the genre, a brief set of instructions (e.g., perspective, style, etc.), global themes and a list of 8-12 items describing the plot. \\
1. Retain the original order of events\\
2. Do not omit details provided in the outline\\
3. You may add minimal details to make stories coherent\\

\textit{Inputs:} \texttt{\{\{GENRE\}\}}, \texttt{\{\{DOCUMENT DESCRIPTION\}\}}, \texttt{\{\{GLOBAL THEMES\}\}}, \texttt{\{\{OUTLINE\}\}}
\end{pbox}

\subsubsection{Annotation Details}
All the computer science graduate students recruited for this task consented to their stories being used for research purposes. Students took approximately 1-2 hours to complete writing each story, depending on the length and complexity of the outline.